\documentclass{ecai} 

\usepackage{latexsym}
\usepackage{amssymb}
\usepackage{amsmath}
\usepackage{amsthm}
\usepackage{booktabs}
\usepackage{enumitem}
\usepackage{graphicx}
\usepackage{color}
\usepackage{caption}
\usepackage{subcaption}
\usepackage{url}
\usepackage{seqsplit}
\usepackage{fancyhdr} 
\newcommand{\BibTeX}{B\kern-.05em{\sc i\kern-.025em b}\kern-.08em\TeX}

\begin{document}


\begin{frontmatter}


\paperid{93} 


\title{Generative Pretrained Embedding and Hierarchical Irregular Time Series Representation for Daily Living Activity Recognition}


\author[A]{\fnms{Damien}~\snm{Bouchabou}\orcid{0000-0003-3623-3626}\thanks{Email: damien.bouchabou@gmail.com} }
\author[B]
{\fnms{Sao Mai}~\snm{Nguyen}\orcid{0000-0003-0929-0019}\thanks{Email: nguyensmai@gmail.com}}

\address[A]{U2IS, ENSTA, IP Paris, France}

\address[B]{FLOWERS, U2IS, ENSTA, IP Paris \& Inria and IMT Atlantique Lab-STICC, UMR 6285}


\begin{abstract}
Within the evolving landscape of smart homes, the precise recognition of daily living activities using ambient sensor data stands paramount. This paper not only aims to bolster existing algorithms by evaluating two distinct pretrained embeddings suited for ambient sensor activations but also introduces a novel hierarchical architecture. We delve into an architecture anchored on Transformer Decoder-based pre-trained embeddings, reminiscent of the GPT design, and contrast it with the previously established state-of-the-art (SOTA) ELMo embeddings for ambient sensors. Our proposed hierarchical structure leverages the strengths of each pre-trained embedding, enabling the discernment of activity dependencies and sequence order, thereby enhancing classification precision. To further refine recognition, we incorporate into our proposed architecture an hour-of-the-day embedding. Empirical evaluations underscore the preeminence of the Transformer Decoder embedding in classification endeavors. Additionally, our innovative hierarchical design significantly bolsters the efficacy of both pre-trained embeddings, notably in capturing inter-activity nuances. The integration of temporal aspects subtly but distinctively augments classification, especially for time-sensitive activities. In conclusion, our GPT-inspired hierarchical approach, infused with temporal insights, outshines the SOTA ELMo benchmark.
\end{abstract}

\end{frontmatter}

\lhead{ }
\rhead{ }
\lfoot{Bouchabou, D. and Nguyen, S. M. (2024). Generative Pretrained \\Embedding and Hierarchical Irregular Time Series Representation \\for Daily Living Activity Recognition. ECAI 2024(4764 - 4771). IOS Press.
}
\rfoot{http://doi.org/10.3233/FAIA241075}

\section{Introduction}
Recognizing and analyzing temporal event sequences is a fundamental challenge in artificial intelligence and data science, involving pattern identification in time-ordered observations. This task requires capturing short and long-term dependencies, handling irregular sampling, and accounting for noise and variability.
Human Activity Recognition (HAR) in smart homes exemplifies this challenge. With the Internet of Things (IoT) advancements, homes are increasingly equipped with ambient sensors (e.g., motion, door open/close, temperature), with the outlook of offering services for improving the quality of daily life or health. HAR algorithms aim to recognize Activities of Daily Living (ADL) like cooking or sleeping from these sensor activations, forming the basis for health and well-being services.
Deep learning has emerged as a leading approach in HAR due to its proficiency in interpreting raw sensor data \cite{gochoo2018unobtrusive,mohmed2020employing,wang2016human,singh2017convolutional}. However, challenges persist \cite{bouchabou2021survey}, including noisy, irregularly sampled data from battery-powered sensors, limited contextual information due to privacy concerns, and the complexity of ADLs involving variable-length actions with multilevel temporal dependencies. These factors strain traditional modeling methods.

To address long-term dependencies, hierarchical models like ontologies \cite{Hong2009PMC} and hidden Markov models \cite{Asghari2019} have been proposed. Recent deep learning paradigms offer single-level sequence models \cite{medina2018ensemble,liciotti_lstm,sedky2018evaluating}, but struggle with irregular time series and long-range contexts. Recurrent Neural Networks (RNN) and language models (ELMo \cite{peters2018deep}, GPT-2 \cite{radford2018improving}) show promise but have limitations for longer-range contexts \cite{bouchabou2021using,10179111}.

This work tackles a multivariate non-Markovian irregular time series in the context of ADLs recognition into smart homes, by examining event timestamps, contextual encoding, and long-term dependency. Our approach combines: (1) attention mechanisms for discerning the importance of sensor signals across a whole sequence, (2) pre-trained generative transformer embeddings capturing sensor interrelations, (3) a hierarchical model emphasizing activity succession for long-horizon dependency, and (4) a temporal encoding model to harness the timing of events.

We propose a multi-timescale architecture for wider temporal dependency, contextualizing non-Markov time series events. 

\textbf{Code available at:} \seqsplit{https://github.com/dbouchabou/Generative-Pretrained-Embedding-and-Hierarchical-Representation-to-Unlock-ADL-Rhythm-in-Smart-Homes.git}.

\textbf{Annexes available at:} \seqsplit{https://github.com/dbouchabou/Generative-Pretrained-Embedding-and-Hierarchical-Representation-to-Unlock-ADL-Rhythm-in-Smart-Homes/blob/Master/Paper/Annexes.pdf}.

\section{Related Works}

\subsection{Human Activity Recognition and Pre-trained Embeddings}

Recent advances in deep learning have significantly impacted various fields, including HAR. In HAR, sensor-based deep learning techniques are pertain to Convolutional Neural Networks \cite{singh2017convolutional, mohmed2020employing}, autoencoders \cite{wang2016human}, semantics-based approaches \cite{yamada2007applying}, and sequence models \citep{ghods2019activity2vec}. Despite their efficacy, these methods often struggle with temporal aspects, long-term dependencies, and pattern similarities crucial for recognizing ADLs.

In \cite{huang2023human} Huang et Zhang have employed a transformer architecture \cite{vaswani2017attention} with handcrafted features to classify pre-segmented activities on the Aruba dataset from the CASAS benchmark \cite{cook2012casas}. However, their approach had limitations: it omitted the infrequent "respirate" activity and the "Other" category, which includes unidentified activities that share patterns with recognized ones. Additionally, their model had difficulty distinguishing similar activities like "wash dishes" and "meal preparation," and did not fully utilize deep learning's potential for automatic feature extraction.

Advancements in self-supervised learning and sequence modeling, such as bi-LSTM \cite{liciotti_lstm}, ELMo \cite{bouchabou2021using}, and GPT-2 \cite{10179111}, have influenced HAR. \cite{bouchabou2021using} enhanced bi-LSTM classifiers with a frozen, pre-trained ELMo-based sensor embedding. \cite{10179111} used a GPT-2 architecture for predicting sensor events, but this approach was limited to feature engineering and did not leverage GPT-2's capabilities for classification tasks. Additionally, it lacked extensive analysis of embeddings, attention mechanisms, and ablation studies.

\subsection{Hierarchical Models of Actions}

A hierarchical approach is crucial for modeling human actions and recognizing ADLs in smart homes. Various hierarchical models have been developed using the CASAS dataset \cite{cook2012casas}. For example, \cite{Hong2009PMC} explored ontology models for context-aware activities, and \cite{Asghari2019} used Hierarchical Hidden Markov Models. These methods, however, have limitations in recognizing long-term dependencies.

Hierarchical LSTM models have also been applied to activity recognition with wearable sensors \cite{Wang2020CSSP} and RGB-D videos \cite{devanne2019recognition}. \cite{devanne2019recognition} found that hierarchical structures improved recognition performance by processing more information from longer time windows and better understanding activity relationships, highlighting the importance of multi-level time dependencies.

ADLs, such as cooking or cleaning, can vary significantly daily based on the inhabitant's context and goals. Each activity comprises a sequence of unit actions, as identified in \cite{devanne2019recognition} and \cite{Wang2020CSSP}, organized to achieve a distant goal. Activities may also be interdependent, influencing and relying on each other.

\section{Approach}

\subsection{Problem Formulation}
Let $e_i = (i_i, v_i, t_i)$ be a sensor event, where $i_i \in \{1,\ldots,n\}$ is the sensor ID, $v_i \in \mathcal{V}$ is the sensor value (binary or scalar), and $t_i \in \mathcal{T}$ is the timestamp. 
Events arrive asynchronously and  form a history, or event stream. It is denoted as $h_t$, where $t$ represents the current timestamp :\\
$ h_t = \langle e_1=(i_1, v_1, t_1), e_2=(i_2, v_2, t_2), \ldots, e_k=(i_k, v_k, t_k) \rangle \\ \text{ where } \forall i \in \{1,..,k\}, t_i \leq t $

We note $\mathcal{H}$ the set of histories supposed segmented into sensor event sequences, defined as a set of sensor events $s_j = \{e_1,\ldots,e_L\}$. We note $\mathcal{S}$ as the set of all sequences and $t_{s_j} = max(\{t_i, \text{ where } e_i=(i_i, v_i, t_i) \in s_j\} )$ the timestamp of the last event of $s_j$.

Let $\mathcal{A} = \{a_1,\ldots,a_m\}$ be the set of activity labels. The HAR problem learns to associate a sequence to a label given the history $f: \mathcal{S} \times \mathcal{H} \rightarrow \mathcal{A}$, where $f(s_j,h_{t_{s_j}}) \mapsto a_k$. 
Our approach decomposes $f$ into three functions :\\
(1) Embedding $E_{\text{sequence}}: \{(i_1, v_1),\ldots,(i_L, v_L)\} \rightarrow \mathbb{R}^d_{\text{sensor}}$\\
(2) Embedding $E_{\text{timestamp}}: \{t_1,\ldots,t_L\} \rightarrow \mathbb{R}^d_{\text{time}}$\\
(3) Classification $C: (\mathbb{R}^d_{\text{sensor}}, \mathbb{R}^d_{\text{time}})^c \rightarrow \mathcal{A}$, with $c$ context length. Our implementation uses $c=3$.

The hierarchical model is expressed as: \\
$f(s_j,h_{t_{s_j}}) = 
C(E_{\text{sequence}}(s_{j-2}), E_{\text{timestamp}}(s_{j-2}), E_{\text{sequence}}(s_{j-1}), \\ E_{\text{timestamp}}(s_{j-1}), E_{\text{sequence}}(s_j), E_{\text{timestamp}}(s_j))$ 
\\where $s_{j-2}, s_{j-1}, s_j$ are three consecutive event sequences. Dataset details  are in Annex \ref{sec:dataset_details}.

\subsection{Decoder Transformer Embedding}

In the realm of HAR within smart homes, our approach builds upon the innovative application of pre-trained ELMo embeddings for ambient sensors, as proposed by \cite{bouchabou2021using}, and the transformative successes of transformer architectures like BERT \cite{devlin2018bert} and GPT \cite{radford2018improving}. We advance this field by adopting a Transformer decoder embedding architecture to enhance the performance of HAR algorithms.

Our primary goal is to improve HAR algorithms' ability to accurately categorize pre-segmented, temporally connected sequences of sensor events into identifiable daily activities. For instance, consider a scenario where a resident prepares to take a bath, involving a sequence of sensor-triggered events: leaving a room, going to the bathroom, opening the bathroom door, turning on the light, entering the bathroom, and using the shower. These activities imply a sequential relationship, where actions such as the bathroom door opening and entering the bathroom precede shower usage. This sequencing becomes critical in a multi-resident setting, where concurrent activities, like cooking, may occur. The model must distinguish related events from unrelated ones, focusing its attention to accurately extract the current activity context.

While ELMo and BERT excel in contextual understanding within sequences, their training approach limits their efficacy in addressing contexts beyond immediate sequences. They primarily capture context within pre-segmented sequences. In contrast, a Transformer decoder-based architecture like GPT, trained on data chunks that do not correspond to pre-segmented activity sequences, is more suitable for our needs as it learns longer-term dependencies and broader cause-and-effect relationships among sensor events.

Furthermore, the use of both past and future information in ELMo and BERT architectures to contextualize elements within a sequence prevents the reuse of these approaches for real-time ADL recognition. The Transformer decoder architecture, exemplified by GPT, addresses this limitation through its causal masking feature, which ensures that the network considers only previous sensor states to predict subsequent ones. This training approach enables the network to recognize and understand cause-and-effect relationships effectively.

\begin{figure*}[hbt]
\centering
\includegraphics[width=0.68\textwidth]{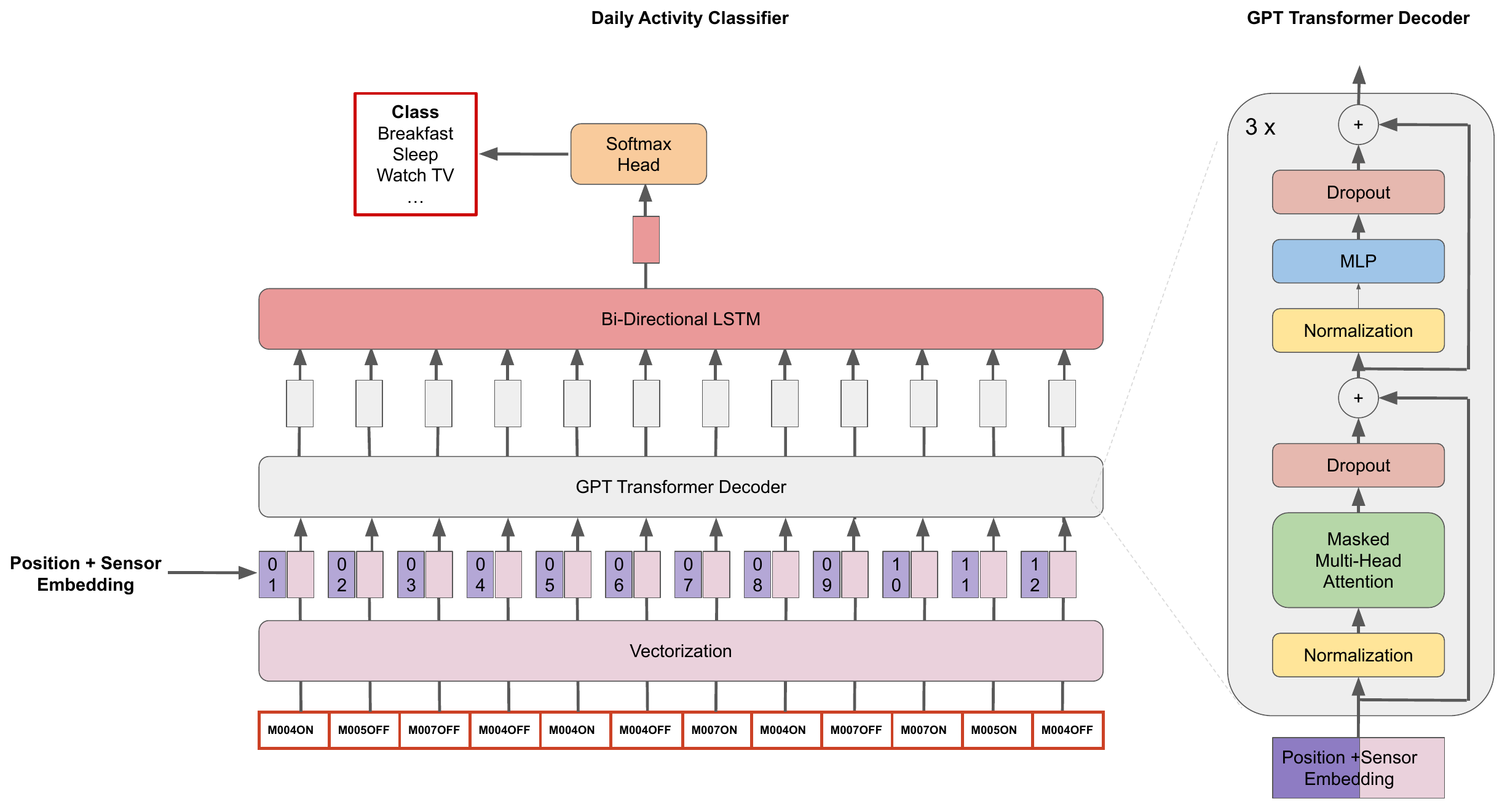} 
\caption{Model architecture of GPTAR and its GPT transformer decoder. GPTAR  embeds the sensor signal  with 3 layers of GPT transformer decoder embedding and a bi-LSTM. 
The illustration of the Transformer decoder was inspired by \cite{vaswani2017attention}
}
\label{fig:proposed_architecture}
\end{figure*}

As illustrated in Fig. \ref{fig:proposed_architecture}, we integrate a pre-trained, frozen GPT Transformer decoder embedding as a replacement for the previous ELMo setup within our single-level temporal  framework. This GPT embedding encodes sensor events $E_{\text{sequence}}$, which are subsequently processed by  a Bi-LSTM layer, followed by a classification head with a Softmax layer $C$. This model forms the foundation of our classifier GPTAR.

\subsection{Multi-Timescale Architecture}
 
\begin{figure*}[hbt]
\centering
\includegraphics[width=\textwidth]{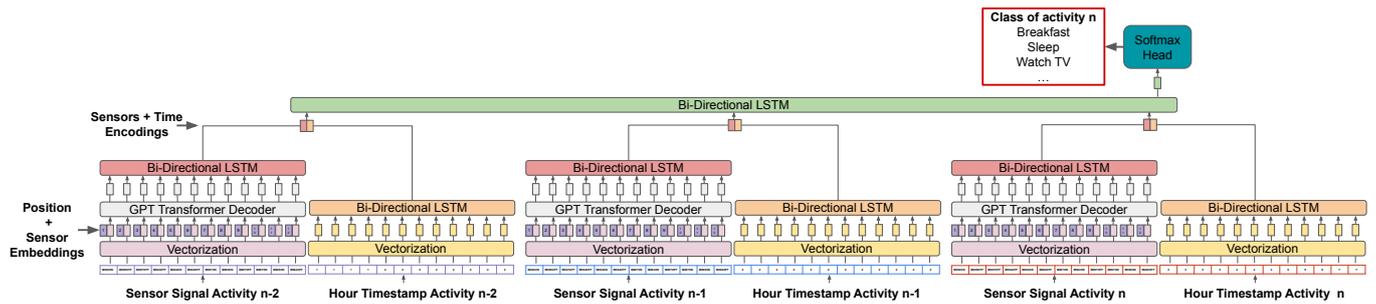} 
\caption{Complete architecture of the Generative Pre-trained Transformer for Hierarchical Activity Recognition (GPTHAR), composed of 3 low-level modules to compute 3 successive activities, and a top-level composed of a bi-LSTM and a softmax classifier. The low-level module processes in parallel the hour timestamp with a bi-LSTM and the sensor signal  with a GPT transformer decoder embedding and a bi-LSTM}.
\label{fig:complete_arch}
\end{figure*}

Motivated by the need to discern intricate temporal relationships both within individual activities and across successive ones, we designed a hierarchical architecture, as depicted in Fig. \ref{fig:complete_arch}. Recognizing that human behaviors manifest across diverse observational scales, our model comprehensively understands both the immediate sequences of events and the broader dynamics between activities.

We enhanced our architecture with an additional bi-directional LSTM layer, drawing methodological inspiration from the strategies proposed in \cite{devanne2019recognition}. This enhancement allows our model to not only discern immediate sequences of events but also to contextualize activities within a broader temporal framework. The bi-directionality of this LSTM layer enables it to capture insights from both prior and subsequent events, effectively integrating immediate event transitions with overarching behavioral patterns, thus providing a comprehensive understanding of human activity sequences.

For the input, our model processes a chronologically ordered sequence of three activities. This architectural choice is strategically made to predict the label of the current activity by leveraging the contextual representations of its two preceding activities. We opted for a sequence of three activities  ($c=3$) based on observations that activities often intersperse with a category termed "other," representing unlabeled sequences of sensor activations.

\subsection{Time Encoding}

In smart home human activity recognition, human behaviors often exhibit rhythmic patterns driven by ingrained habits. For example, sensor activation in the kitchen at 8 am might indicate breakfast, whereas the same trigger at 8 pm could relate to dinner.

We incorporate a specialized temporal encoding $E_{\text{timestamp}}$, as depicted in Fig. \ref{fig:complete_arch}. A supplementary input maps to the hours corresponding to each sensor activation timestamp. For every sensor activation in the main input sequence, a corresponding hour-of-the-day value is aligned in this secondary time input sequence. These hour values are vectorized through an embedding layer, which are then processed by a bi-directional LSTM. The output from this LSTM combines with the output from the bi-directional LSTM that encodes the sensor activations, directed to a terminal LSTM layer designed to discern the intricate relationships and order of activities.

\section{Methods}

\subsection{Datasets}

To evaluate the robustness and adaptability of our model, we utilized from the CASAS collection \cite{cook2012casas} three datasets : Aruba to test single-resident scenarios, Milan to assess the impact of pets and sensor issues, and Cairo to observe performance in multi-resident settings with overlapping activities. Collected from volunteers' homes over several months, these datasets feature unbalanced classes and vary in house structures and resident numbers\footnote{Additional details are provided in Annex \ref{sec:dataset_details} of the supplementary material}.

\subsection{Generative Pre-trained Transformer Sensor Embedding}

Following the methodology outlined by \cite{bouchabou2021using}, we employ a GPT transformer decoder \cite{radford2018improving} for training our sensor embedding. This model predicts the next sensor event based on the current context, employing natural language processing techniques to comprehend the logical sequence of human actions through sensor activations. As depicted in Figure \ref{fig:proposed_architecture}, our GPT Sensor Embedding model includes token and positional embeddings, along with transformer decoders featuring pre-normalization \cite{xiong2020layer}, aligning with the GPT-2 \cite{radford2019language} architecture. We utilized a context length of 1024 tokens for input contexts, consistent with the GPT-2 model's configuration.

\subsection{Pre-processing, Training, Evaluation and Metrics}

Datasets are divided weekly to preserve temporal relationships. Once partitioned, the weeks are shuffled and split into training (70\%) and testing (30\%) subsets. The training subset serves multiple purposes, including training of pre-trained embeddings, hyperparameter optimization through cross-validation, and training of the classifier.

During the embedding pre-training step, 80\% of the training subset is used for model training, with the remaining 20\% for validation. Early stopping, based on validation perplexity, is employed to prevent overfitting. 
For hyperparameter tuning, we utilize a 3-fold cross-validation method. From the first two folds, 20\% is reserved for validation and early stopping, with the third fold serving as the test set for this phase.  
In the final classification stage, 20\% of the whole training subset is earmarked for validation and early stopping, followed by testing on the initially defined test set.

Given the significant imbalance in our activity classes, we primarily report the F1-score metric to provide an unbiased estimate of performance. For statistical robustness, results are averaged over 10 repetitions, and all experiments are conducted with a fixed value for random seeds to ensure reproducibility.

\subsection{Vectorization of Sensor Activations}

Sensor activations are categorized as distinct symbols, facilitating the model's ability to identify patterns and relationships among activations, forming a comprehensive vocabulary of sensor actions. Our datasets include readings from motion (M), door (D), and temperature (T) sensors, each event logged with a unique sensor ID, corresponding value, and timestamp. 
Events are converted into unique tokens by merging sensor ID (\(i_i\)) and value (\(v_i\)), while the timestamp (\(t_i\)) is excluded. For example, a motion sensor M001 turning ON is tokenized as 'M001ON', and a temperature reading from sensor T004 at \(24.5^{\circ}C\) is represented as 'T00424.5'.

These tokens are indexed using natural language processing techniques, where indexing starts at 1, with 0 reserved for padding. The frequency of occurrence determines the index of each token, ensuring frequently appearing tokens are indexed with lower numbers. Thus, a sensor activation sequence like {[M005OFF, M007OFF, M004OFF, M004ON]} is transformed into an indexed sequence like {[1, 4, 8, 2]}, reflecting the prevalence of each token.

\section{Results}

We address these research questions through empirical studies:
\begin{enumerate}[label=RQ\arabic*]
    \item Can GPT-based embeddings better capture long-term dependencies and improve activity recognition accuracy?
    \item Is a single-level long-term dependency model sufficient for modeling activities of daily living, and what additional benefits does a hierarchical model provide?
    \item Is temporal information relevant for HAR, especially given the irregular sampling of time-series data from event-triggered sensors?
\end{enumerate}

In this section we provide an empirical study of the following research questions: 
(RQ1) Can GPT-based models capture better long-term dependencies and lead to better activity recognition ? 
(RQ2) Is a single-level long-term dependency enough to model activities of daily living and what does a hierarchical model add ?
(RQ3) Is time a relevant information for HAR, especially in the case of this irregularly sampled time-series from event-triggered sensors ?

To investigate these questions, we conducted ablation studies with: 
\begin{itemize}
    \item GPTHAR, our proposed method, employs a GPT transformer decoder with time-encoding and a hierarchical architecture as depicted in Fig. \ref{fig:complete_arch}.
    \item GPTHAR-note (no temporal encoding) uses a GPT transformer decoder in a hierarchical architecture (no timestamp information).
    \item GPTAR (see Fig. \ref{fig:proposed_architecture}) uses a GPT transformer decoder as embedding in a single-level architecture (no timestamp information).
    \item ELMoHAR, which uses ELMo as an embedding, combined with time-encoding and a hierarchical architecture.
    \item ELMoHAR-note (no temporal encoding) uses ELMo and a hierarchical architecture without timestamp information.
    \item ELMoAR, which applies ELMo for embedding in a single-level architecture without timestamp information.
\end{itemize}

\subsection{Hyperparameters Search and Comparative Study}

To address RQ1, we compared the embeddings: our GPT decoder Transformer-based model (GPTAR) and the ELMo-based model (ELMoAR) as described in \cite{bouchabou2021using}. Both models adhere to the architecture shown in Fig. \ref{fig:proposed_architecture}, differing in the embedding. Each embedding was pre-trained using sensor activations, with the GPT decoder applied to unsegmented data and ELMo to pre-segmented data.

At first, we conducted hyperparameter tuning for each embedding technique, adjusting the context window size for ELMoAR and the number of layers and attention heads for GPTAR\footnote{The results of a 3-fold cross-validation on three datasets are reported in Table \ref{tab:cross_validation_results_parameters_search_f1} in Annex \ref{sec:cross_validation_results_parameters_search_f1} of the supplementary material}. The ELMoAR model, configured with a 60-token context window, provided the best results across datasets for ELMo-based models. For GPTAR, a configuration with 8 attention heads and 3 decoder layers yielded the best average F1 scores, especially in the noisier Milan and Cairo datasets.

\begin{table}[bt]
\vspace{5pt}
\centering
\caption{Hierarchical Model: Test Mean F1-scores for FCN, Liciotti et al., and both hierarchical and non-hierarchical architectures using ELMo and GPT  Embeddings. The ELMoAR and ELMoHAR models utilized a 60-token context window, while the GPTAR  and GPTHAR model was configured with 8 attention heads and 3 layers.}
\label{tab:hierarchical_comparison_f1}
\resizebox{\columnwidth}{!}{%
\begin{tabular}{lcccccl}
\toprule
                         & \multicolumn{2}{c}{Aruba}        & \multicolumn{2}{c}{Milan}        & \multicolumn{2}{c}{Cairo}                   \\
                         & F1 Score   & std           & F1 Score   & std           & F1 Score   & std                      \\ \hline
\begin{tabular}[c]{@{}l@{}}FCN \cite{bouchabou2021fully} \end{tabular}                      & 33.10\%          & 2.23          & 15.10            & 1.52          & 7.60\%           & \multicolumn{1}{c}{2.46} \\
\begin{tabular}[c]{@{}l@{}}Bi-LSTM \cite{liciotti_lstm} \end{tabular}          & 32.00\%          & 1.56          & 17.40\%          & 2.07          & 26.60\%          & \multicolumn{1}{c}{3.24} \\
\begin{tabular}[c]{@{}l@{}}ELMoAR (W-size 60)   \end{tabular}      & 84.80\%          & 1.99          & 70.80\%          & \textbf{0.92} & 70.50\%          & \textbf{1.43}            \\
\begin{tabular}[c]{@{}l@{}}GPTAR (8H 3L) \end{tabular}  & 86.10\%          & \textbf{1.20} & 70.80\%          & 1.40          & 73.20\%          & 1.99                     \\
ELMoHAR-note                  & \textbf{88.10\%} & \textbf{1.20} & 77.40            & 1.65          & 75.90\%          & 2.88                     \\
GPTHAR-note                   & 87.30\%          & 2.98          & \textbf{79.90\%} & 1.52          & \textbf{84.80\%} & 1.81\\           \bottomrule       
\end{tabular}%
}
\end{table}

Subsequently, using the selected embedding parameters, we conducted the final training and classification tasks for the ELMoAR and GPTAR architectures. 
In addition, we evaluated their performance against other methodological approaches: the Fully Convolutional Network (FCN), which has shown effective performance in time series classification \cite{fawaz2019deep} and ADL recognition \cite{bouchabou2021fully}, and a SOTA LSTM-based method known for its strong results in ADL recognition \cite{liciotti_lstm} \footnote{For details of the LSTM-based method, please refer to Annex \ref{sec:liciotti_rep}.} The F1 scores are reported in Table \ref{tab:hierarchical_comparison_f1}\footnote{additional metrics and details are provided in Annex \ref{sec:resultsDetailedScores} of the supplementary material}. 
The results demonstrate that both ELMoAR and GPTAR surpass the LSTM-based approach and the FCN in performance. Notably, while the LSTM method excels during training, it exhibits a significant drop in performance during testing. FCN struggles to effectively manage the length variability inherent in the sequences. In particular, the results indicate that GPTAR outperformed ELMoAR in the Aruba and Cairo datasets, although both models achieved the same F1 score on the Milan dataset.

\begin{table}[th]
\centering
\caption{F1-Score by activity for each algorithm on the Aruba dataset}
\label{tab:f1-score-aruba}
\resizebox{\columnwidth}{!}{%
\begin{tabular}{lllllll}
\toprule
                           & \textbf{ELMoAR} & \textbf{GPTAR} & \textbf{ELMoHAR-note} & \textbf{GPTHAR-note} & \textbf{ELMoHAR} & \textbf{GPTHAR} \\ \hline
\textbf{Bed\_to\_Toilet}   & 0,987           & 0,996          & 0,991            & 0,992           & 0,992                   & \bf{0,996}                  \\
\textbf{Eating}            & 0,925           & 0,935          & 0,937            & \bf{0,948}           & 0,932                   & 0,936                  \\
\textbf{Enter\_Home}       & 0,8             & 0,797          & 0,992            & \bf{0,994}           & 0,99                    & 0,992                  \\
\textbf{Housekeeping}      & 0,83            & 0,84           & 0,829            & 0,907           & 0,852                   & \bf{0,918}                  \\
\textbf{Leave\_Home}       & 0,827           & 0,811          & \bf{0,992}            & \bf{0,992}           & 0,99                    & 0,99                   \\
\textbf{Meal\_Preparation} & \bf{0,974}           & 0,971          & 0,972            & 0,968           & 0,971                   & 0,964                  \\
\textbf{Other}             & 0,988           & 0,99           & 0,99             & 0,99            & 0,99                    & 0,99                   \\
\textbf{Relax}             & 0,99            & 0,99           & 0,991            & 0,993           & \bf{0,994}                   & 0,992                  \\
\textbf{Respirate}         & 0,867           & \bf{0,967}          & 0,651            & 0,502           & \bf{0,967}                   & 0,548                  \\
\textbf{Sleeping}          & 0,988           & 0,99           & 0,982            & 0,984           & \bf{0,991}                   & 0,981                  \\
\textbf{Wash\_Dishes}      & 0,008           & 0,047          & \bf{0,266}            & 0,243           & 0,225                   & 0,199                  \\
\textbf{Work}              & 0,984           & \bf{0,993}          & 0,979            & 0,987           & 0,985                   & 0,975\\
\bottomrule
\end{tabular}%
}
\end{table}
\begin{table}[th]
\centering
\caption{F1-Score by activity for each algorithm on Milan dataset}
\label{tab:f1-score-milan}
\resizebox{\columnwidth}{!}{%
\begin{tabular}{lllllll}
\toprule
                                   & \textbf{ELMoAR} & \textbf{GPTAR} & \textbf{ELMoHAR-note} & \textbf{GPTHAR-note} & \textbf{ELMoHAR} & \textbf{GPTHAR} \\ \hline
\textbf{Bed\_to\_Toilet}           & 0,551           & 0,532          & 0,79             & 0,749           & 0,845                   & \bf{0,902}                  \\
\textbf{Chores}                    & 0               & 0,011          & 0,088            & 0,122           & 0,15                    & \bf{0,161}                  \\
\textbf{Desk\_Activity}            & 0,976           & \bf{0,996}          & 0,952            & 0,982           & 0,958                   & 0,976                  \\
\textbf{Dining\_Rm\_Activity}      & 0,416           & 0,252          & 0,516            & 0,481           & 0,491                   & \bf{0,522}                  \\
\textbf{Eve\_Meds}                 & 0               & 0,095          & 0,421            & \bf{0,587}           & 0,566                   & 0,545                  \\
\textbf{Guest\_Bathroom}           & 0,978           & 0,981          & 0,979            & 0,98            & 0,98                    & \bf{0,984}                  \\
\textbf{Kitchen\_Activity}         & 0,91            & 0,919          & 0,93             & \bf{0,936}           & 0,927                   & 0,931                  \\
\textbf{Leave\_Home}               & 0,901           & 0,911          & 0,934            & 0,952           & 0,941                   &\bf{0,957}                  \\
\textbf{Master\_Bathroom}          & 0,884           & 0,858          & 0,942            & 0,943           & 0,968                   & \bf{0,977}                  \\
\textbf{Master\_Bedroom\_Activity} & 0,788           & 0,792          & 0,799            & 0,836           & 0,836                   & \bf{0,872}                  \\
\textbf{Meditate}                  & 0,86            & 0,874          & 0,863            & 0,933           & 0,866                   & \bf{0,972}                  \\
\textbf{Morning\_Meds}             & 0,58            & 0,546          & 0,642            & 0,669           & 0,698                   & \bf{0,712}                  \\
\textbf{Other}                     & 0,888           & 0,9            & 0,906            & 0,92            & 0,909                   & \bf{0,924}                  \\
\textbf{Read}                      & 0,92            & 0,946          & 0,917            & 0,947           & 0,914                   & \bf{0,953}                  \\
\textbf{Sleep}                     & 0,897           & 0,921          & 0,927            & 0,925           & \bf{0,943}                   & 0,932                  \\
\textbf{Watch\_TV}                 & 0,767           & 0,778          & 0,782            & \bf{0,807}           & 0,789                   & 0,806\\
\bottomrule
\end{tabular}%
}
\end{table}
\begin{table}[!h]
\centering
\caption{F1-Score by activity for each algorithm on Cairo dataset}
\label{tab:f1-score-cairo}
\resizebox{\columnwidth}{!}{%
\begin{tabular}{lllllll}
\toprule
                              & \textbf{ELMoAR} & \textbf{GPTAR} & \textbf{ELMoHAR-note} & \textbf{GPTHAR-note} & \textbf{ELMoHAR} & \textbf{GPTHAR} \\ \hline
\textbf{Bed\_to\_toilet}      & 0,382           & 0,359          & 0,374            & 0,483           & \bf{0,545}                   & 0,311                  \\
\textbf{Breakfast}            & 0,533           & 0,61           & 0,822            & 0,89            & 0,886                   & \bf{0,94}                   \\
\textbf{Dinner}               & 0,415           & 0,412          & 0,711            & 0,713           & 0,961                   & \bf{0,991}                  \\
\textbf{Laundry}              & 0,911           & 1              & 0,309            & \bf{0,951}           & 0,358                   & 0,794                  \\
\textbf{Leave\_home}          & 0,938           & 0,958          & 0,891            & 0,97            & 0,895                   & \bf{0,973}                  \\
\textbf{Lunch}                & 0,337           & 0,331          & 0,609            & 0,67            & 0,912                   & \bf{0,929}                  \\
\textbf{Night\_wandering}     & 0,759           & 0,795          & 0,805            & 0,829           & 0,823                   & \bf{0,837}                  \\
\textbf{Other}                & 0,915           & 0,923          & 0,948            & 0,968           & 0,955                   & \bf{0,977}                  \\
\textbf{R1\_sleep}            & 0,702           & 0,629          & 0,852            & 0,868           & \bf{0,897}                   & 0,877                  \\
\textbf{R1\_wake}             & 0,871           & 0,899          & 0,903            & 0,906           & 0,895                   & \bf{0,909}                  \\
\textbf{R1\_work\_in\_office} & 0,853           & 0,939          & 0,918            & 0,985           & 0,92                    & \bf{0,991}                  \\
\textbf{R2\_sleep}            & 0,706           & 0,715          & 0,824            & 0,857           & 0,856                   & \bf{0,863}                  \\
\textbf{R2\_take\_medicine}   & 0,781           & 0,913          & 0,859            & 0,94            & 0,87                    & \bf{0,931}                  \\
\textbf{R2\_wake}             & 0,745           & 0,746          & 0,799            & 0,821           & 0,852                   & \bf{0,87  }\\
\bottomrule
\end{tabular}%
}
\end{table}

\begin{figure*}[hbt]
     \begin{subfigure}{0.33\textwidth}
         \centering
         \includegraphics[width=\textwidth]{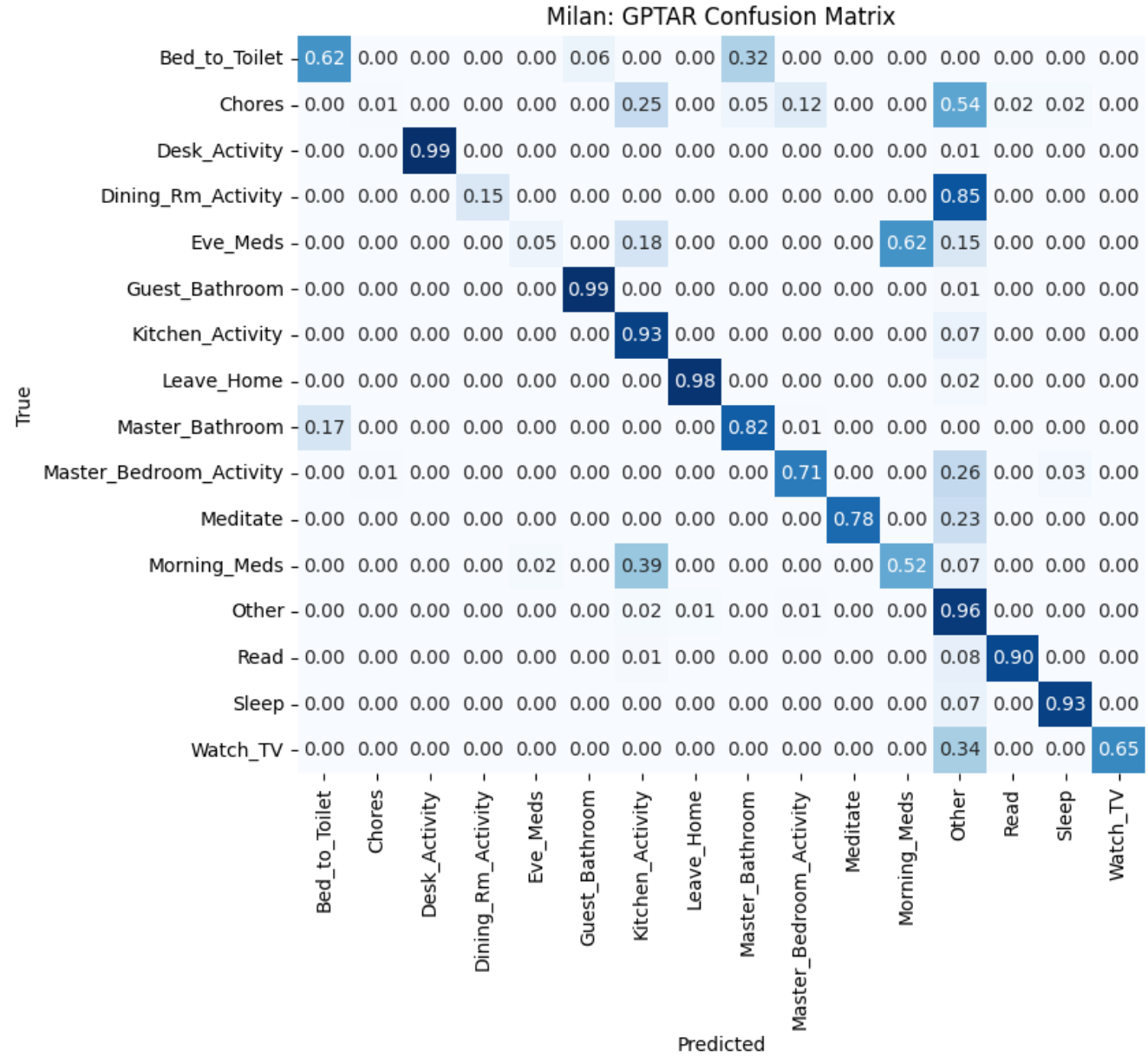}
         \caption{GPTAR}
         \label{fig:Cairo_GPTAR}
     \end{subfigure}
     \begin{subfigure}{0.32\textwidth}
         \centering
         \vspace{-35pt}
         \includegraphics[width=\textwidth]{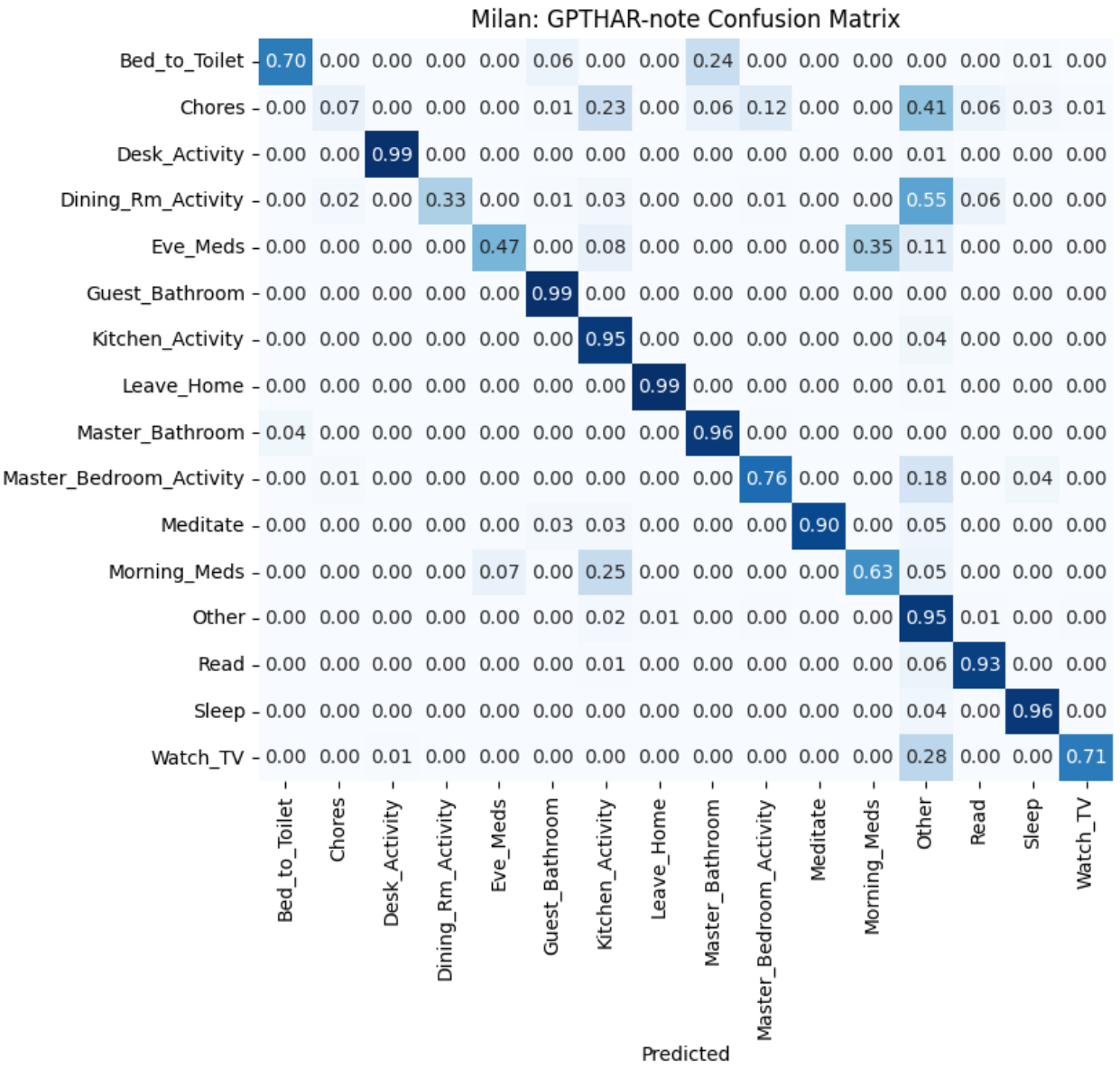}
         \vspace{-50pt}
         \caption{GPTHAR-note}
         \label{fig:Cairo_GPTHAR_N}
     \end{subfigure}
     \begin{subfigure}{0.35\textwidth}
         \centering
         \includegraphics[width=\textwidth]{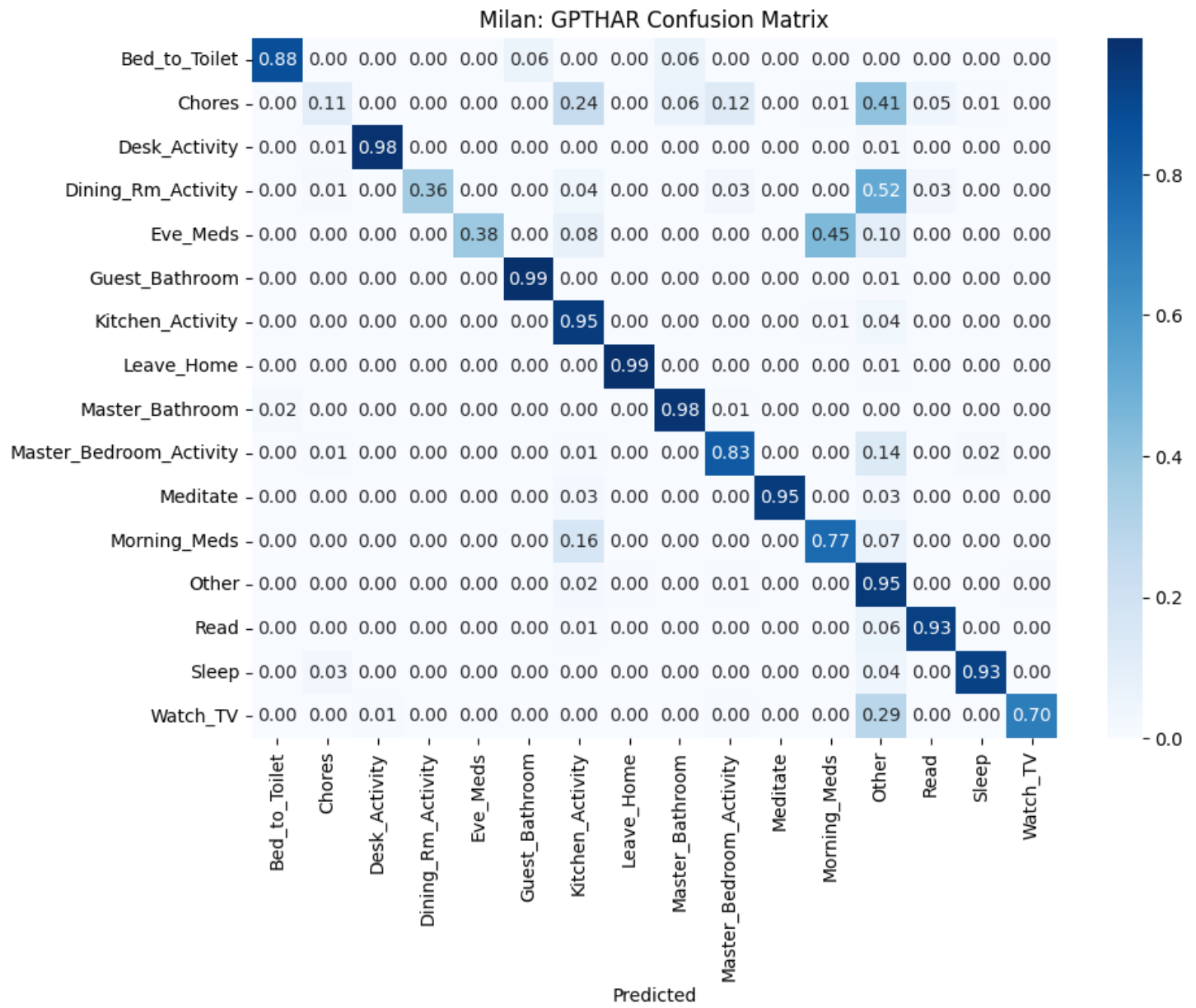}
         \caption{GPTHAR}
         \label{fig:Cairo_GPTHAR}
     \end{subfigure}
          \hfill
    \vspace{5pt}
    \caption{Confusion matrices per algorithm on the  Milan dataset}
    \label{fig:matrix_Milan_main}
    \vspace{15pt}
\end{figure*}

\begin{figure*}[hbt]

     \begin{subfigure}{0.33\textwidth}
         \centering
         \vspace{-40pt}
         \includegraphics[width=\textwidth]{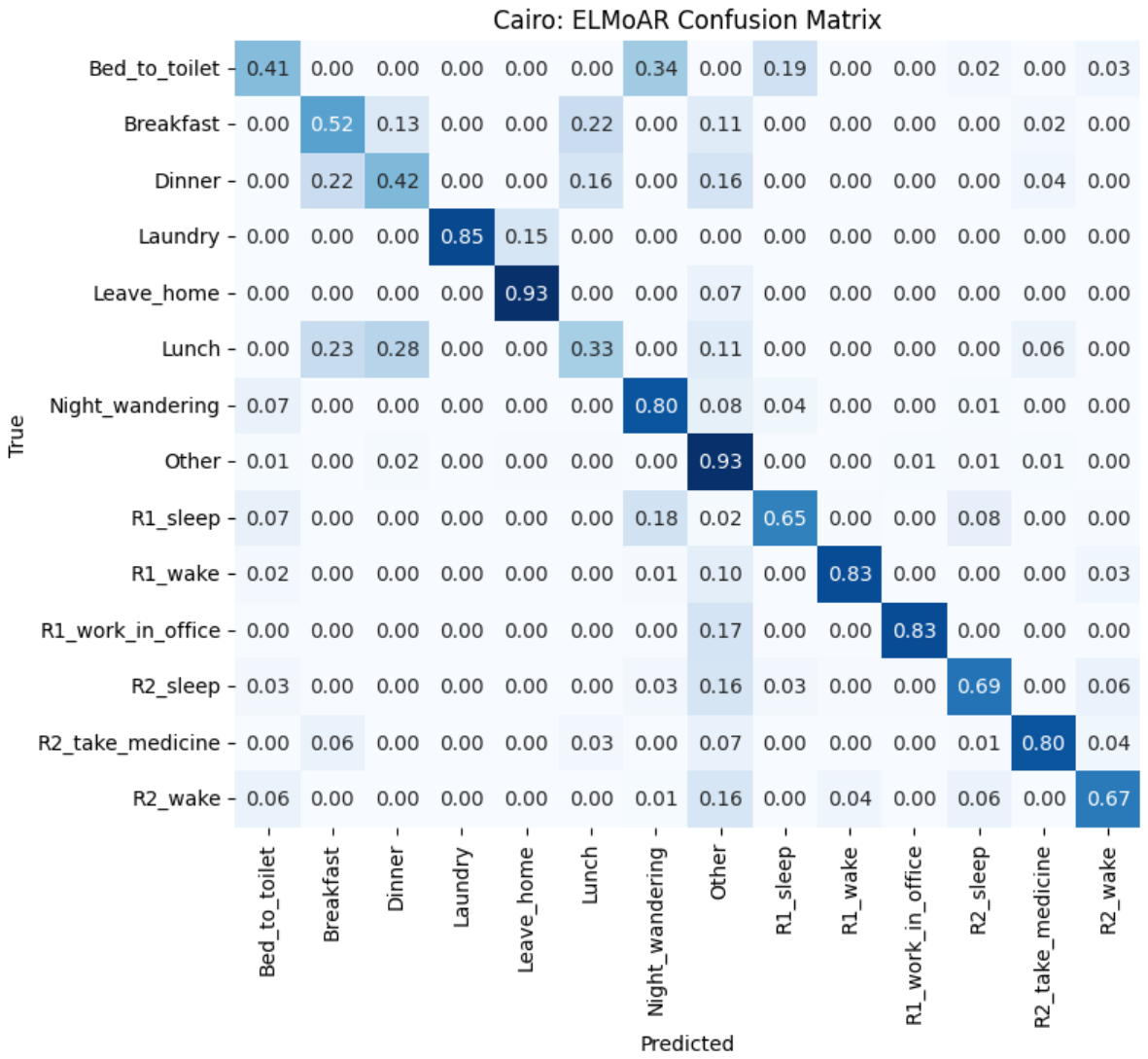}
         \vspace{-60pt}
         \caption{ELMoAR}
         \label{fig:Cairo_ELMoAR}
     \end{subfigure}
     \begin{subfigure}{0.32\textwidth}
         \centering
         \vspace{-40pt}
         \includegraphics[width=\textwidth]{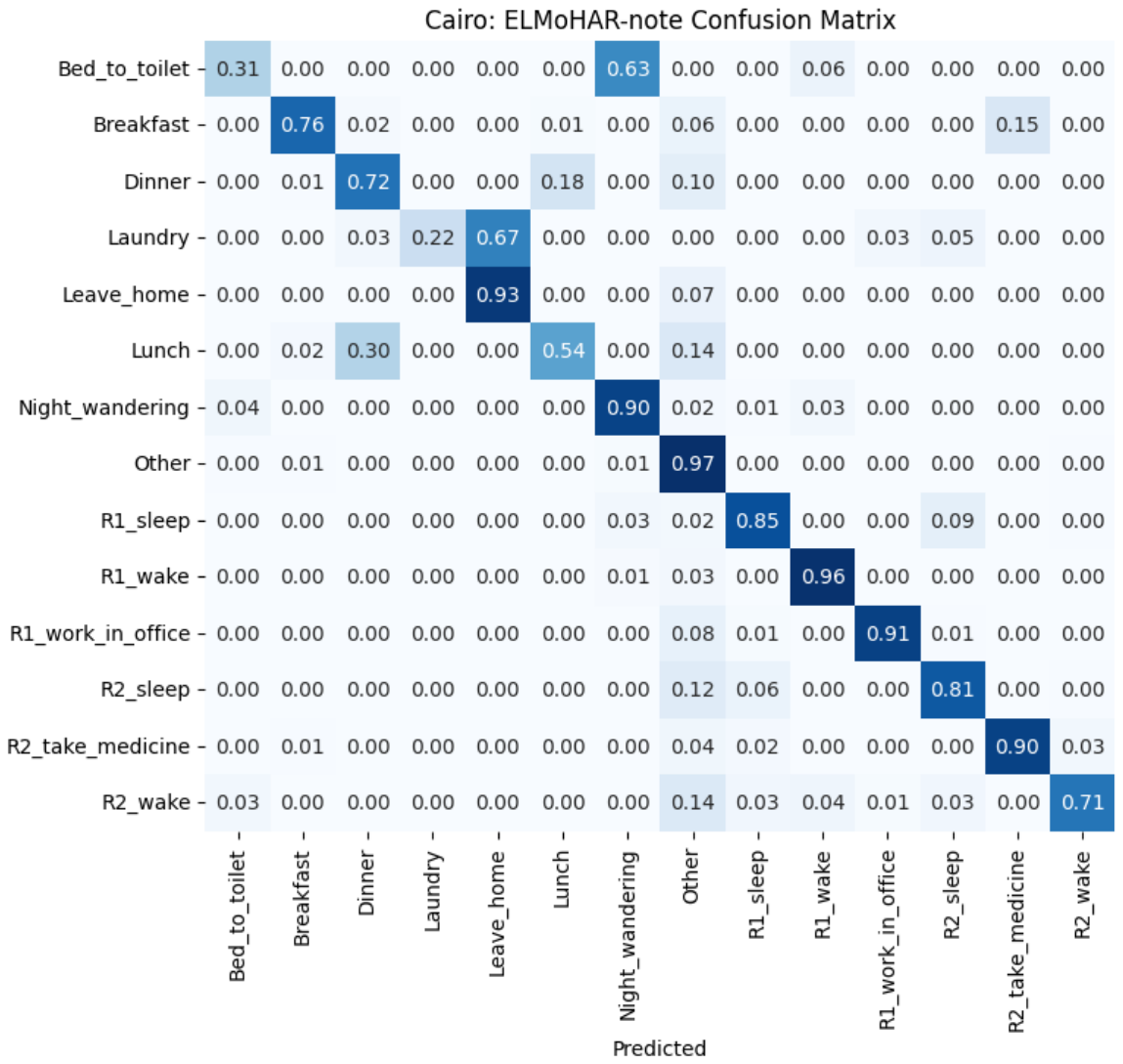}
         \vspace{-60pt}
         \caption{ELMoHAR-note}
         \label{fig:Cairo_ELMoHAR_N}
     \end{subfigure}
     \begin{subfigure}{0.35\textwidth}
         \centering
         \includegraphics[width=\textwidth]{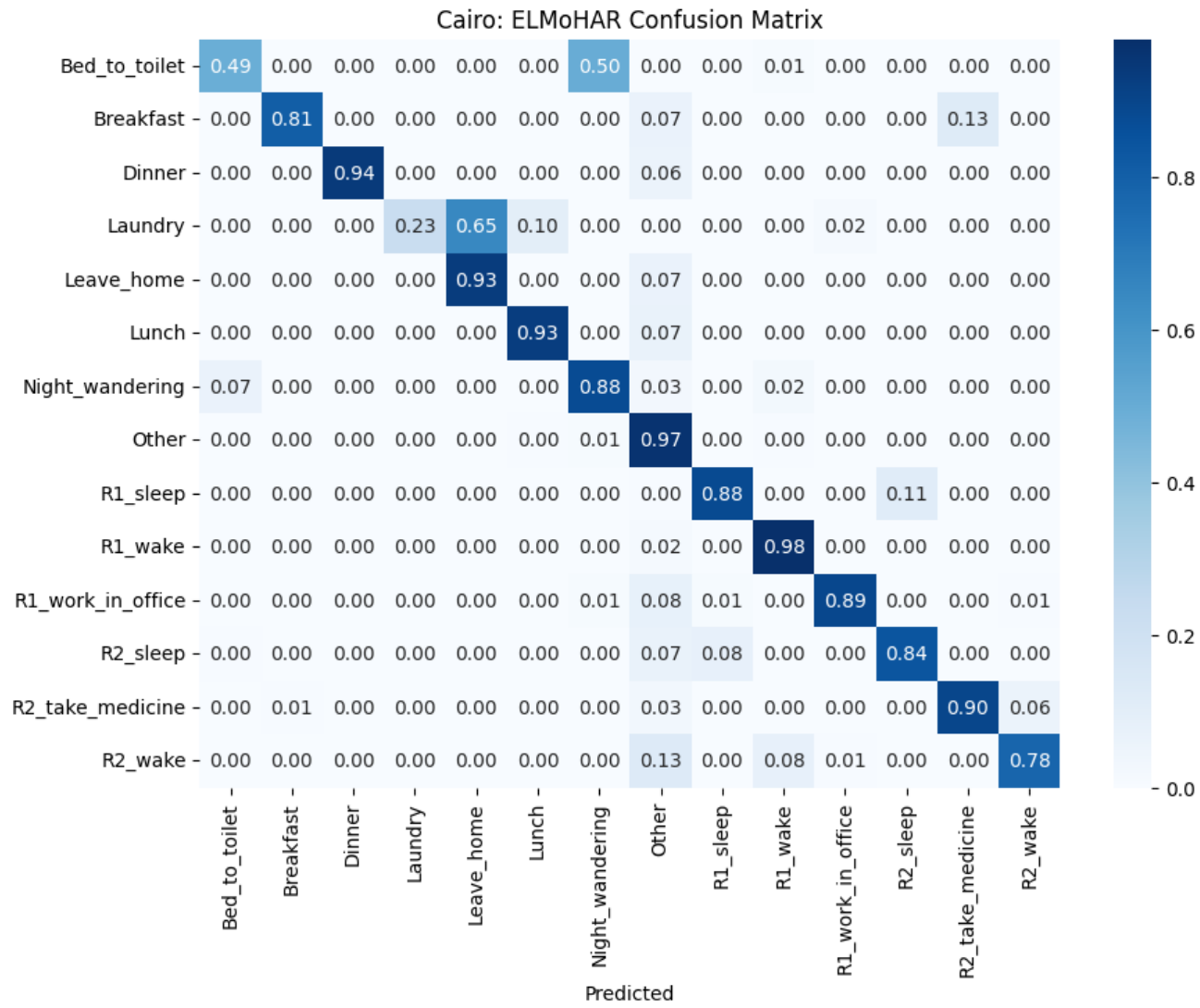}
         \caption{ELMoHAR}
         \label{fig:Cairo_ELMoHAR}
     \end{subfigure}
          
     \begin{subfigure}{0.33\textwidth}
         \centering
         \vspace{-40pt}
         \includegraphics[width=\textwidth]{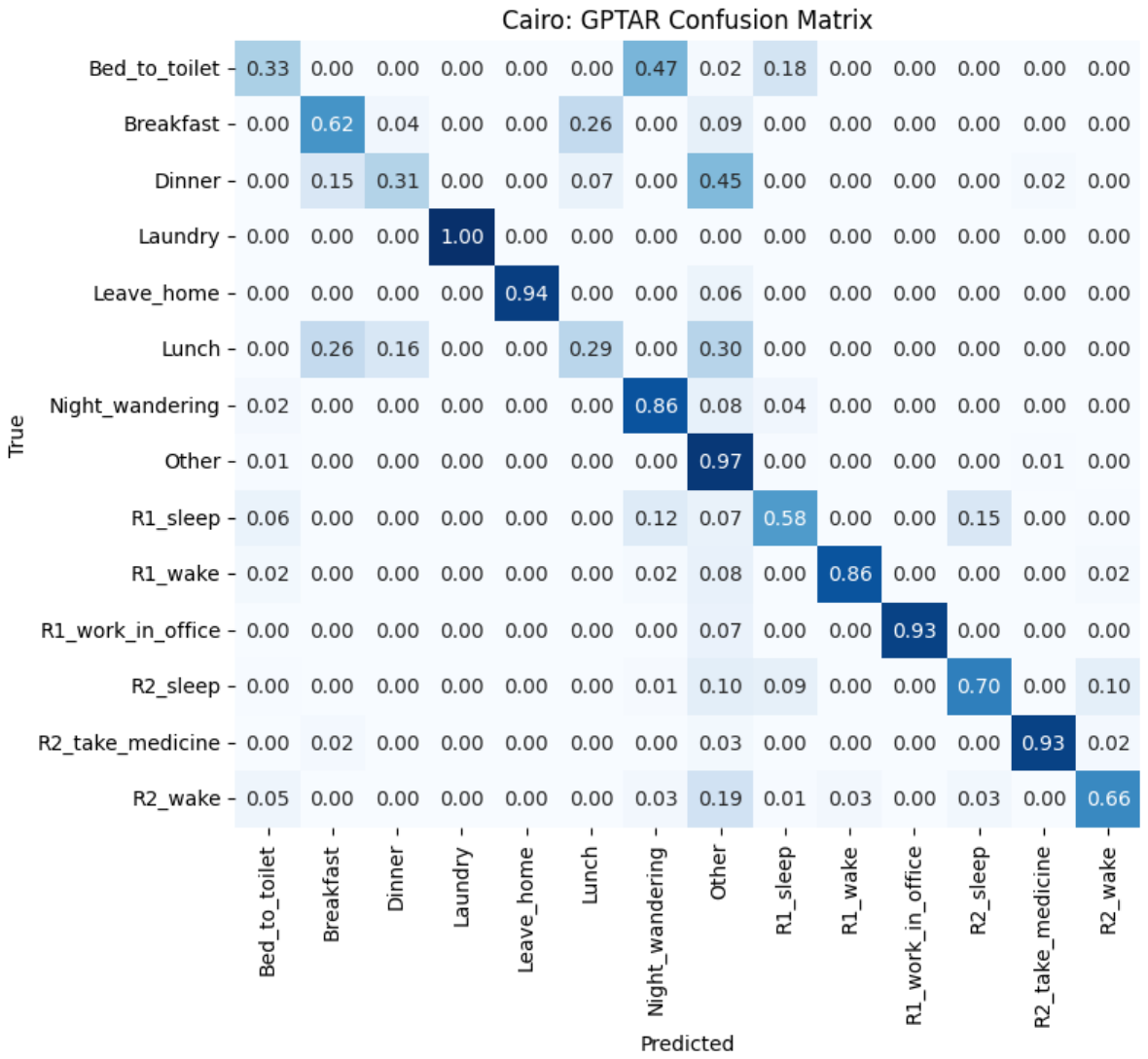}
         \vspace{-60pt}
         \caption{GPTAR}
         \label{fig:Cairo_GPTAR}
     \end{subfigure}
     \begin{subfigure}{0.32\textwidth}
         \centering
         \vspace{-40pt}
         \includegraphics[width=\textwidth]{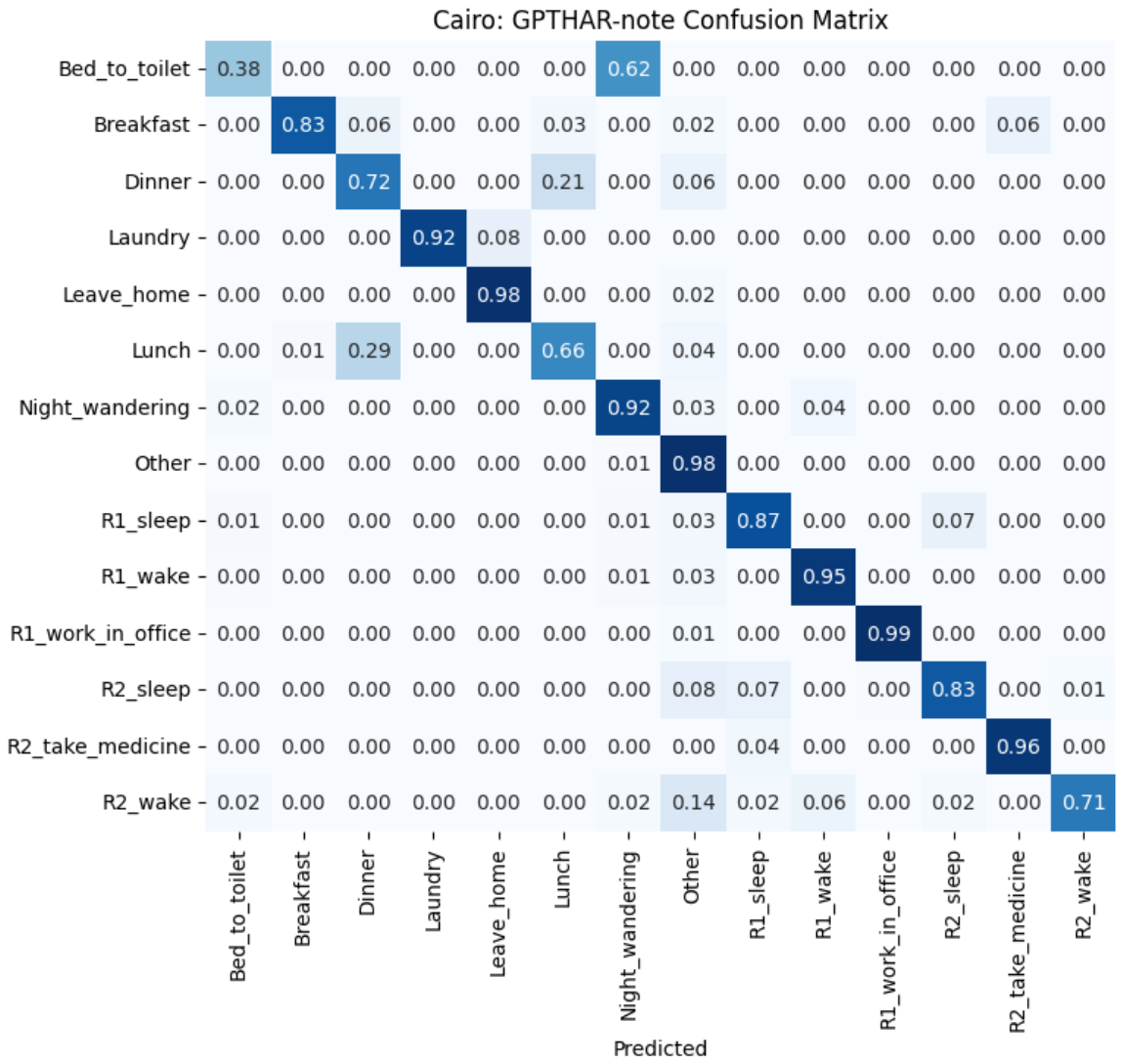}
         \vspace{-60pt}
         \caption{GPTHAR-note}
         \label{fig:Cairo_GPTHAR_N}
     \end{subfigure}
     \begin{subfigure}{0.35\textwidth}
         \centering
         \includegraphics[width=\textwidth]{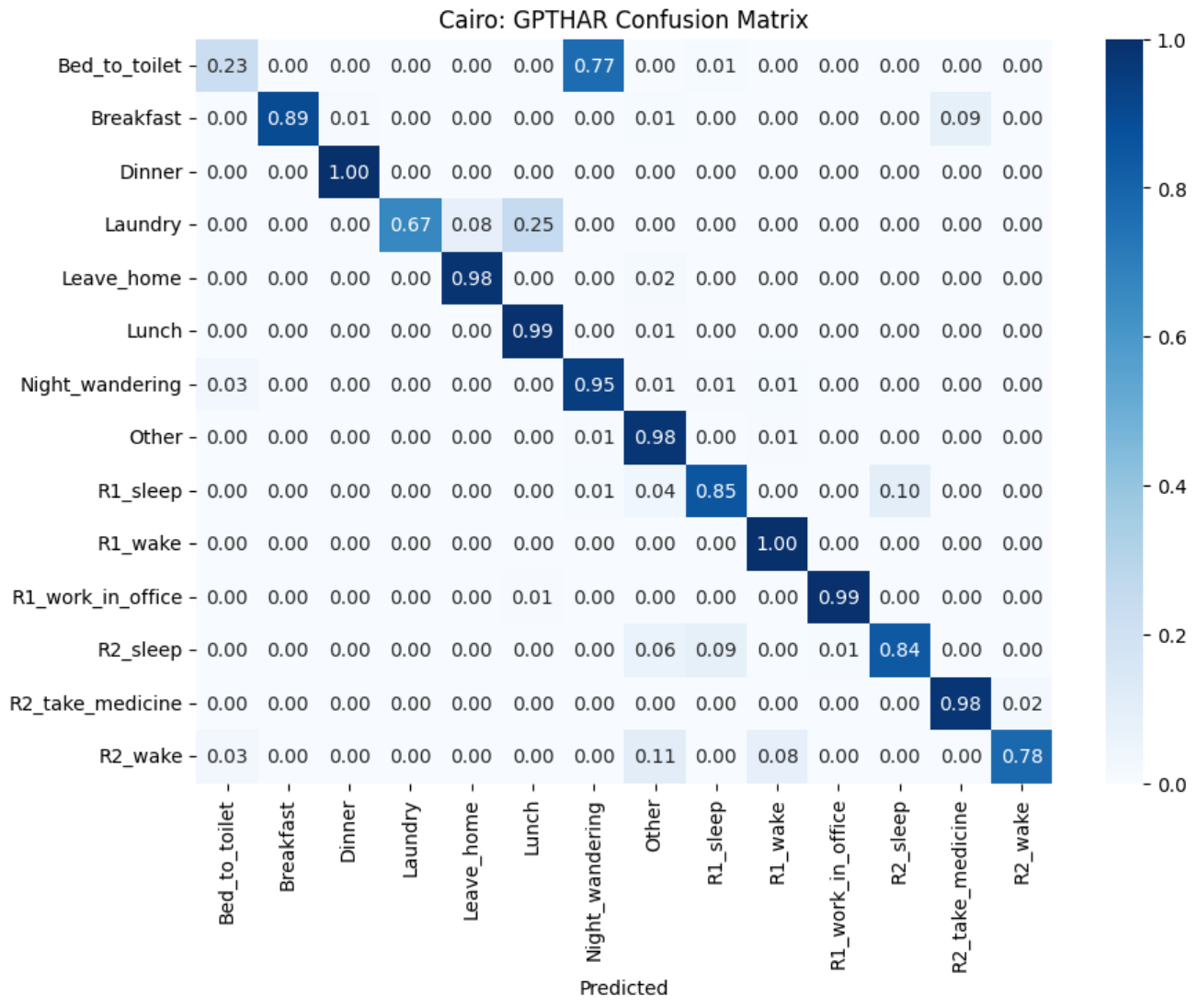}
         \caption{GPTHAR}
         \label{fig:Cairo_GPTHAR}
     \end{subfigure}
          \hfill
    \vspace{10pt}
    \caption{Confusion matrices per algorithm on the  Cairo dataset}
    \label{fig:matrix_Cairo_main}
    \vspace{10pt}
\end{figure*}

\begin{figure}[bht]
     \begin{subfigure}{0.22\textwidth}
         \includegraphics[width=\textwidth]{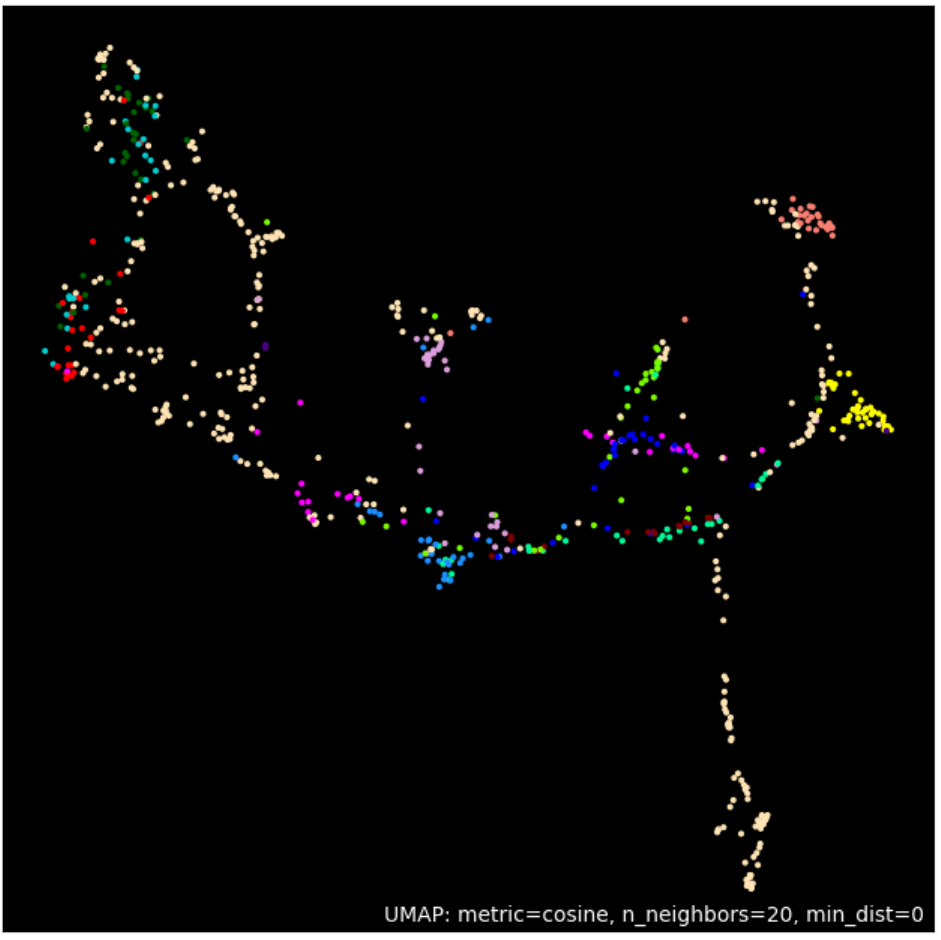}
         \caption{ELMo}
         \label{fig:Cairo_ELmo_Emb}
     \end{subfigure}
     \hfill
     \begin{subfigure}{0.22\textwidth}
         \includegraphics[width=\textwidth]{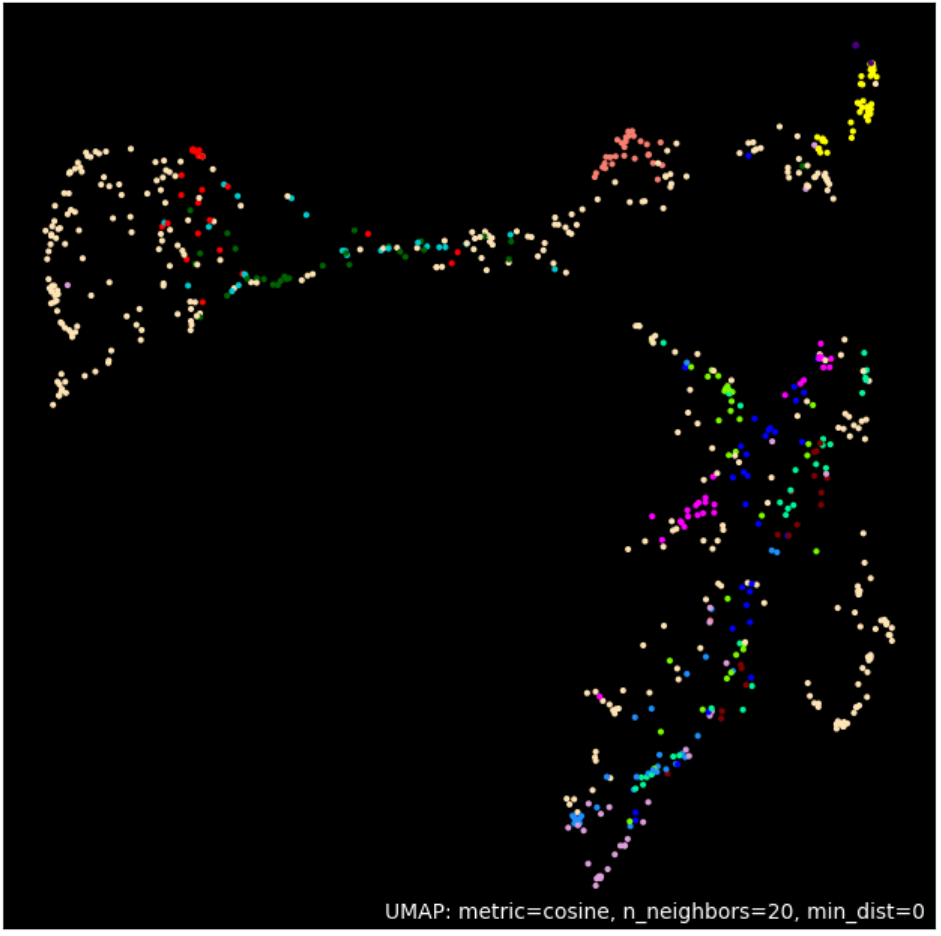}
         \caption{GPT}
         \label{fig:Cairo_GPT_Emb}
     \end{subfigure}
     \begin{subfigure}{0.033\textwidth}
         \includegraphics[width=\textwidth]{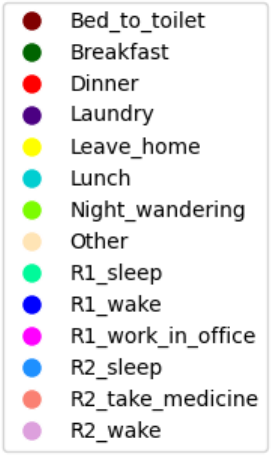}
     \end{subfigure}
    \vspace{7pt}
    \caption{Visualization of Activity Embeddings for the Cairo Dataset Generated by ELMo and GPT }
    \label{fig:emb_cairo_m}
    \vspace{15pt}
\end{figure}

However, an analysis of individual activity performance with F1-scores per activity in Tables \ref{tab:f1-score-aruba},\ref{tab:f1-score-milan} \& \ref{tab:f1-score-cairo} and the confusion matrices in Fig.\ref{fig:matrix_Milan_main} \& \ref{fig:matrix_Cairo_main}  revealed specific challenges\footnote{see all confusion matrices in Annex \ref{sec:cm}}. Activities such as `Wash Dishes' in Aruba, `Eve Meds' in Milan, and `Laundry' in Cairo showed better recognition by GPTAR. These activities exhibit similar sensor patterns to other activities, such as between `Meal Preparation' and `Wash Dishes' in Aruba.  `Eve Meds' 's confusion with `Kitchen Activity' by ELMo (see green dots top right corner) is partially resolved by GPT .
Fig. \ref{fig:emb_cairo_m} shows that  the `Laundry' data in Cairo are embedded closer together by GPT than ELMo.

\subsection{Hierarchical Activity Recognition}

To investigate RQ2, we compare GPTAR and ELMoAR with the hierarchical models versions without time encoding, GPTHAR-note and ELMoHAR-note respectively. These models omit the timestamp inputs depicted in Fig. \ref{fig:complete_arch}. Our analysis in Sections \ref{sec:resultsHierarchical} and \ref{sec:resultsExtend} evaluates whether hierarchical models, which aim to capture complex inter-activity relationships, outperform non-hierarchical models that utilize longer input contexts and activity segmentation markers.

\subsubsection{Comparison of Hierarchical Architectures}
\label{sec:resultsHierarchical}

As with previous experiments, the pre-trained embeddings were maintained frozen, with only the dual bi-directional LSTM layers and the softmax layer undergoing further training for classification.

Table \ref{tab:hierarchical_comparison_f1} illustrates that the hierarchical structure effectively leverages both types of pre-trained embeddings to enhance performance.   While GPTHAR-note shows substantial improvements over ELMoHAR-note in the Milan and especially Cairo datasets, it slightly lags behind in the Aruba dataset, suggesting that GPTHAR-note performs better in more complex, noisy environments. 
However, employing a hierarchical structure has led to increased standard deviation, indicating a reduction in performance consistency. We hypothesize that a regularization layer following the embedding output could stabilize the model's performance.

Comparison of GPTHAR-note to GPTAR, and ELMoHAR-note to ELMoAR for each activity F1 scores in Tables \ref{tab:f1-score-aruba}, \ref{tab:f1-score-milan} \& \ref{tab:f1-score-cairo} and confusion matrices\footnote{see complete results in Annex \ref{sec:cm}}   in Fig. \ref{fig:matrix_Milan_main} \& \ref{fig:matrix_Cairo_main} reveals that sequential activities such as, in Milan, `Dining Room Activity',`Meditate' and `Watch TV'  are less confused with `Other' activity; and `Eve Med’, `Morning Meds' are less confused with `Kitchen activity'. Moreover, in Cairo, `Dinner', `Breakfast' and `Lunch' are less confused by the hierarchical models.

\subsubsection{Input Context Extended}
\label{sec:resultsExtend}

To assess the efficacy of the hierarchical structure, we conducted a comparative analysis between hierarchical and non-hierarchical model configurations using an augmented input context. Specifically, we extended the input for the base models, ELMoAR and GPTAR, by: 1) appending two preceding activities to the current one, while maintaining temporal sequences order, and 2) incorporating two preceding activities with a separation token, $<SEP>$ delineating distinct sensor sequences. This token is designed to clearly mark activity boundaries, a task inherently managed by the hierarchical models.

The F1-scores presented in Table \ref{tab:hierarchical_comparison_extended_f1}, confirm that hierarchical models consistently outperform their non-hierarchical counterparts with extended input contexts. Despite the addition of a separation token to non-hierarchical models to indicate sequence segmentation, leading to some improvements in F1-score, yet, hierarchical structures still demonstrated superior performance across all three datasets. 

Furthermore, our observations revealed that increasing the number of nodes in the last bi-directional LSTM layer of the non-hierarchical models improved performance for both the extended context and the extended context with separator setups. However, these enhancements were still insufficient to match the performance levels achieved by the hierarchical structures. 

\begin{table}[bth]
\centering
\caption{Long-term dependency : Test mean F1 Score and its standard deviation of the classification when using either (1) a hierarchical architecture, or (2) the  baseline models using extended input context, or (3) extended input context incorporating the "$<sep>$" token. For GPTAR we use 8 heads and 3 layers.}
\label{tab:hierarchical_comparison_extended_f1}
\resizebox{\columnwidth}{!}{%
\begin{tabular}{lcccccl}
\toprule
                                                                                              & \multicolumn{2}{c}{Aruba}                   & \multicolumn{2}{c}{Milan}        & \multicolumn{2}{c}{Cairo}                   \\
                                                                                              & F1 Score   & std                      & F1 Score   & std           & F1 Score   & std                      \\ \hline
ELMoHAR-note                                                                                        & \textbf{88.10\%} & \textbf{1.20}            & 77.40\%          & 1.65          & 75.90\%          & 2.88                     \\
GPTHAR-note                                                                                         & 87.30\%          & 2.98                     & \textbf{79.90\%} & 1.52          & \textbf{84.80\%} & \textbf{1.81}            \\
\begin{tabular}[c]{@{}l@{}}ELMoAR (Window 60) \\  entended\end{tabular}                & 76.90\%          & 1.2                      & 60.05\%          & 3.5           & 59.10\%          & \multicolumn{1}{c}{3.84} \\
\begin{tabular}[c]{@{}l@{}}ELMoAR (Window 60) \\  extended + $<sep>$\end{tabular}       & 77.90\%          & 3.63                     & 59.50\%          & 1.27          & 56.80\%          & \multicolumn{1}{c}{5.45} \\
\begin{tabular}[c]{@{}l@{}}GPTAR (8H 3L) \\ \ extended\end{tabular}          & 81.00\%          & \multicolumn{1}{l}{3.53} & 61.80\%          & 1.4           & 58.80\%          & 2.7                      \\
\begin{tabular}[c]{@{}l@{}}GPTAR (8H 3L) \\ \ extended + $<sep>$\end{tabular} & 81.60\%          & \multicolumn{1}{l}{3.81} & 61.40\%          & \textbf{1.26} & 60.20\%          & 2.94\\           
\bottomrule
\end{tabular}%
}
\end{table}

\subsection{The Impact of Time Encoding}
To address RQ3, 
 we compare models with and without this time encoding (ELMoHAR and GPTHAR with encoding; ELMoHAR-note and GPTHAR-note without) 
in Table \ref{tab:hierarchical_time_comparison_f1}\footnote{additional metrics in Annex \ref{sec:resultsDetailedScores}}.

\begin{figure*}
\centering
\includegraphics[width=\textwidth, trim=0cm 8.3cm 0cm 8.5cm, clip]{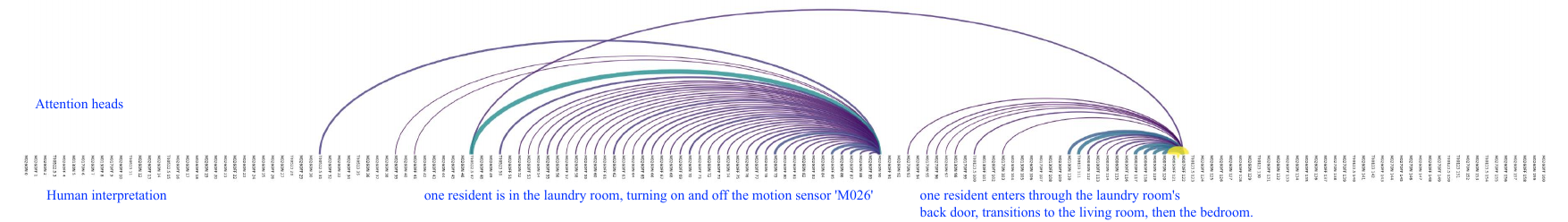} %
\caption{Visualization for GPT Transformer decoder on a "Laundry" sequence, of the attention heads leading to tokens 90 (M026ON) and 122 (M007OFF). 
The yellow and wide arcs indicate high attention; the blue and narrow arcs, a low attention. Time is read from left to right. }
\label{fig:case_study}
\vspace{8pt}
\end{figure*}

\begin{table}[bth]
\centering
\caption{Time encoding : Test mean F1 Score and its standard deviation comparing when the embedding is either (1) ELMoHAR, (2) GPTHAR with and without Time Encoding.}
\label{tab:hierarchical_time_comparison_f1}
\resizebox{\columnwidth}{!}{%
\begin{tabular}{lcccccl}
\toprule
                        & \multicolumn{2}{c}{Aruba}        & \multicolumn{2}{c}{Milan}       & \multicolumn{2}{c}{Cairo}      \\
                        & F1 Score   & std           & F1 Score   & std          & F1 Score & std           \\ \hline
ELMoHAR-note                  & 88.10\%          & \textbf{1.20} & 77.40\%          & 1.65         & 75.90\%        & 2.88          \\
GPTHAR-note                   & 87.30\%          & 2.98          & 79.90\%          & 1.52         & 84.80\%        & 1.81          \\
ELMoHAR & \textbf{90.70\%} & 1.25          & 80.00            & 1.33         & 83.30\%        & 2.95          \\
GPTHAR  & 89.70\%          & 3.06          & \textbf{81.90\%} & \textbf{1.1} & \textbf{87.20} & \textbf{0.92}\\
\bottomrule
\end{tabular}%
}
\end{table}

The integration of the temporal component significantly enhances the classification efficacy of both models, notably by reducing the standard deviation.  In the Milan and Cairo datasets, GPTHAR substantially outperforms ELMoHAR and ELMoHAR-note.  
An examination of the confusion matrices  \footnote{see all in Annex \ref{sec:cm} in the supplementary material} in Fig. \ref{fig:matrix_Cairo_main} reveals that in the Cairo dataset, activities like `Laundry' are better recognised. In the Aruba dataset, activities such as `Washing Dishes', `Meal Preparation', `Enter Home', and `Leave Home' exhibit improved classification accuracy. In Fig. \ref{fig:matrix_Milan_main} for Milan, activities like `Eve Med' and `Morning Meds' show a reduction in misclassifications.

\subsection{Attention Heads through a Case Study}

This case study examines the efficacy of GPT-based models and the role of 
 its causal model.  
Fig. \ref{fig:matrix_Cairo_main} previously showed that the recognition of the `Laundry' activity within the Cairo dataset improved both in GPTHAR (vs ELMoHAR) and in GPTAR (vs ELMoAR).   
 are often erroneously identified as `Leave Home' in models employing pre-trained ELMo embeddings (ELMoAR, ELMoHAR-note, and ELMoHAR). Indeed, Fig. \ref{fig:emb_cairo} shows that an embedding of Laundry is very close to embeddings of Leave Home.

  The dual-resident setting of the Cairo dataset exacerbates the issue of concurrent activities. 
Fig. \ref{fig:case_study} plots for a recording of `Laundry' the activation sequence and  attention heads by the GPT embedding, with our interpretation of the activity at the bottom.
 The activations of the back door motion sensor brings confusion with `Leave Home'.  
 We filtered the heads connected and preceding tokens 90 and 122 to represent only 2 sets of attention heads on two different moments: for tokens 31-90: a first resident turning on and off the motion sensor 'M026' in the laundry room; for tokens 93-122 :  the path from the back door to the bedroom. The attention head notably links the last bedroom sensor activation 'M007OFF' (token 122) and the back door motion sensor 'M027ON' (token 93), but not  the tokens 94, 102, 106 of the  'M026' sensor activities  during that period. Separate attention heads for these two moments show a selective attention by the GPT embeddings effectively tracks concurrent activities.

\section{Discussion}

\subsection{Summary}

In our study on recognizing daily living activities through ambient sensor classifications within smart homes, we embarked on a comparison of two distinct pretrained embeddings applied to ambient sensor activations. Notably, the Transformer Decoder-based embedding (akin to the GPT design) was shown superior in classification tasks when compared to the ELMo pretrained embeddings. By introducing a hierarchical structure, we aimed to exploit the inherent relationships among activity sequences, thereby refining the classification outcomes. The effectiveness of this approach was evident across all three datasets, with the GPTHAR version standing out especially. Furthermore, the inclusion of an hour-of-the-day embedding subtly yet significantly enhanced classification precision, particularly for activities with time-sensitive natures.

\subsection{Limitations and Future Works}

Our study used pre-segmented data, which restricts the potential application of these algorithms in real-time services. 
Throughout our experiments, we observed that the transformer architecture resulted in longer training times compared to the LSTM embedding-based architecture. 
Additionally, although we selected datasets that encompass various lifestyle configurations, our evaluation was limited to just three datasets of the CASAS benchmark. This raises potential concerns regarding the wider applicability and generalizability of our conclusions. Broadening the scope of our evaluations to encompass a more diverse range of datasets will be pivotal. This would not only test the model's robustness but also enhance its generalizability across various scenarios and environments. 

Our results show an increase in standard deviation values in datasets Aruba and Milan, with a hierarchical structure. We hypothesize for future works that model stability could be improved by the introduction of normalization layers.
Lastly, we see a promising avenue in automated segmentation learning. Moving away from pre-segmented data, investigating methods that allow for more natural and continuous activity recognition could pave the way for more realistic and adaptive models. This would potentially overcome the constraints posed by our current segmentation approach, leading to real-world applicable results.

\subsection{Conclusion}
In our smart home study, we highlight the benefits of using Transformer Decoder-based methods as an embedding, for recognising daily living activities. Our framework improves ADL recognition, with results favouring a hierarchical approach that excels in identifying inter-activity relationships. Incorporating temporal data significantly enhanced performance, particularly in noisy datasets.


\begin{ack}
This research work is supported by the Hi! PARIS Center.
\end{ack}


\bibliography{biblio}

\clearpage
\newpage

\clearpage
\appendix

\section{Hyper parameter search}
\label{sec:cross_validation_results_parameters_search_f1}
In this appendix, we present F1 scores during cross-validation of ELMoAR and GPTAR with, respectively, different window sizes and different head and layer numbers. The best hyperparameters are selected for the ELMoHAR-note, ELMoHAR and GPTHAR-note and GPTHAR algorithms. 

For ELMoAR, the best hyperparameters are: a window size of 60. For GPTAR, the best hyperparameters are : 8 attention heads and 3 layers of decoder.


\begin{table}[h]
\centering
\caption{Cross-validation F1-Scores with their standard deviation for GPTAR and ELMoAR models with various hyperparameters across the three datasets, to assess the Impact of context window size in ELMoAR and the number of layers and attention heads in GPTAR}
\label{tab:cross_validation_results_parameters_search_f1}
\resizebox{\columnwidth}{!}{%
\begin{tabular}{l|cc|cc|cc|l}
\toprule
                                & \multicolumn{2}{c}{Aruba}      & \multicolumn{2}{c}{Milan}      & \multicolumn{2}{c}{Cairo}      &                \\
                                & Macro F1 Score & std           & Macro F1 Score & std           & Macro F1 Score & std           & Average        \\ \hline
ELMoAR (Window 20)              & 83.47          & 1.83          & 71.10          & 2.25          & 66.37          & 3.24 & 73.65          \\
ELMoAR (Window 40)              & 82.93          & 2.12          & 70.73          & 2.43          & 66.70          & 4.07          & 73.45          \\
{ELMoAR (Window 60)}        & 83.47          & 2.61          & \it{72.40}          & \it{2.49}          & \it{67.23}          & \it{4.70}          & \textit{74.37} \\
{GPTAR  (8 Heads 3 Layers)} & 83.20          & 1.45 & \textbf{73.77} & \bf{2.19}          & \textbf{70.90} & \bf{3.48}          & \textbf{75.96} \\
GPTAR (8 Heads 4 Layers)        & 83.53 & 1.48          & 72.30          & 1.62 & 69.03          & 4.42          & 74.95          \\
GPTAR (12 Heads 6 Layers)       & \bf{83.57}          & \bf{1.57}          & 73.07          & 2.13          & 68.27          & 4.08          & 74.97\\
\bottomrule
\end{tabular}%
}
\end{table}

\section{Detailed Algorithms Metrics}
\label{sec:resultsDetailedScores}

In this appendix, we present a comprehensive breakdown of the performance metrics for the algorithms used in our study. The table encompasses test results from three distinct datasets: Aruba (Table \ref{tab:detailed_scores_aruba}), Milan (Table \ref{tab:detailed_scores_milan}), and Cairo (Table \ref{tab:detailed_scores_cairo}). These metrics show that for the most simple dataset, Aruba, ELMoHAR and GPTHAR perform closely, ranking first or second depending on the chosen metric.
For the more complex datasets, Milan and Cairo, GPTHAR outperforms all the other algorithms, regardless of the choice of metric.

\begin{table}[!h]
\centering
\caption{Detailed Algorithms Scores over the datasets Aruba}
\label{tab:detailed_scores_aruba}
\resizebox{\columnwidth}{!}{%
\begin{tabular}{lcccccc}
\toprule
                   & ELMoAR  & GPTAR   & ELMoHAR-note & GPTHAR-note & ELMoHAR          & GPTHAR           \\ \hline
Accuracy           & 97.00\% & 97.10\% & 98.40\%      & 98.20\%     & 98.50\%          & \textbf{98.52\%} \\
Precision          & 86.00\% & 90.30\% & 89.90\%      & 87.40\%     & \textbf{94.30\%} & 91.20\%          \\
Recall             & 84.70\% & 85.10\% & 88.10\%      & 88.60\%     & 89.70\%          & \textbf{89.80\%} \\
F1 Score           & 84.80\% & 86.10\% & 88.10\%      & 87.30\%     & \textbf{90.70\%} & 89.70\%          \\
Balanced Accuracy  & 84.76\% & 85.18\% & 88.22\%      & 88.55       & 89.71\%          & \textbf{90.10\%} \\
Weighted Precision & 96.70\% & 97.00\% & 98.10\%      & 98.00\%     & \textbf{98.50\%} & 98.00\%          \\
Weighted Recall    & 97.00\% & 97.00\% & 98.40\%      & 92.20\%     & \textbf{98.50\%} & 98.30\%          \\
Weighted F1 Score  & 96.9\%  & 97.00\% & 98.00\%      & 98.00\%     & \textbf{98.30\%} & 98.20\%\\
\bottomrule
\end{tabular}%
}
\end{table}
\begin{table}[!h]
\centering
\caption{Detailed Algorithms Scores over the dataset Milan}
\label{tab:detailed_scores_milan}
\resizebox{\columnwidth}{!}{%
\begin{tabular}{lcccccc}
\toprule
                   & ELMoAR  & GPTAR   & ELMoHAR-note & GPTHAR-note & ELMoHAR & GPTHAR           \\ \hline
Accuracy           & 87.50\% & 88.20\% & 90.00\%      & 91.30\%     & 90.60\% & \textbf{91.9\%}  \\
Precision          & 75.90\% & 80.20\% & 85.60\%      & 87.60\%     & 88.90\% & \textbf{90.00\%} \\
Recall             & 68.40   & 68.50\% & 73.90        & 76.90\%     & 75.60\% & \textbf{79.20\%} \\
F1 Score           & 70.80\% & 70.80\% & 77.40\%      & 79.90\%     & 80.00\% & \textbf{81.90\%} \\
Balanced Accuracy  & 68.51\% & 68.55\% & 73.91\%      & 76.87\%     & 77.84\% & \textbf{79.22\%} \\
Weighted Precision & 86.80\% & 88.20\% & 89.60\%      & 91.20\%     & 90.70\% & \textbf{91.90\%} \\
Weighted Recall    & 87.50\% & 88.20\% & 90.00\%      & 91.30\%     & 90.60\% & \textbf{91.90\%} \\
Weighted F1 Score  & 86.70\% & 87.60\% & 89.20\%      & 90.70\%     & 90.00\% & \textbf{91.70\%}\\
\bottomrule
\end{tabular}%
}
\end{table}
\begin{table}[!h]
\centering
\caption{Detailed Algorithms Scores over the dataset Cairo}
\label{tab:detailed_scores_cairo}
\resizebox{\columnwidth}{!}{%
\begin{tabular}{lcccccc}
\toprule
                   & ELMoAR  & GPTAR   & ELMoHAR-note & GPTHAR-note & ELMoHAR & GPTHAR           \\ \hline
Accuracy           & 81.1\%  & 83.40\% & 87.30\%      & 91.00\%     & 90.80\% & \textbf{93.20}   \\
Precision          & 72.70\% & 76.60\% & 79.70\%      & 87.40\%     & 87.30\% & \textbf{89.80\%} \\
Recall             & 69.20\% & 71.40\% & 74.70\%      & 83.40\%     & 82.00\% & \textbf{86.60\%} \\
F1 Score           & 70.50\% & 73.20\% & 75.90\%      & 84.80\%     & 83.30\% & \textbf{87.20\%} \\
Balanced Accuracy  & 69.12\% & 71.33\% & 74.75\%      & 83.58\%     & 81.87\% & \textbf{86.74\%} \\
Weighted Precision & 81.10\% & 82.40\% & 86.90\%      & 91.10\%     & 90.60\% & \textbf{93.20}   \\
Weighted Recall    & 81.10\% & 83.40\% & 87.30\%      & 91.00\%     & 90.80\% & \textbf{93.20}   \\
Weighted F1 Score  & 80.90\% & 82.30\% & 86.80\%      & 90.50\%     & 90.30\% & \textbf{92.70\%}\\
\bottomrule
\end{tabular}%
}
\end{table}

%
%
%


\section{Confusion Matrix}
\label{sec:cm}
In this section, we report on the confusion matrices for various algorithms across three datasets. A notable observation is that the more complex architectures, which include time encodings (namely ELMoHAR and GPTHAR), exhibit fewer misclassifications compared to simpler models.

This improvement is particularly evident in specific activities. For instance, in the Aruba dataset, activities like 'Washing Dishes', 'Meal Preparation', 'Enter Home', and 'Leave Home' are classified more accurately. Similarly, in the Milan dataset, activities such as 'Eve Med' and 'Morning Meds' show a marked decrease in misclassifications. Additionally, in the Cairo dataset, meal-related activities like 'Breakfast', 'Lunch', and 'Dinner' are more accurately identified.

These findings highlight the effectiveness of our improved algorithms in reducing misclassifications.
\begin{figure*}[h]

     \begin{subfigure}{0.49\textwidth}
         \centering
         \includegraphics[width=\textwidth]{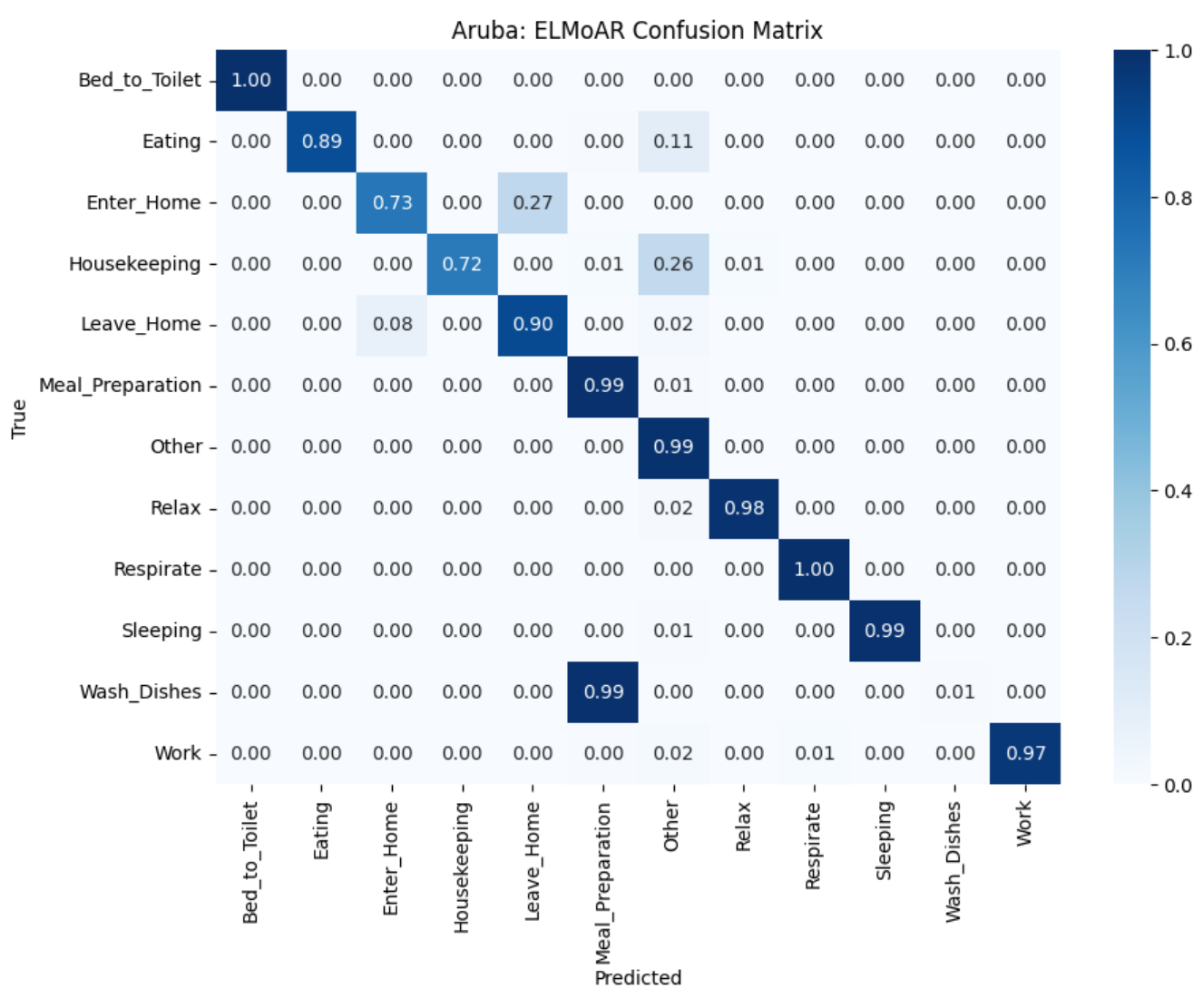}
         \caption{ELMoAR}
         \label{fig:aruba_ELMoAR}
     \end{subfigure}
     \begin{subfigure}{0.49\textwidth}
         \centering
         \includegraphics[width=\textwidth]{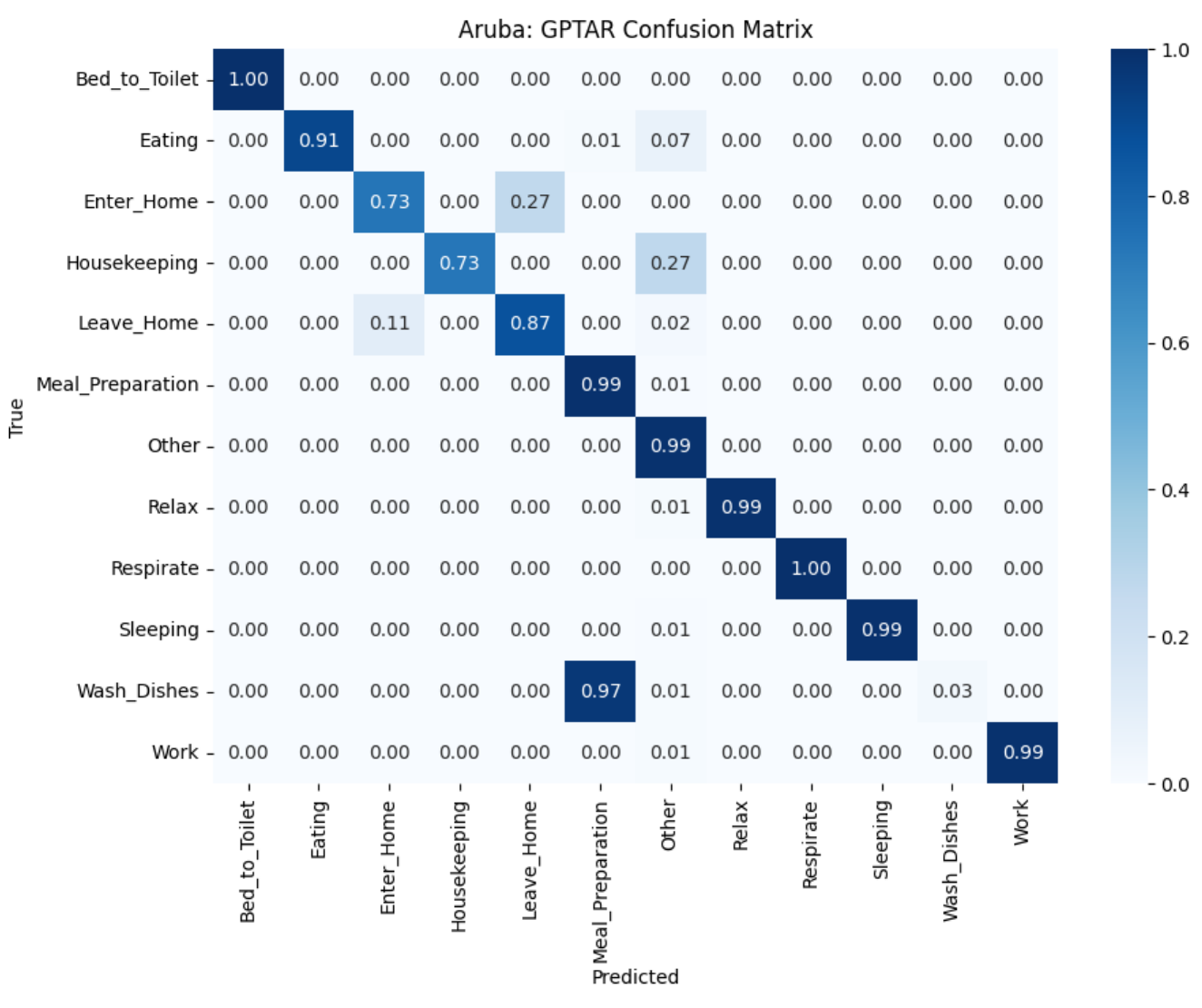}
         \caption{GPTAR}
         \label{fig:aruba_GPTAR}
     \end{subfigure}
     \begin{subfigure}{0.49\textwidth}
         \centering
         \includegraphics[width=\textwidth]{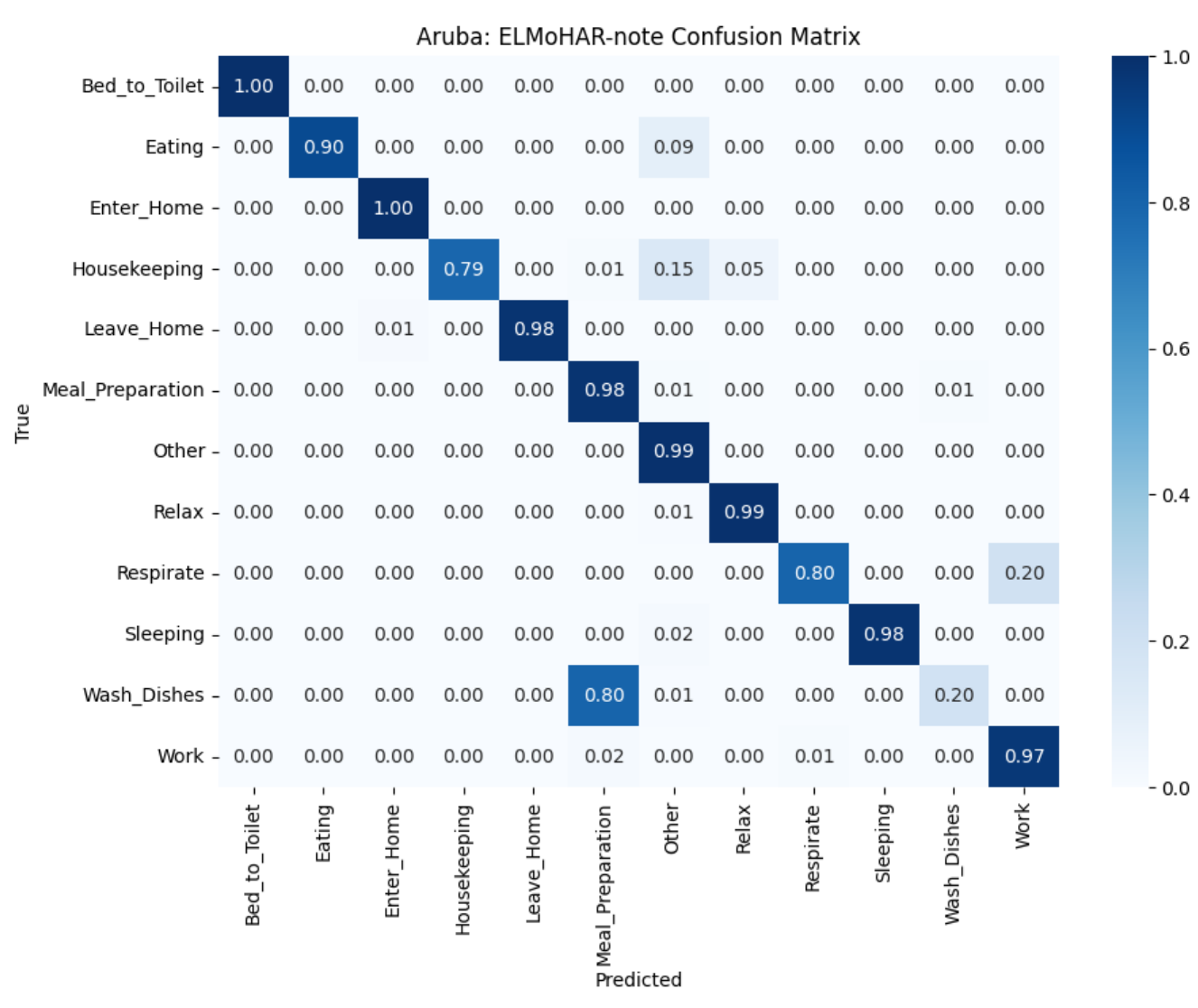}
         \caption{ELMoHAR-note}
         \label{fig:aruba_ELMoHAR_N}
     \end{subfigure}
     \begin{subfigure}{0.49\textwidth}
         \centering
         \includegraphics[width=\textwidth]{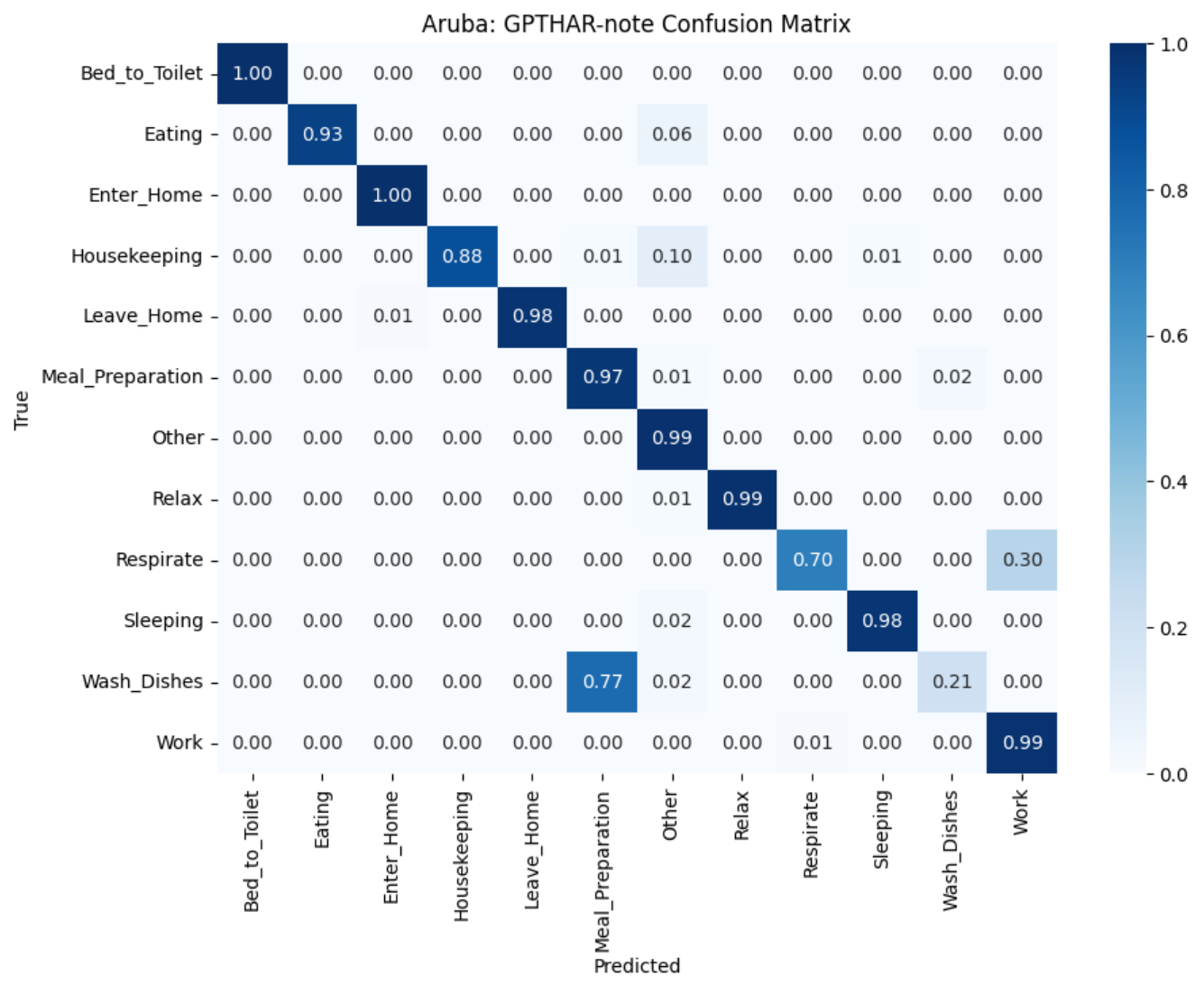}
         \caption{GPTHAR-note}
         \label{fig:aruba_GPTHAR_N}
     \end{subfigure}
     \begin{subfigure}{0.49\textwidth}
         \centering
         \includegraphics[width=\textwidth]{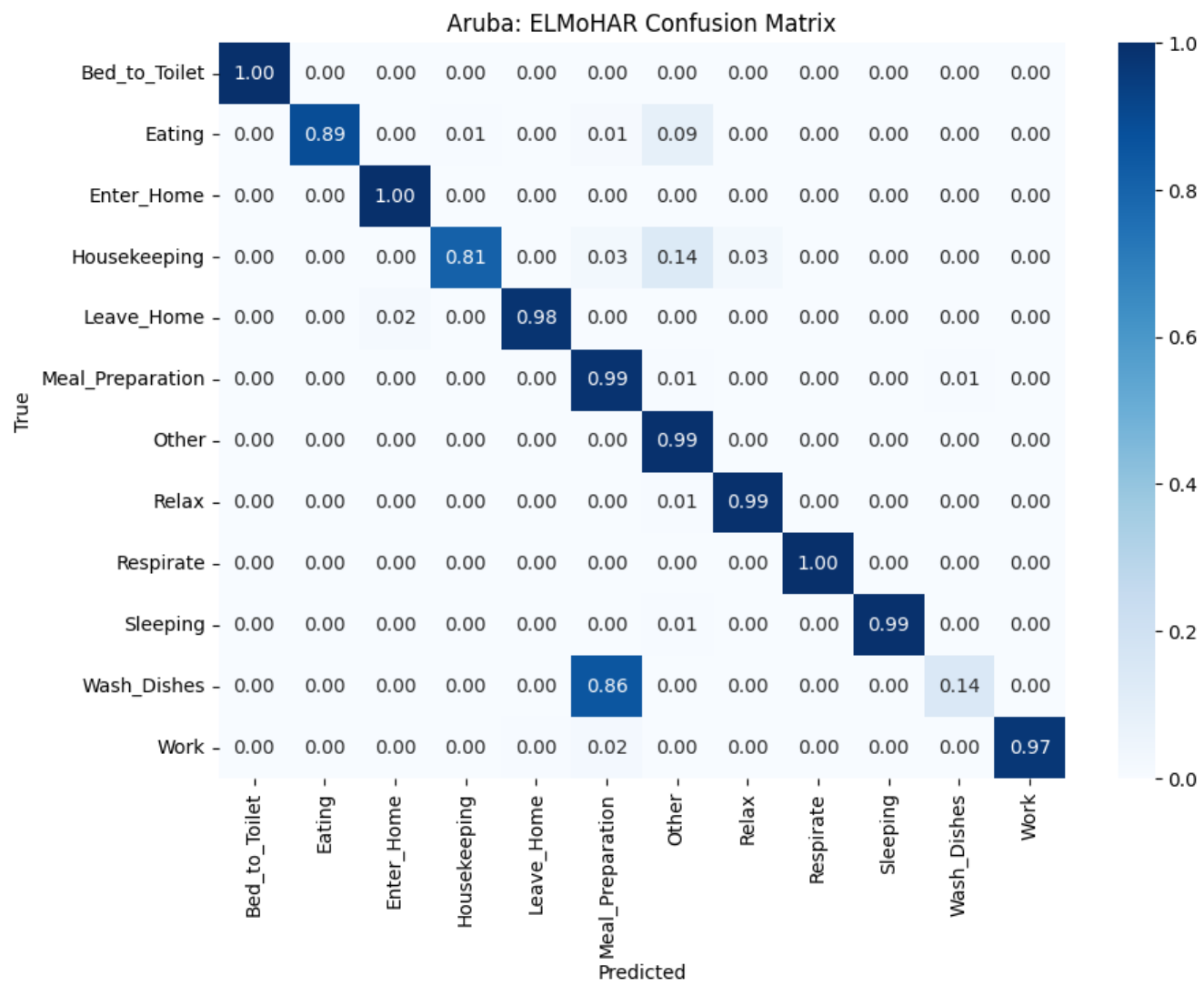}
         \caption{ELMoHAR}
         \label{fig:aruba_ELMoHAR}
     \end{subfigure}
     \begin{subfigure}{0.49\textwidth}
         \centering
         \includegraphics[width=\textwidth]{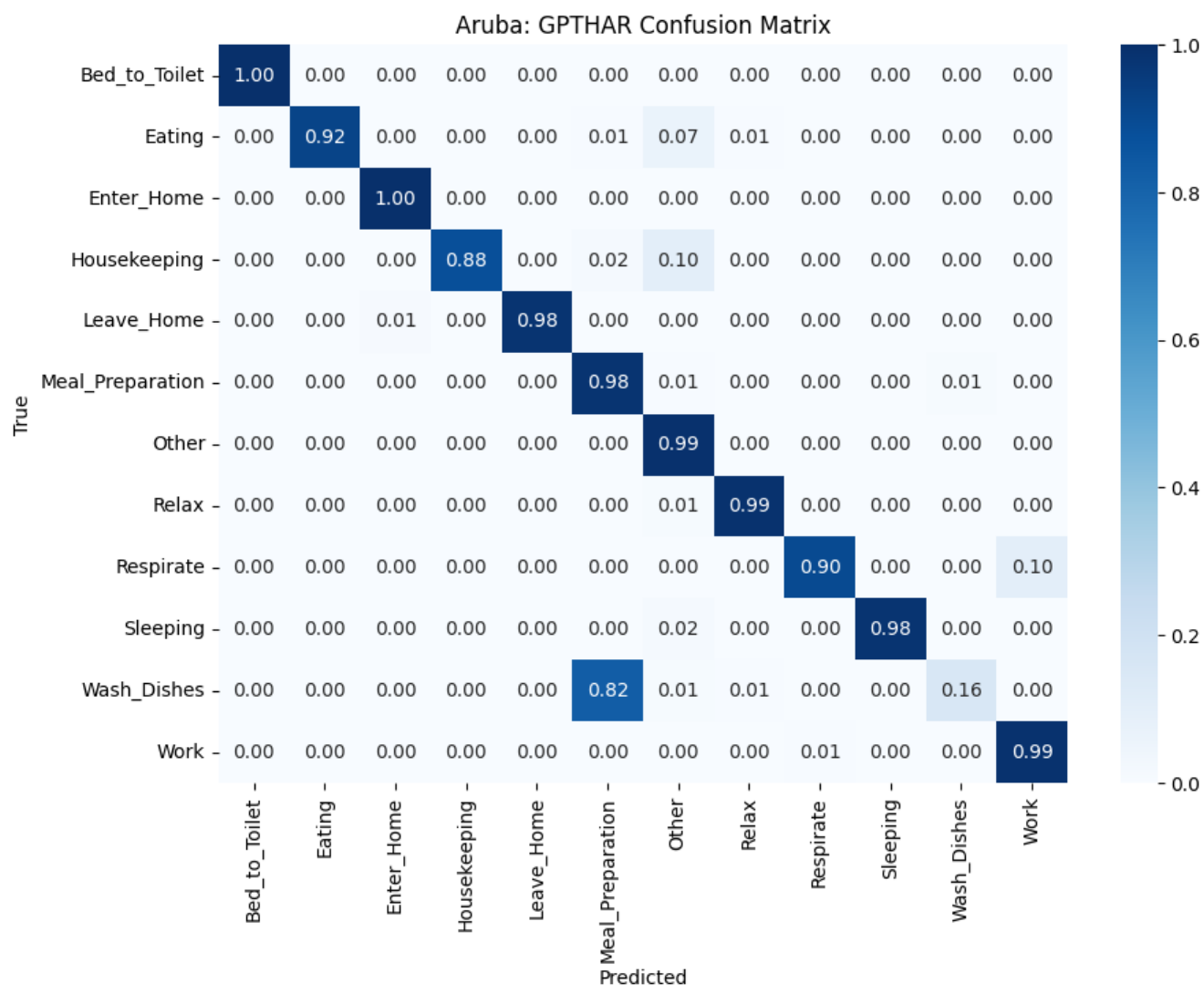}
         \caption{GPTHAR}
         \label{fig:aruba_GPTHAR}
     \end{subfigure}
     \vspace{10pt}
    \caption{ Confusion matrices per algorithm on the Aruba dataset.}
    \label{fig:matrix_aruba}
\end{figure*}
\begin{figure*}[h]

     \begin{subfigure}{0.49\textwidth}
         \centering
         \includegraphics[width=\textwidth]{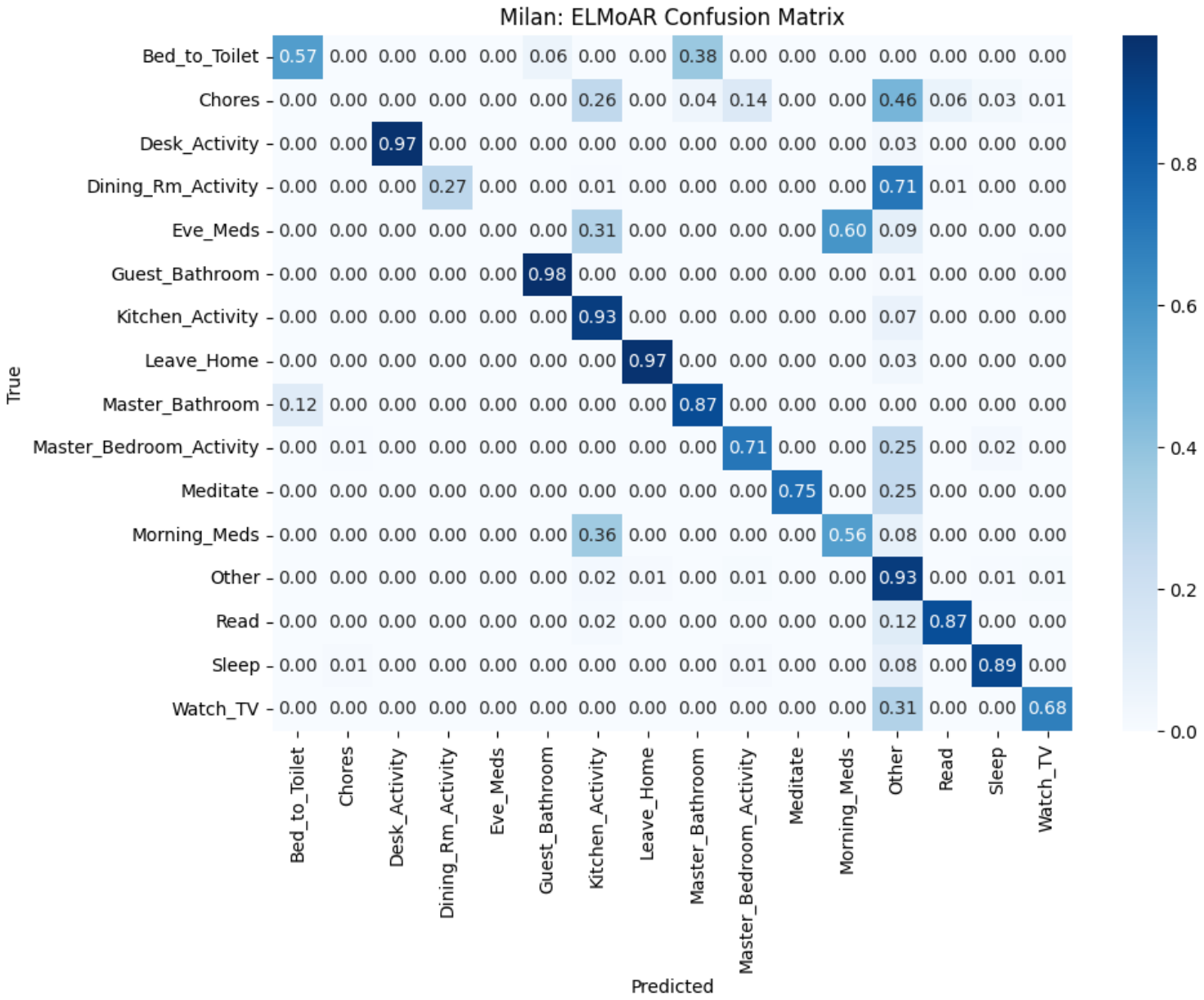}
         \caption{ELMoAR}
         \label{fig:Milan_ELMoAR}
     \end{subfigure}
     \vspace{10pt}
     \begin{subfigure}{0.49\textwidth}
         \centering
         \includegraphics[width=\textwidth]{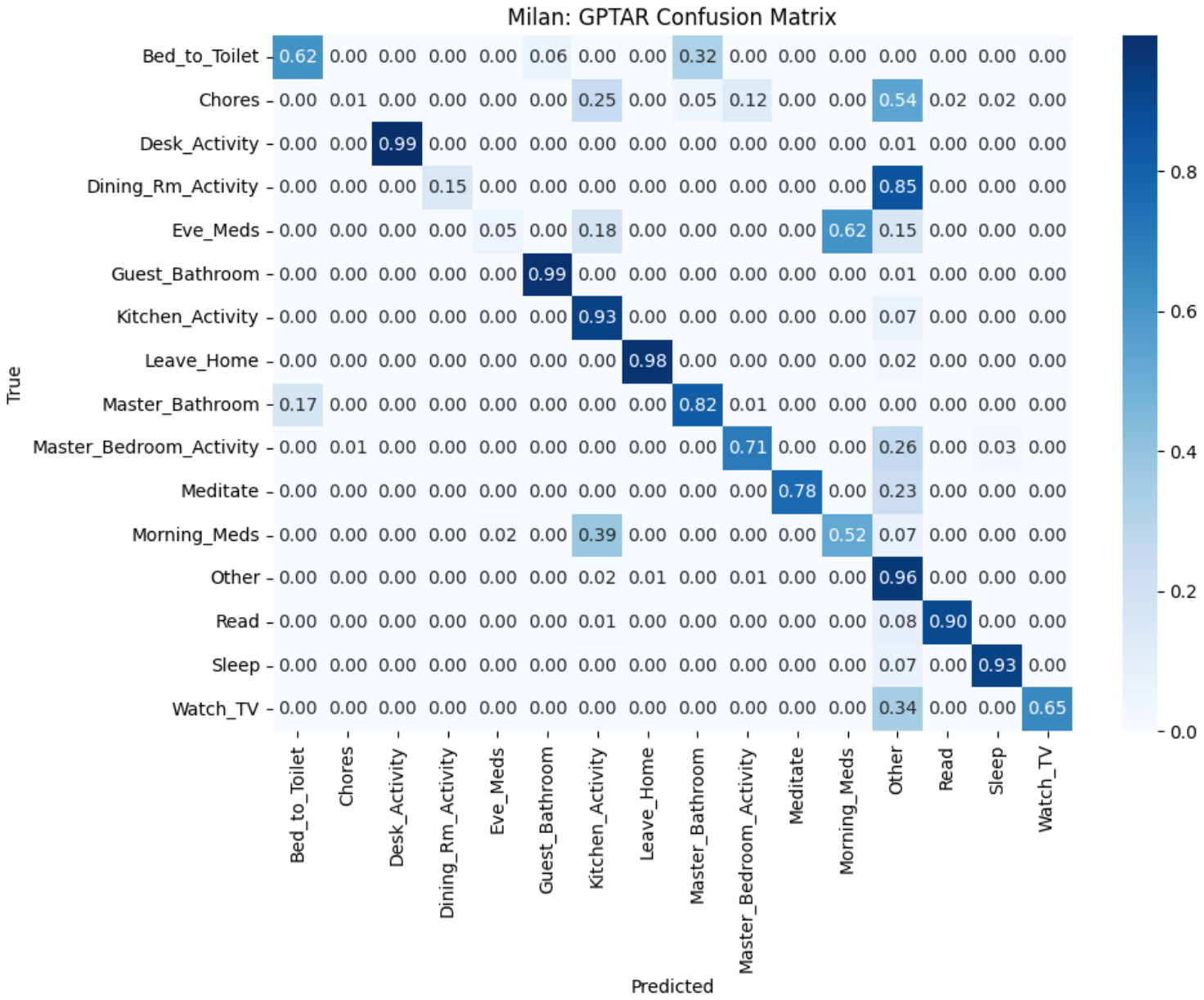}
         \caption{GPTAR}
         \label{fig:Milan_GPTAR}
     \end{subfigure}
     \vspace{10pt}
     \begin{subfigure}{0.49\textwidth}
         \centering
         \includegraphics[width=\textwidth]{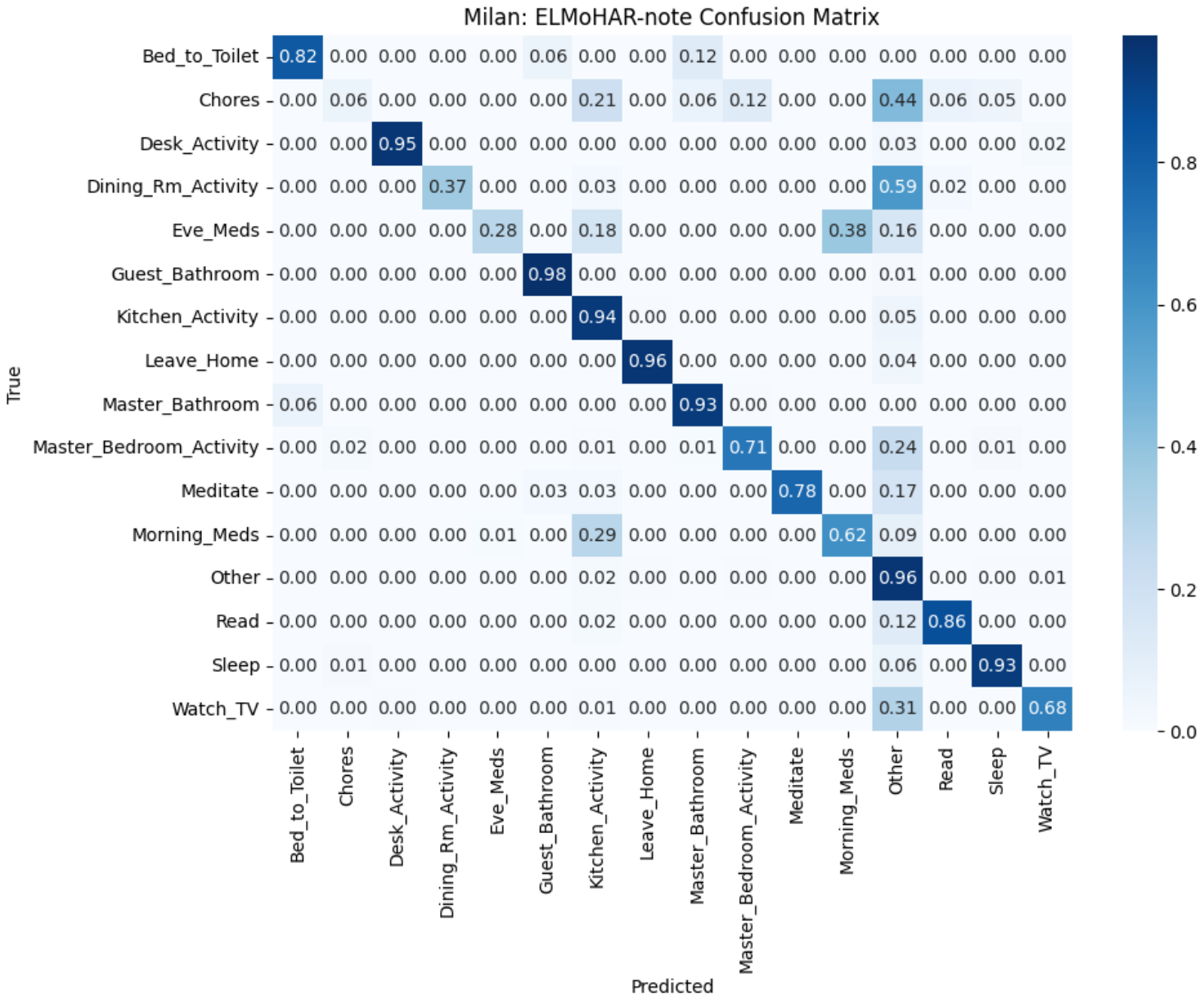}
         \caption{ELMoHAR-note}
         \label{fig:Milan_ELMoHAR_N}
     \end{subfigure}
     \begin{subfigure}{0.49\textwidth}
         \centering
         \includegraphics[width=\textwidth]{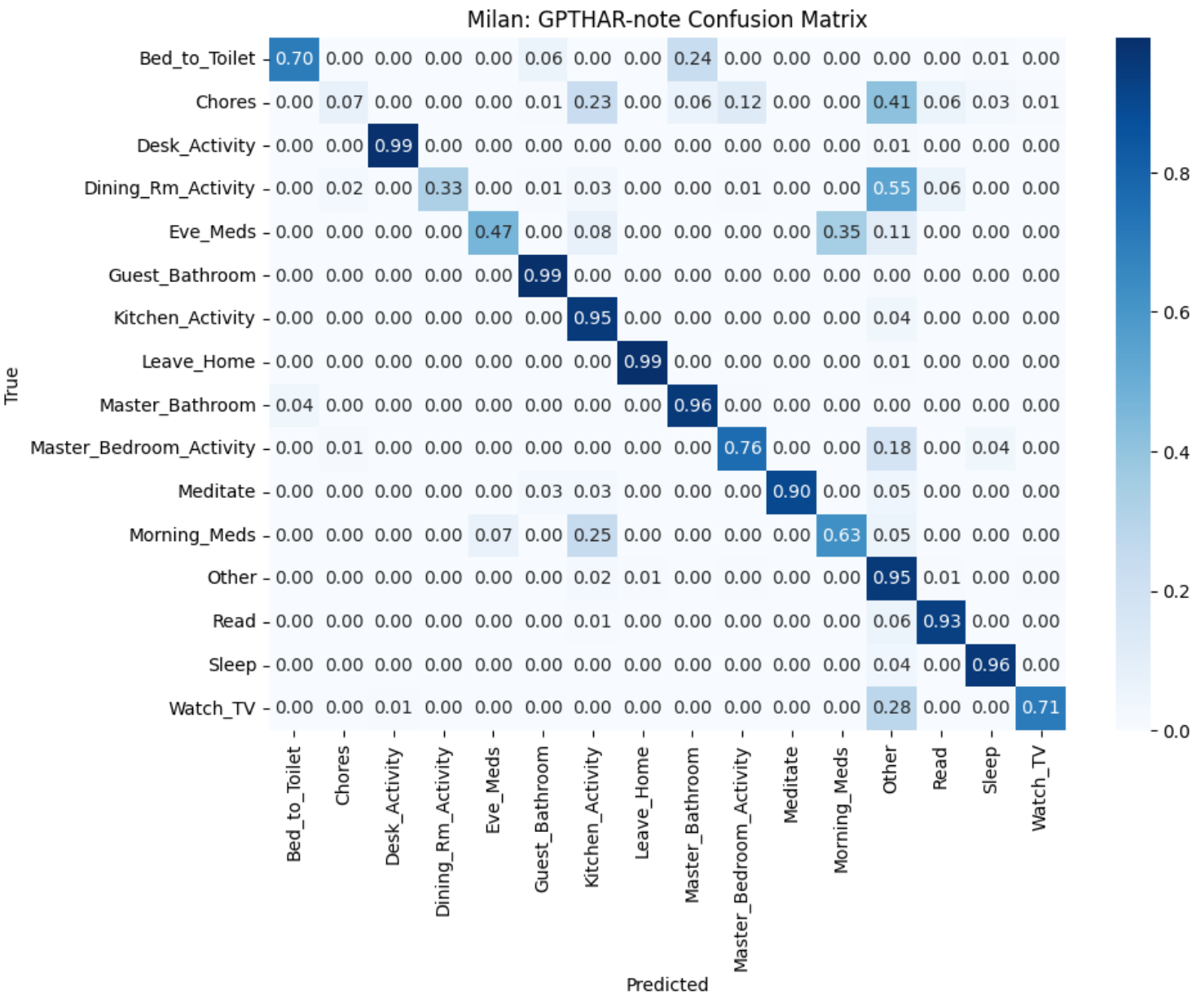}
         \caption{GPTHAR-note}
         \label{fig:Milan_GPTHAR_N}
     \end{subfigure}
     \vspace{10pt}
     \begin{subfigure}{0.49\textwidth}
         \centering
         \includegraphics[width=\textwidth]{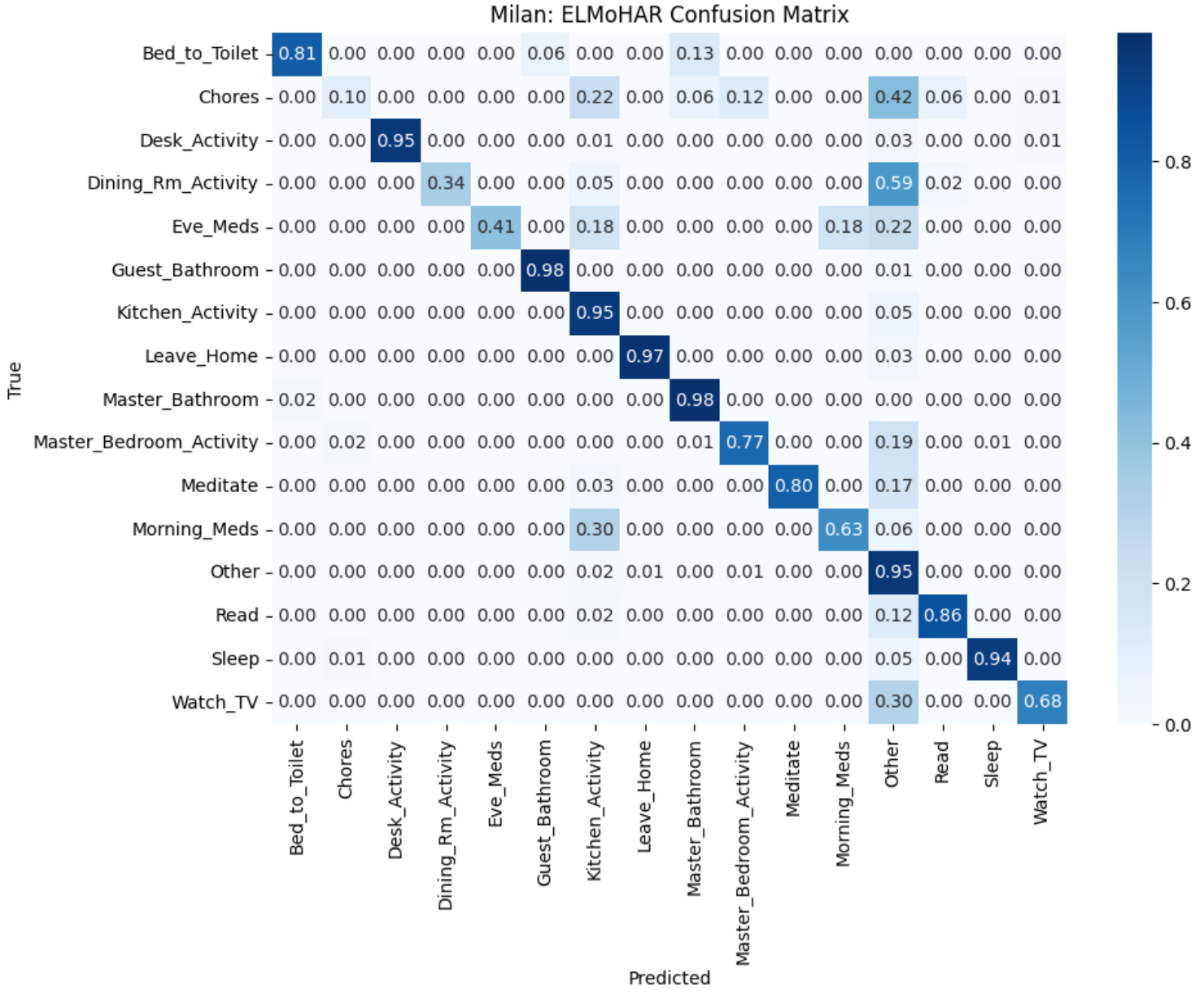}
         \caption{ELMoHAR}
         \label{fig:Milan_ELMoHAR}
     \end{subfigure}
     \begin{subfigure}{0.49\textwidth}
         \centering
         \includegraphics[width=\textwidth]{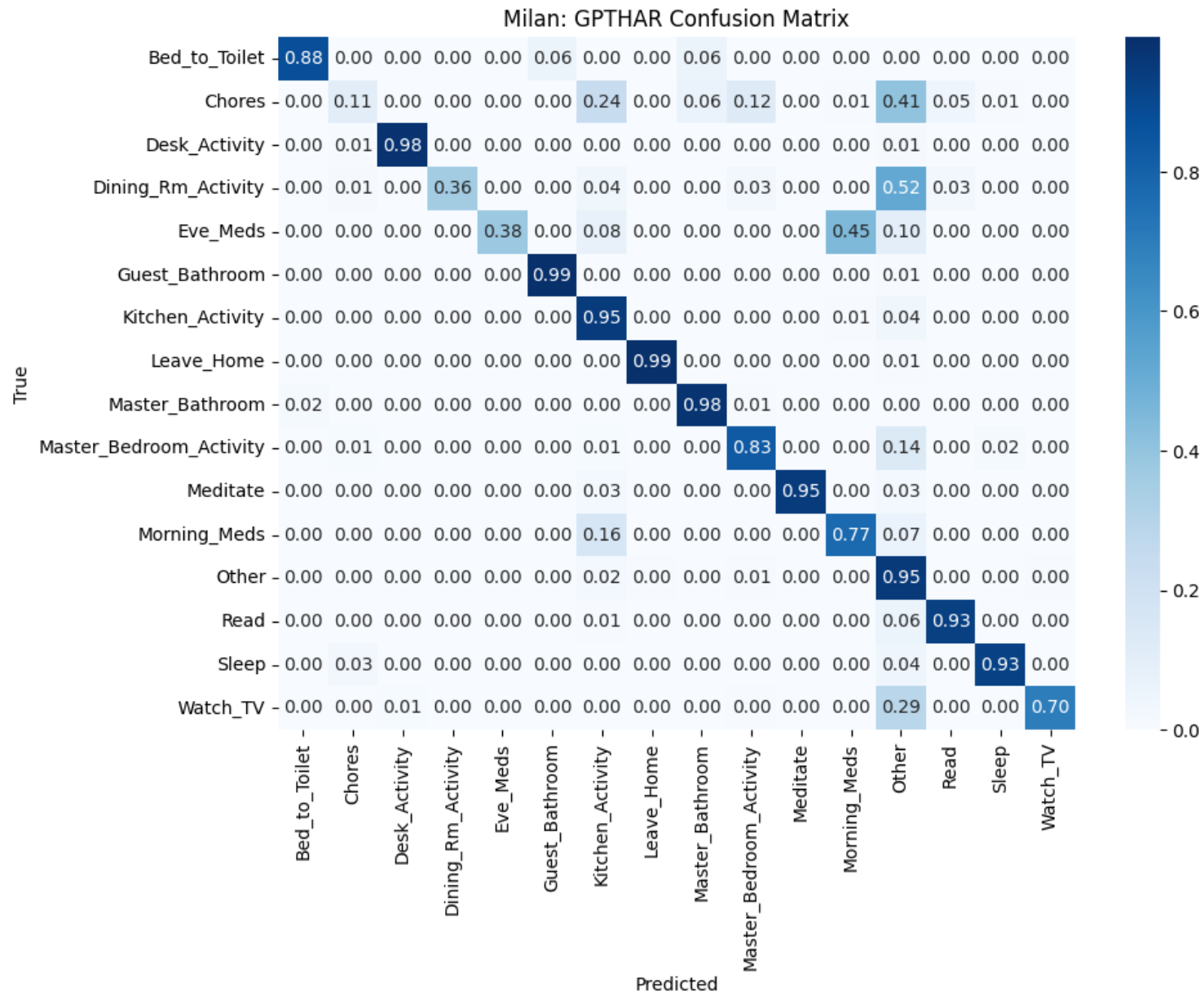}
         \caption{GPTHAR}
         \label{fig:Milan_GPTHAR}
     \end{subfigure}
     \vspace{10pt}
    \caption{Confusion matrices per algorithm on the Milan dataset.}
    \label{fig:matrix_Milan}
\end{figure*}
\begin{figure*}[h]
         \vspace{-90pt}

     \begin{subfigure}{0.49\textwidth}
         \centering
         \includegraphics[width=\textwidth]{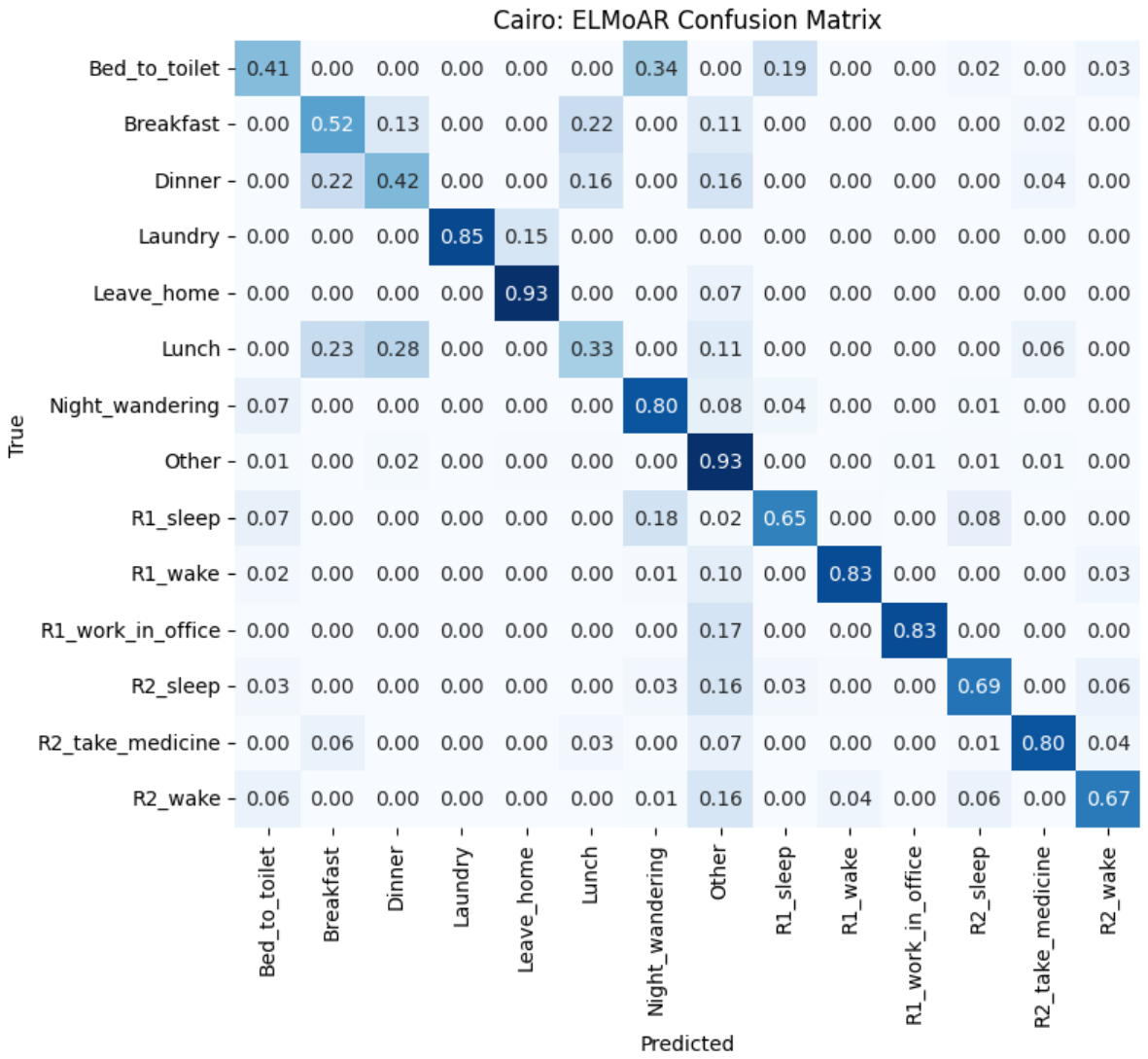}
         \vspace{-80pt}
         \caption{ELMoAR}
         \vspace{-60pt}
         \label{fig:Cairo_ELMoAR}
     \end{subfigure}
     \begin{subfigure}{0.49\textwidth}
         \centering
         \includegraphics[width=\textwidth]{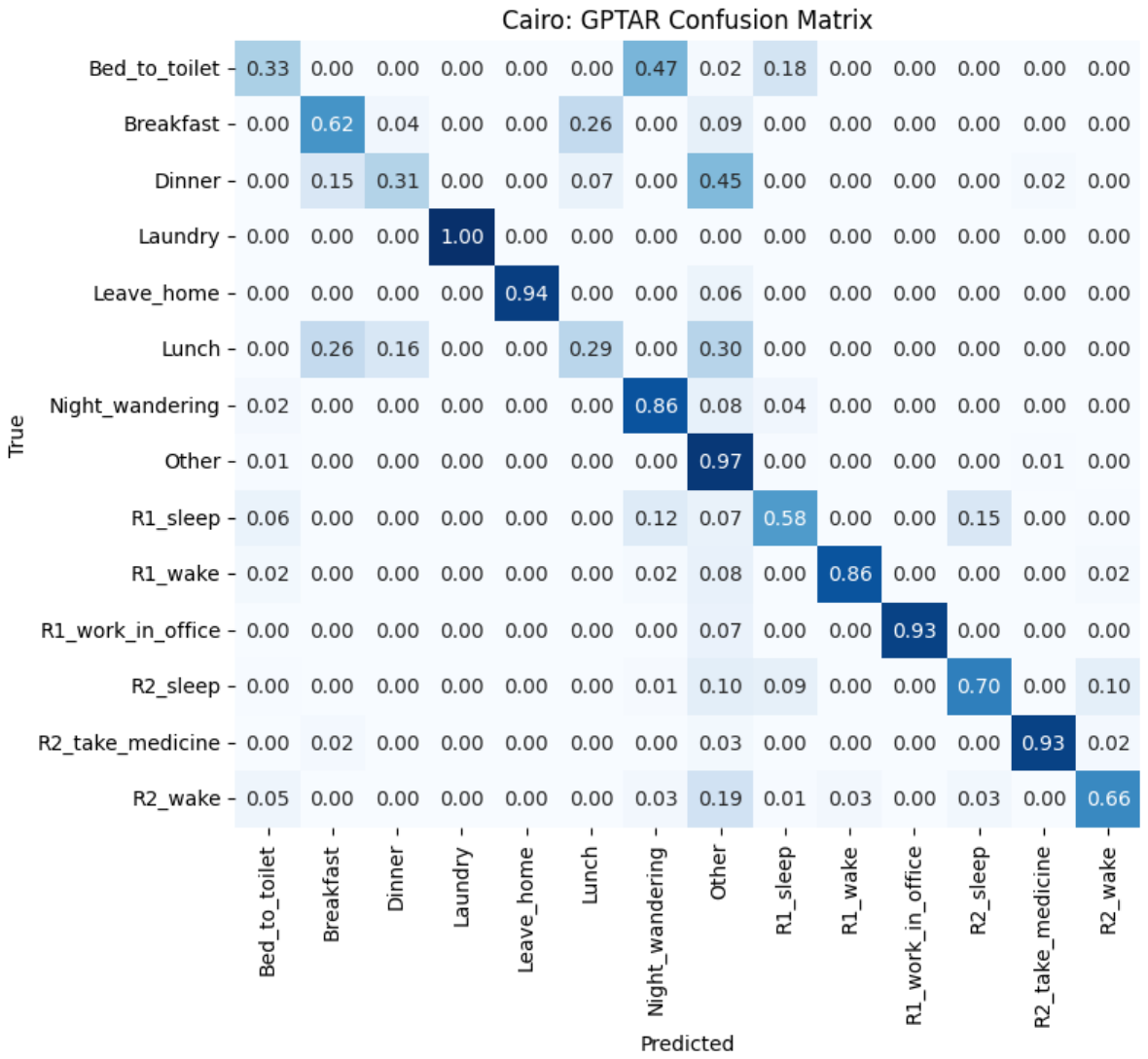}
         \vspace{-80pt}
         \caption{GPTAR}
         \vspace{-60pt}
         \label{fig:Cairo_GPTAR}
     \end{subfigure}
     \begin{subfigure}{0.49\textwidth}
         \centering
         \includegraphics[width=\textwidth]{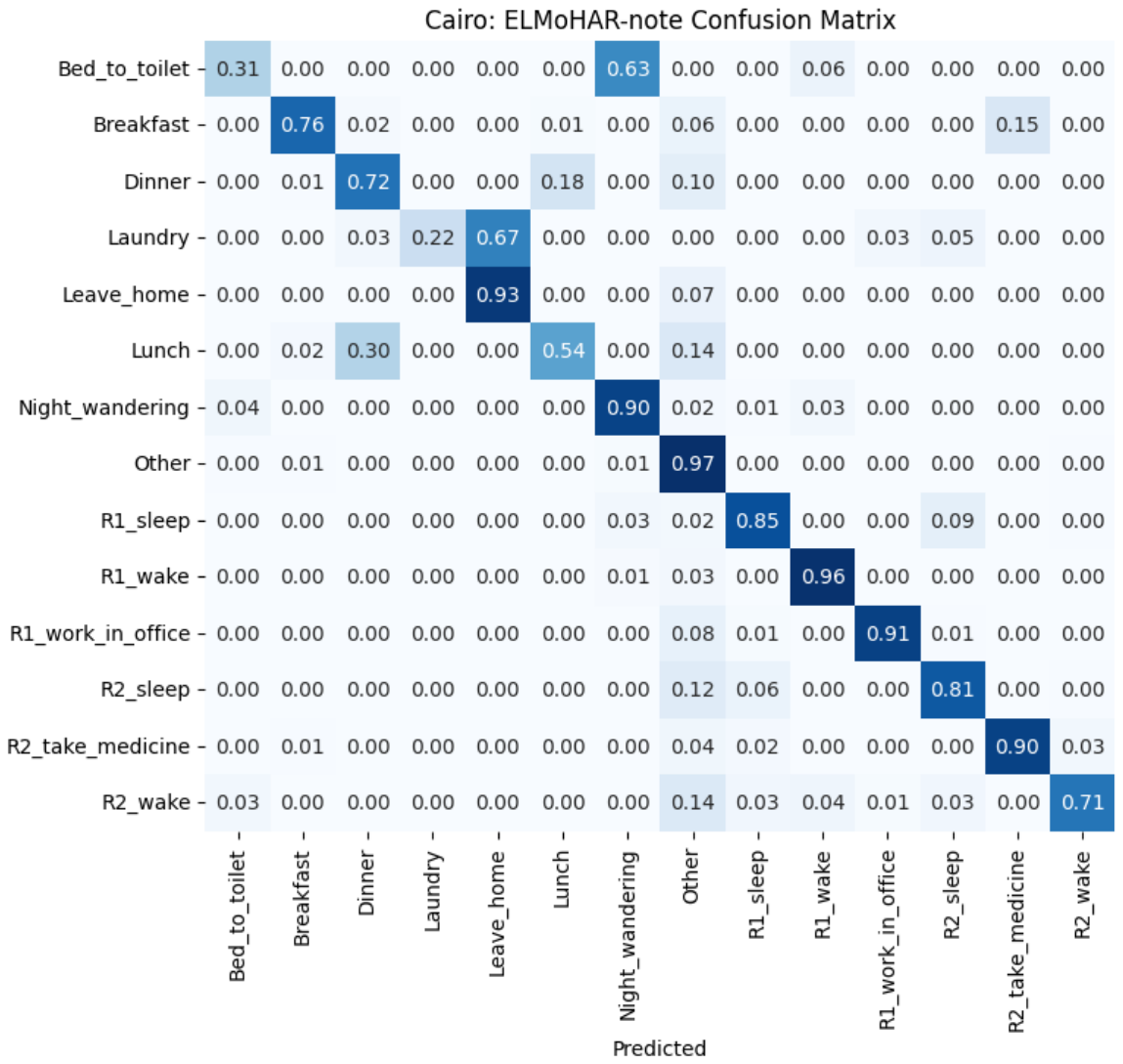}
         \vspace{-80pt}
         \caption{ELMoHAR-note}
         \vspace{10pt}
         \label{fig:Cairo_ELMoHAR_N}
     \end{subfigure}
     \begin{subfigure}{0.49\textwidth}
         \centering
         \includegraphics[width=\textwidth]{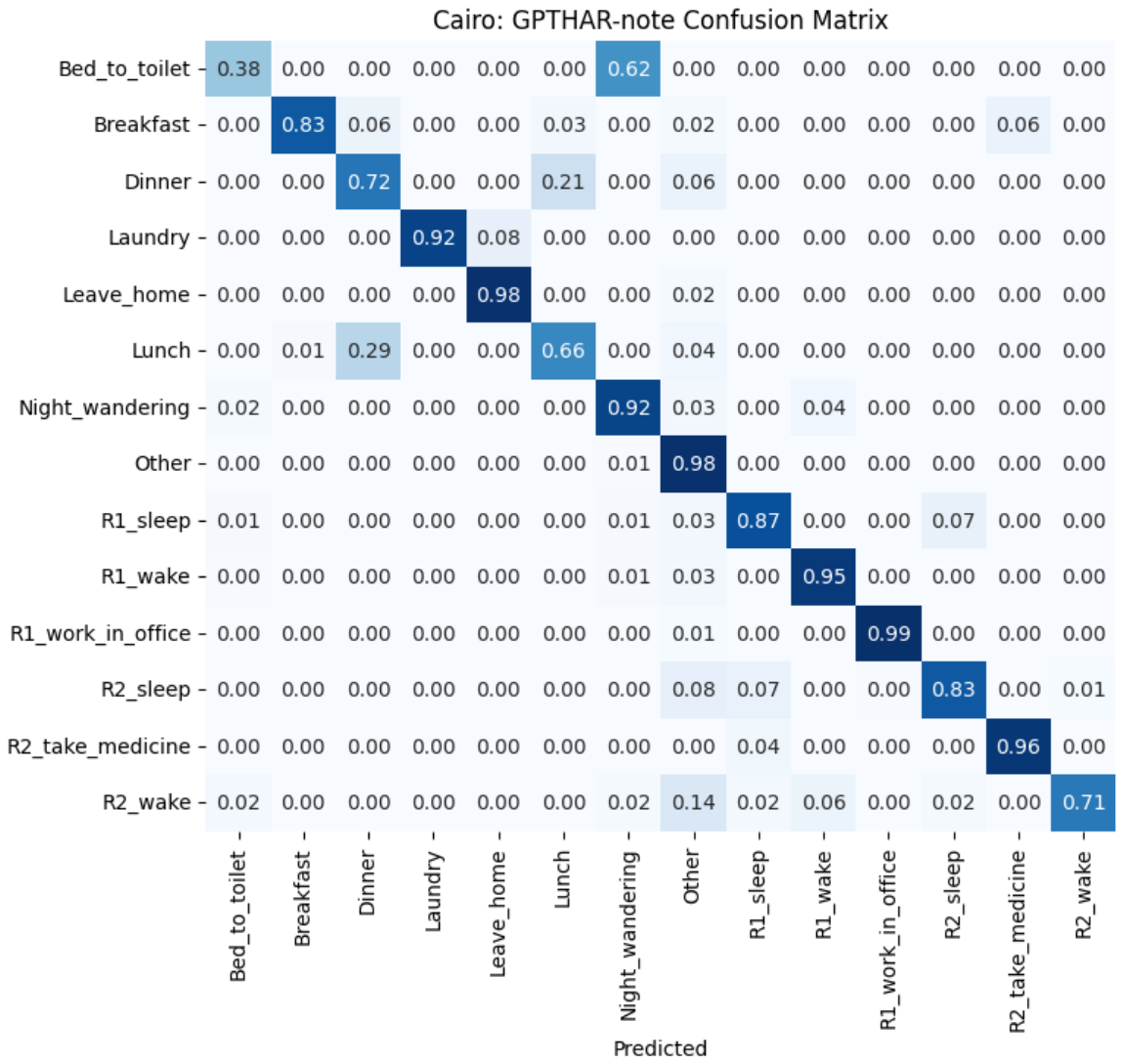}
         \vspace{-80pt}
         \caption{GPTHAR-note}
         \vspace{10pt}
         \label{fig:Cairo_GPTHAR_N}
     \end{subfigure}
     \begin{subfigure}{0.49\textwidth}
         \centering
         \includegraphics[width=\textwidth]{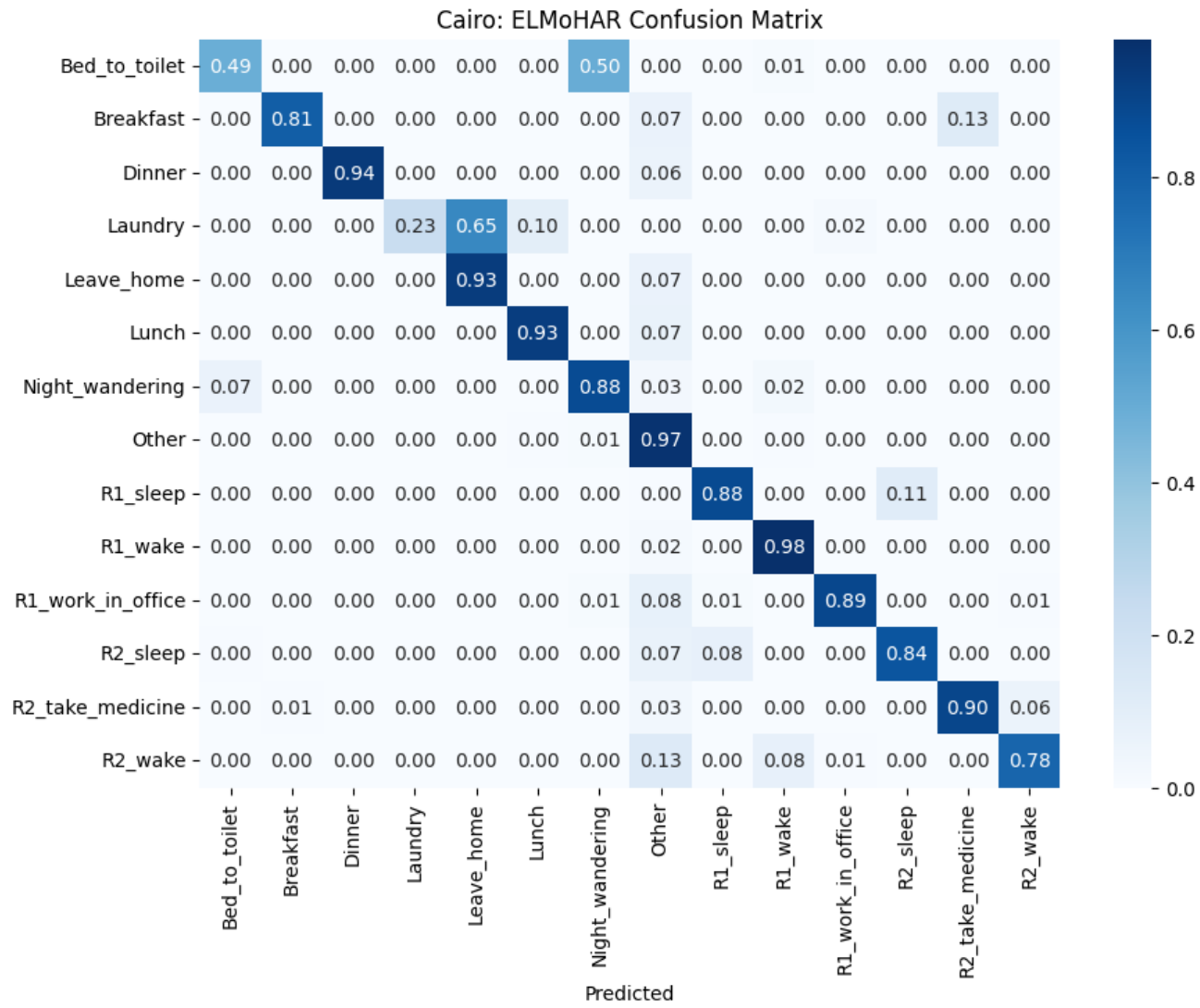}
         \caption{ELMoHAR}
         \vspace{-60pt}
         \label{fig:Cairo_ELMoHAR}
     \end{subfigure}
     \begin{subfigure}{0.49\textwidth}
         \centering
         \includegraphics[width=\textwidth]{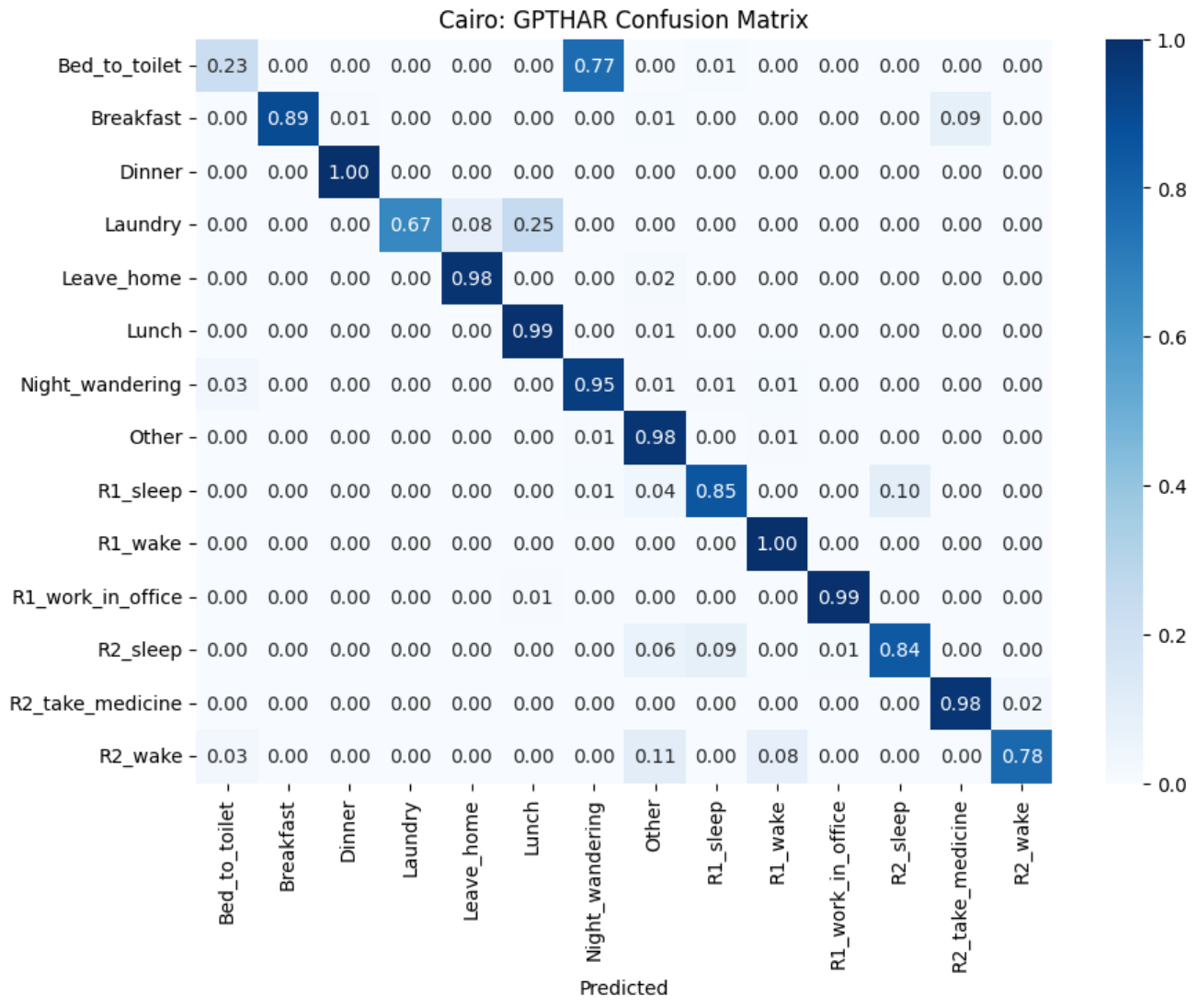}
         \caption{GPTHAR}
         \vspace{-60pt}
         \label{fig:Cairo_GPTHAR}
     \end{subfigure}
     \vspace{65pt}
    \caption{Confusion matrices per algorithm on the  Cairo dataset.}
    \label{fig:matrix_Cairo}
\end{figure*}

\clearpage

\section{Datasets Details}
\label{sec:dataset_details}

\begin{figure}[!hbt]
    \centering
    \includegraphics[width=\columnwidth, trim=0.5cm 2cm 0.2cm 0cm, clip]{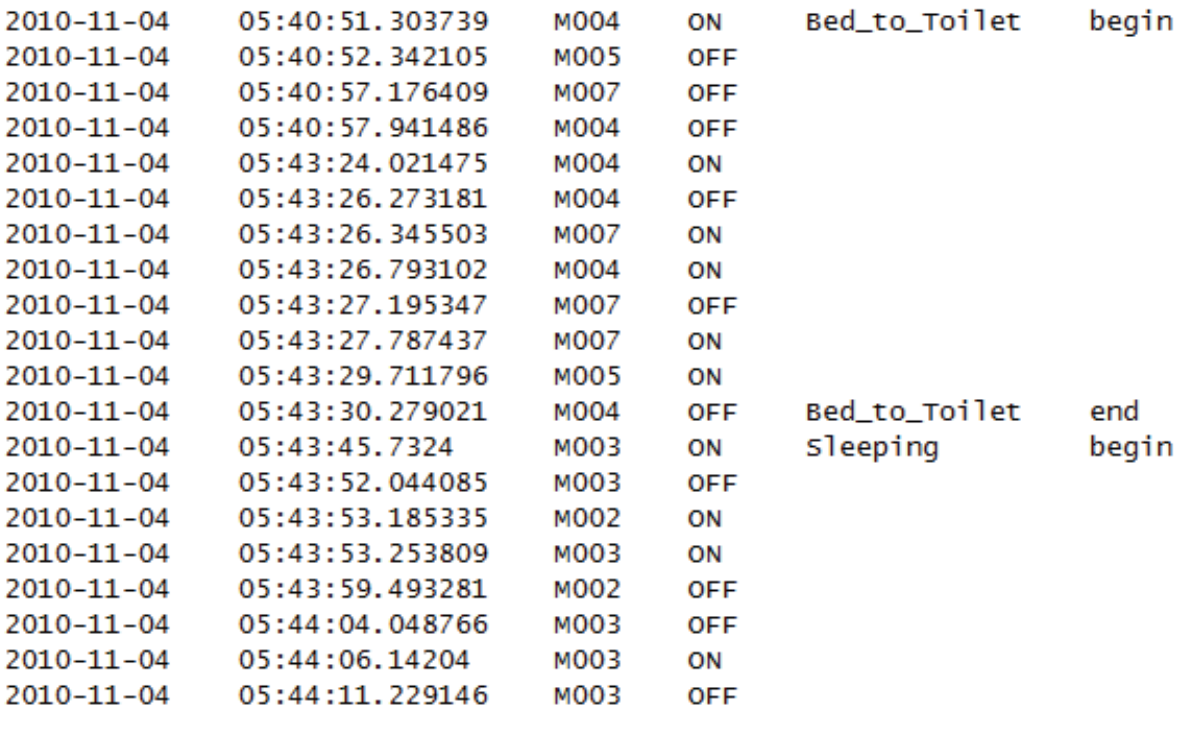}
    \vspace{5pt}
    \caption{Extract of a CASAS dataset log}
    \label{fig:casas_datasets_extract_example}
    \vspace{10pt}
\end{figure}

\begin{figure}[!hbt]
\centering
    \begin{subfigure}{0.48\textwidth}
        \centering
        \includegraphics[width=\textwidth]{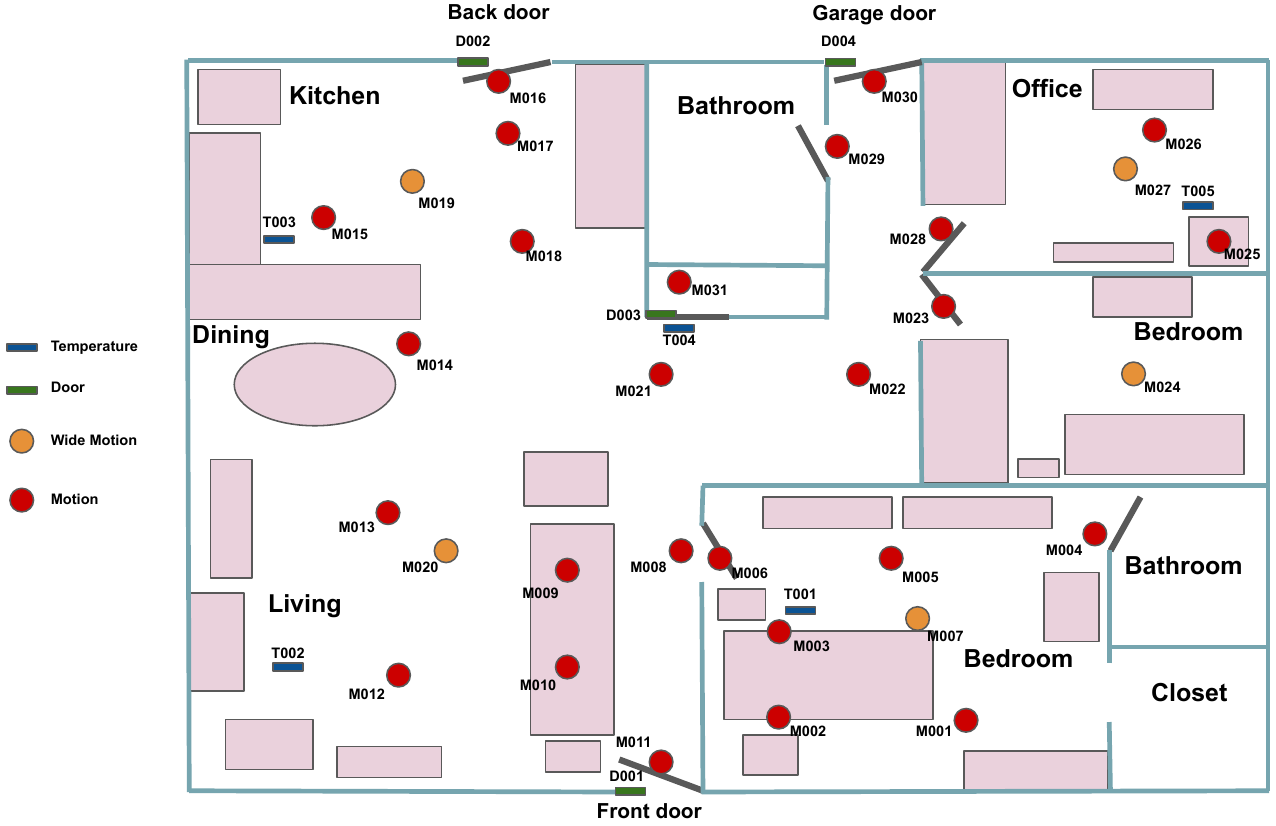}
        \caption{Aruba dataset floor map}
        \label{fig:aruba_floor_map}
        \vspace{10pt}
    \end{subfigure}
    \hfill
    \begin{subfigure}{0.48\textwidth}
        \centering
        \includegraphics[width=\textwidth]{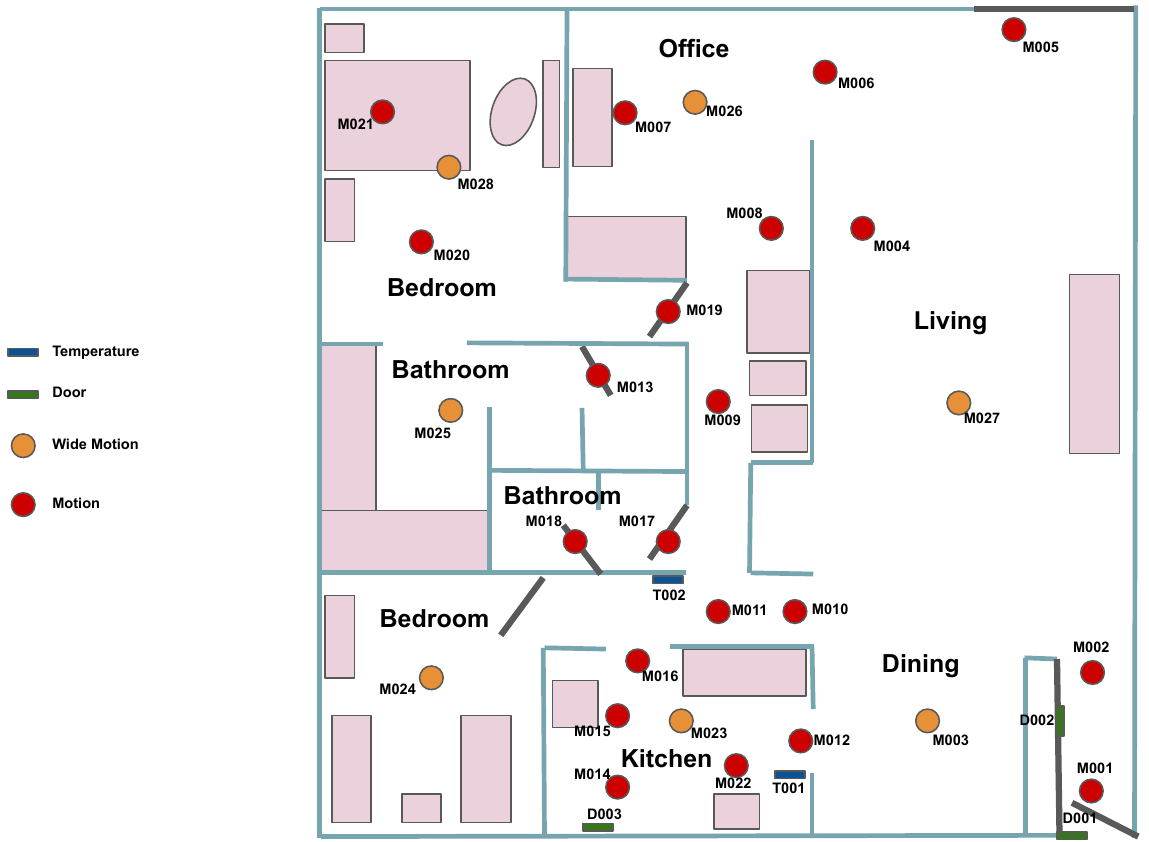}
        \caption{Milan dataset floor map}
        \label{fig:milan_floor_map}
        \vspace{10pt}
    \end{subfigure}
    \hfill
    \begin{subfigure}{0.48\textwidth}
        \centering
        \includegraphics[width=\textwidth]{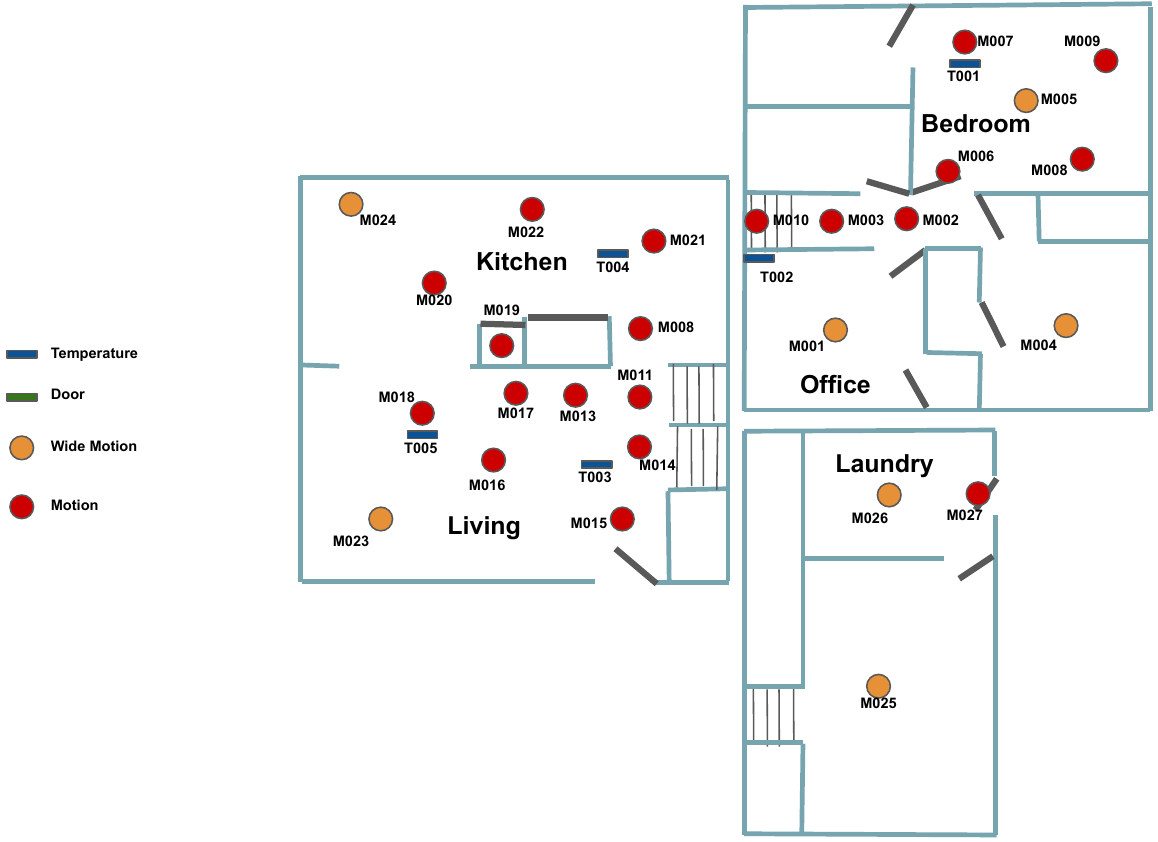}
        \caption{Cairo dataset floor map}
        \label{fig:cairo_floor_map}
        \vspace{10pt}
    \end{subfigure}
    \caption{Floor maps of Aruba, Milan, and Cairo Datasets.}
    \label{fig:casas_datasets_maps}
    \vspace{10pt}
\end{figure}

This appendix details information on three CASAS datasets—Aruba, Milan, and Cairo —created by Washington State University \cite{cook2012casas}. Data were collected from real homes equipped with smart sensors and inhabited by real residents. The sensor events log, as shown in Figure \ref{fig:casas_datasets_extract_example}, comprises motion sensors (sensor IDs beginning with "M"), door closure sensors (sensor IDs beginning with "D"), and temperature sensors (sensor IDs beginning with "T"). Each log entry consists of a single event per row and includes six columns: date, timestamp, sensor ID, sensor value, activity during which the sensor was triggered, and the activity status (its beginning or its end).
The datasets differ in housing structure, number of inhabitants, and the duration of activity labeling, with several months of data for each (details provided in Table \ref{tab:datasets_details}). Floor maps for each dataset are illustrated in Figure \ref{fig:casas_datasets_maps}.

Aruba and Milan feature single-floor homes with the same number of rooms but different layouts, whereas the Cairo home spans three floors and contains more rooms than the former two. The Aruba dataset captures the daily activities of a woman living alone, Milan documents the life of a woman and her dog, and Cairo describes a couple living with a dog.

The three datasets are unbalanced, with some activities less represented than others, as depicted in Table \ref{tab:datasets_activities_details}. This imbalance is due to the real-life settings of these homes and the lifestyles of the inhabitants. Furthermore the datasets do not cover exactly the same activities

\begin{table}[h]
\centering
\caption{Details of the three CASAS datasets}
\label{tab:datasets_details}
\begin{tabular}{cccc}
\toprule
\textbf{Dataset} & \textbf{Aruba} & \textbf{Milan} & \textbf{Cairo} \\
\hline
Residents & 1 & 1 + pet & 2 + pet \\
Number of Sensors & 39 & 33 & 27 \\
Number of Activities & 12 & 16 & 13 \\
Number of Days & 219 & 82 & 56 \\
\bottomrule
\end{tabular}

\end{table}

\begin{table}[h]
\centering
\caption{Aruba, Milan and Cairo labels details.}
\label{tab:datasets_activities_details}
\resizebox{\columnwidth}{!}{%
\begin{tabular}{lrllll}
\toprule
\multicolumn{2}{c}{\textbf{Aruba}}                                              & \multicolumn{2}{c}{\textbf{Milan}}                                              & \multicolumn{2}{c}{\textbf{Cairo}}                                              \\
\multicolumn{1}{c}{\textbf{Labels}} & \multicolumn{1}{c}{\textbf{\# occurence}} & \multicolumn{1}{c}{\textbf{Labels}} & \multicolumn{1}{c}{\textbf{\# occurence}} & \multicolumn{1}{c}{\textbf{Labels}} & \multicolumn{1}{c}{\textbf{\# occurence}} \\ \hline
Bed\_to\_Toilet                     & 155                                       & Bed\_to\_Toilet                     & 86                                        & Bed\_to\_toilet                     & 29                                        \\
Eating                              & 262                                       & Chores                              & 32                                        & Breakfast                           & 49                                        \\
Enter\_Home                         & 427                                       & Desk\_Activity                      & 54                                        & Dinner                              & 42                                        \\
Housekeeping                        & 34                                        & Dining\_Rm\_Activity                & 26                                        & Laundry                             & 10                                        \\
Leave\_Home                         & 427                                       & Eve\_Meds                           & 19                                        & Leave\_home                         & 69                                        \\
Meal\_Preparation                   & 1610                                      & Guest\_Bathroom                     & 335                                       & Lunch                               & 37                                        \\
Other                               & 6079                                      & Kitchen\_Activity                   & 741                                       & Night\_wandering                    & 66                                        \\
Relax                               & 2944                                      & Leave\_Home                         & 211                                       & Other                               & 579                                       \\
Respirate                           & 6                                         & Master\_Bathroom                    & 307                                       & R1\_sleep                           & 55                                        \\
Sleeping                            & 416                                       & Master\_Bedroom\_Activity           & 204                                       & R1\_wake                            & 53                                        \\
Wash\_Dishes                        & 64                                        & Meditate                            & 17                                        & R1\_work\_in\_office                & 46                                        \\
Work                                & 171                                       & Morning\_Meds                       & 41                                        & R2\_sleep                           & 53                                        \\
                                    & \multicolumn{1}{l}{}                      & Other                               & 1776                                      & R2\_take\_medicine                  & 44                                        \\
                                    & \multicolumn{1}{l}{}                      & Read                                & 368                                       & R2\_wake                            & 64                                        \\
                                    & \multicolumn{1}{l}{}                      & Sleep                               & 166                                       &                                     &                                           \\
                                    & \multicolumn{1}{l}{}                      & Watch\_TV                           & 153                                       &                                     &    \\
                                    \bottomrule
\end{tabular}%
}
\end{table}

\clearpage
\section{Experiments Reproduction}
\label{sec:liciotti_rep}

This appendix presents the scores and results of our replication of the Liciotti et al. \cite{liciotti_lstm} experiments. Table \ref{tab:liciotti_reproduced} below includes results from the original paper for the Milan and Cairo datasets, alongside our findings. It's important to note that the Aruba dataset was not explored in the original study. We conducted our experiments 10 times, and the table reflects the average results of these repetitions. In line with the original paper, we employed a 3-fold cross-validation evaluation method for each dataset and regrouping the original dataset labels under meta-activities using the same activity remapping as in the original study. Additionally, we adhered to the same hyperparameters defined in the original paper to ensure consistency in our replication process. Our findings demonstrate that we achieved results closely similar to those in the original paper, confirming the accuracy of our implementation of the algorithm.

\begin{table}[h]
\centering
\caption{Reproduction results of the Bi-LSTM architecture as proposed by Liciotti et al. Our replication of these experiments was conducted 10 times to ensure reliability and consistency of the results.}
\label{tab:liciotti_reproduced}
\resizebox{\columnwidth}{!}{%
\begin{tabular}{lcccccc}
\toprule
                   & \multicolumn{2}{c}{Aruba}                                                                                                                 & \multicolumn{2}{c}{Milan}                                                                                                                 & \multicolumn{2}{c}{Cairo}                                                                                                                 \\
                   & \begin{tabular}[c]{@{}c@{}}Liciotti et al.\\ (paper)\end{tabular} & \begin{tabular}[c]{@{}c@{}}Liciotti et al.\\ (reproduce)\end{tabular} & \begin{tabular}[c]{@{}c@{}}Liciotti et al.\\ (paper)\end{tabular} & \begin{tabular}[c]{@{}c@{}}Liciotti et al.\\ (reproduce)\end{tabular} & \begin{tabular}[c]{@{}c@{}}Liciotti et al.\\ (paper)\end{tabular} & \begin{tabular}[c]{@{}c@{}}Liciotti et al.\\ (reproduce)\end{tabular} \\ \hline
Accuracy           & NA                                                                & 96.17\%                                                               & 94.12\%                                                           & 90.70\%                                                               & 86.90\%                                                           & 86.67\%                                                               \\
Precision          & NA                                                                & 92.73\%                                                               & NA                                                                & 82.33\%                                                               & NA                                                                & 79.67\%                                                               \\
Recall             & NA                                                                & 90.17\%                                                               & NA                                                                & 77.00\%                                                               & NA                                                                & 75.00\%                                                               \\
F1 Score           & NA                                                                & 91.17\%                                                               & NA                                                                & 79.33\%                                                               & NA                                                                & 76.33\%                                                               \\
Balanced Accuracy  & NA                                                                & 90.37\%                                                               & NA                                                                & 76.88\%                                                               & NA                                                                & 74.98\%                                                               \\
Weighted Precision & NA                                                                & 96.30\%                                                               & 94.00\%                                                           & 90.33\%                                                               & 86.67\%                                                           & 86.33\%                                                               \\
Weighted Recall    & NA                                                                & 96.10\%                                                               & 94.00\%                                                           & 90.67\%                                                               & 87.00\%                                                           & 86.67\%                                                               \\
Weighted F1 Score  & NA                                                                & 96.03\%                                                               & 94.00\%                                                           & 90.33\%                                                               & 86.67\%                                                           & 86.33\%\\
\bottomrule
\end{tabular}%
}
\end{table}
\section{Activities Embedding}
\label{sec:embedding}

In this appendix, we present the embeddings of activities from three datasets: Aruba, Milan, and Cairo. These embeddings effectively capture the relationships between various activities and provide a visual representation of the feature space for each activity. We conduct a comparative analysis of the embeddings generated by the ELMo and GPT models. The left images in our figures showcase activities embedded using the ELMo models, while the right images illustrate those embedded by the GPT models.

\begin{figure*}[bt]
\centering
     \begin{subfigure}{\textwidth}
         \centering
         \includegraphics[width=\textwidth]{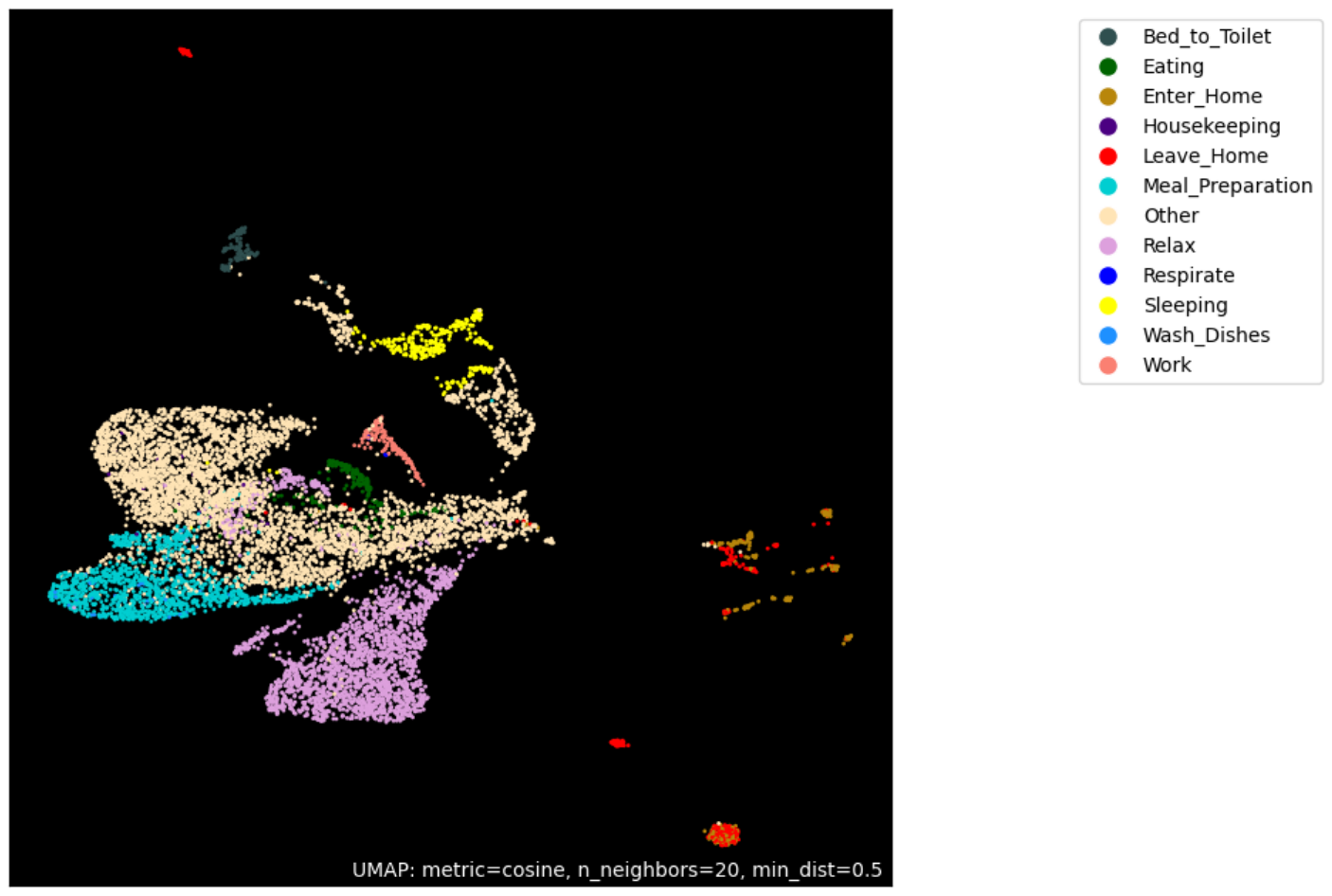}
         \vspace{-50pt}
         \caption{ELMo}
         \label{fig:Aruba_ELmo_Emb}
         \vspace{-30pt}
     \end{subfigure}
     \hfill
     \begin{subfigure}{\textwidth}
         \centering
         \includegraphics[width=\textwidth]{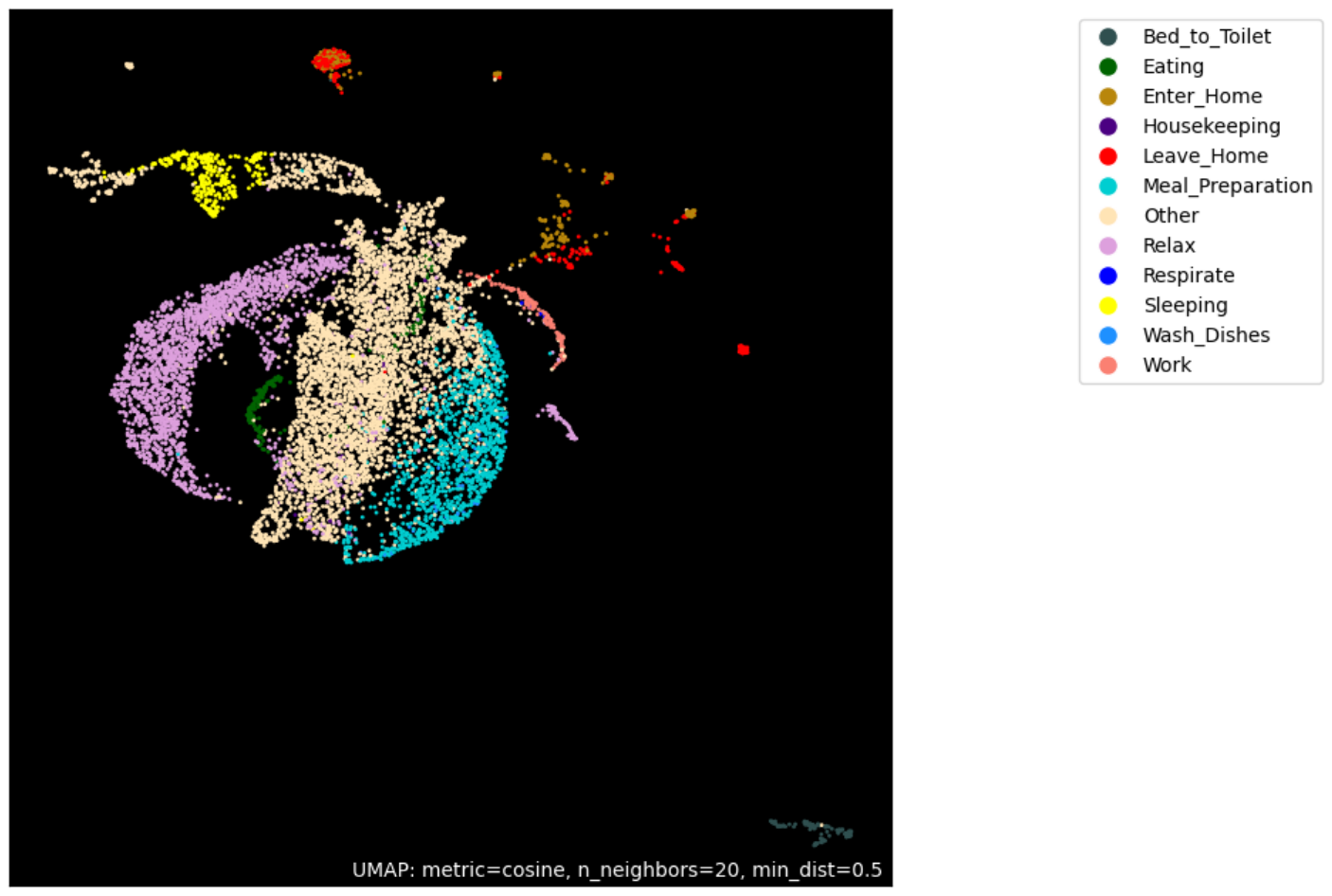}
         \vspace{-50pt}
         \caption{GPT}
         \label{fig:Aruba_GPT_Emb}
         \vspace{10pt}
     \end{subfigure}
    \caption{Visualization of Activity Embeddings for the Aruba Dataset (Test Set) Generated by ELMo (Top) and GPT (Bottom)}
    \label{fig:emb_aruba}
    \vspace{10pt}
\end{figure*}

\begin{figure*}[bt]
\centering
     \begin{subfigure}{\textwidth}
         \centering
         \includegraphics[width=\textwidth]{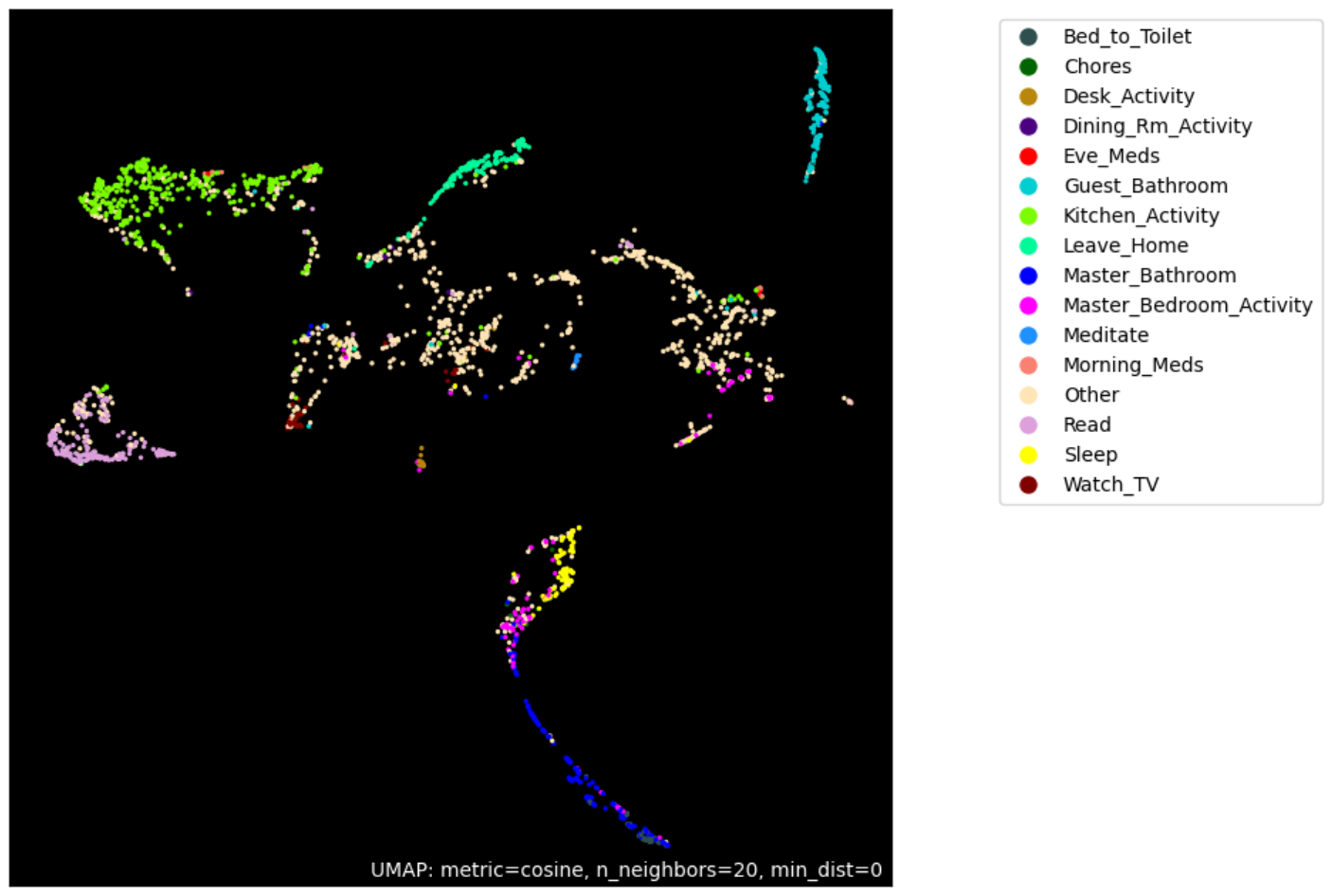}
         \vspace{-50pt}
        \caption{ELMo}
         \label{fig:Milan_ELmo_Emb}
         \vspace{-30pt}
     \end{subfigure}
     \begin{subfigure}{\textwidth}
         \centering
         \includegraphics[width=\textwidth]{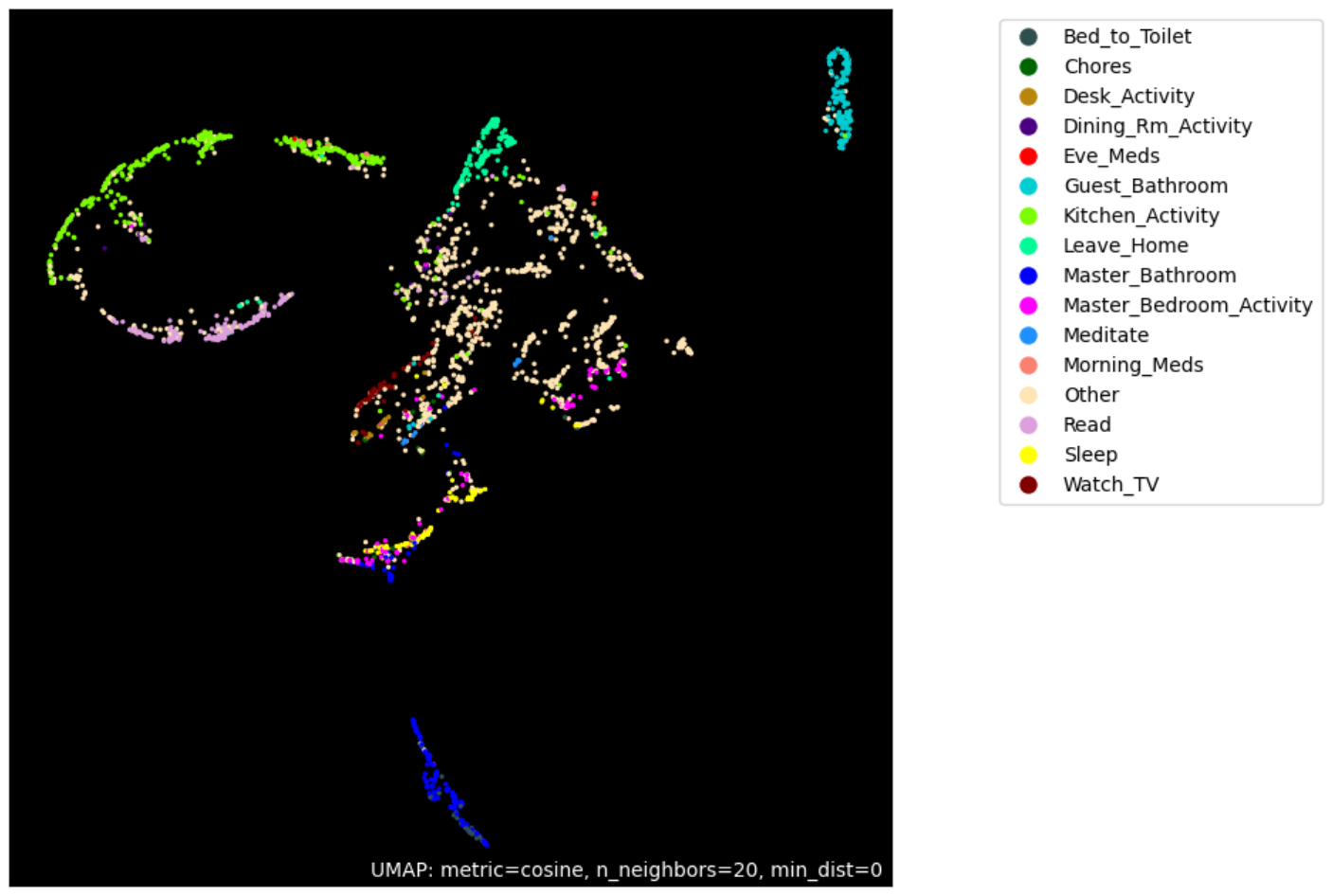}
         \vspace{-50pt}
         \caption{GPT}
         \label{fig:Milan_GPT_Emb}
         \vspace{10pt}
     \end{subfigure}
    \caption{Visualization of Activity Embeddings for the Milan Dataset (Test Set) Generated by ELMo (Top) and GPT (Bottom)}
    \label{fig:emb_milan}
    \vspace{10pt}
\end{figure*}

\begin{figure}[tb]
    \centering
     \begin{subfigure}{\textwidth}
         \centering
         \includegraphics[width=\textwidth]{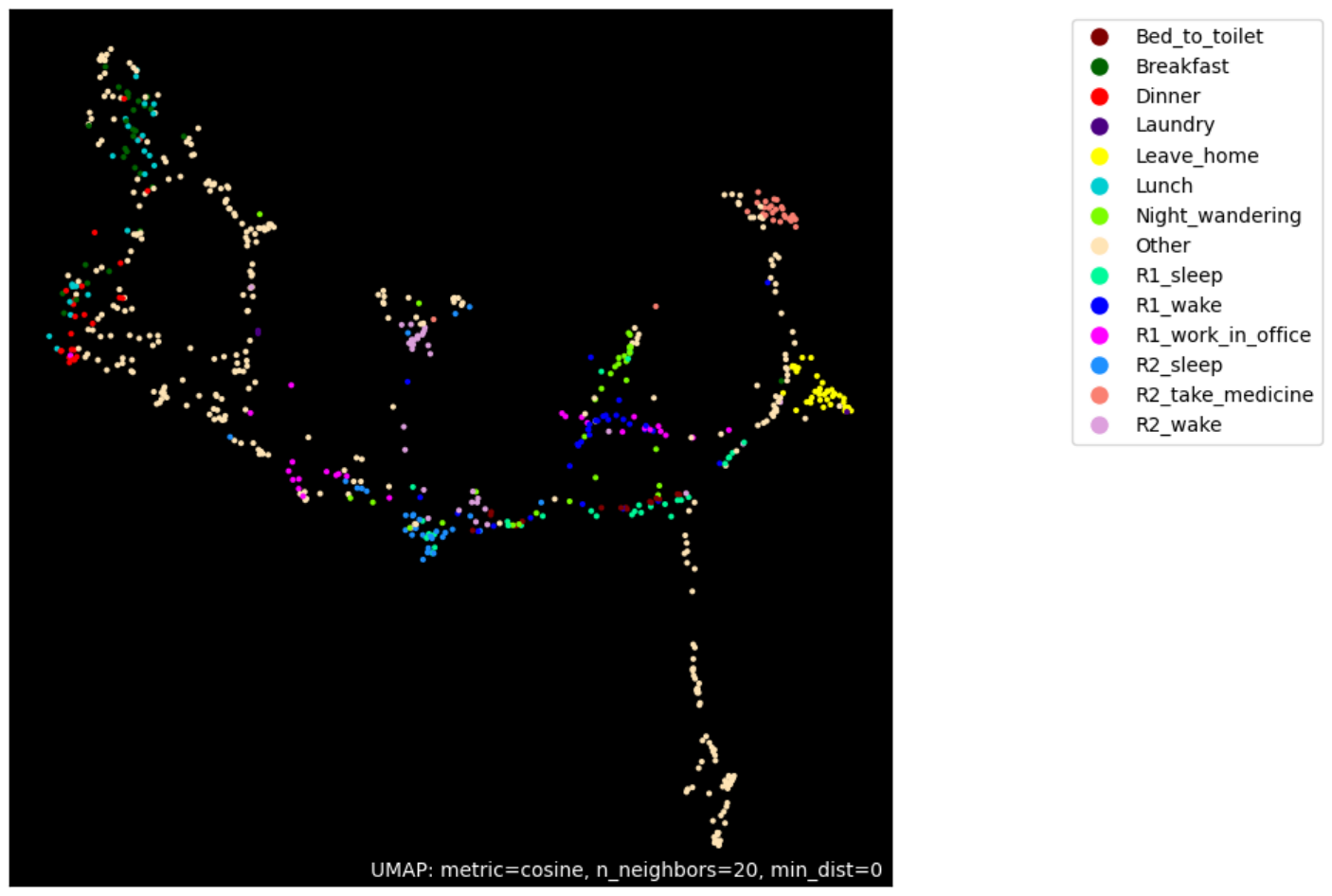}
         \vspace{-50pt}
         \caption{ELMo}
         \label{fig:Cairo_ELmo_Emb}
         \vspace{-30pt}
     \end{subfigure}
     \begin{subfigure}{\textwidth}
         \centering
         \includegraphics[width=\textwidth]{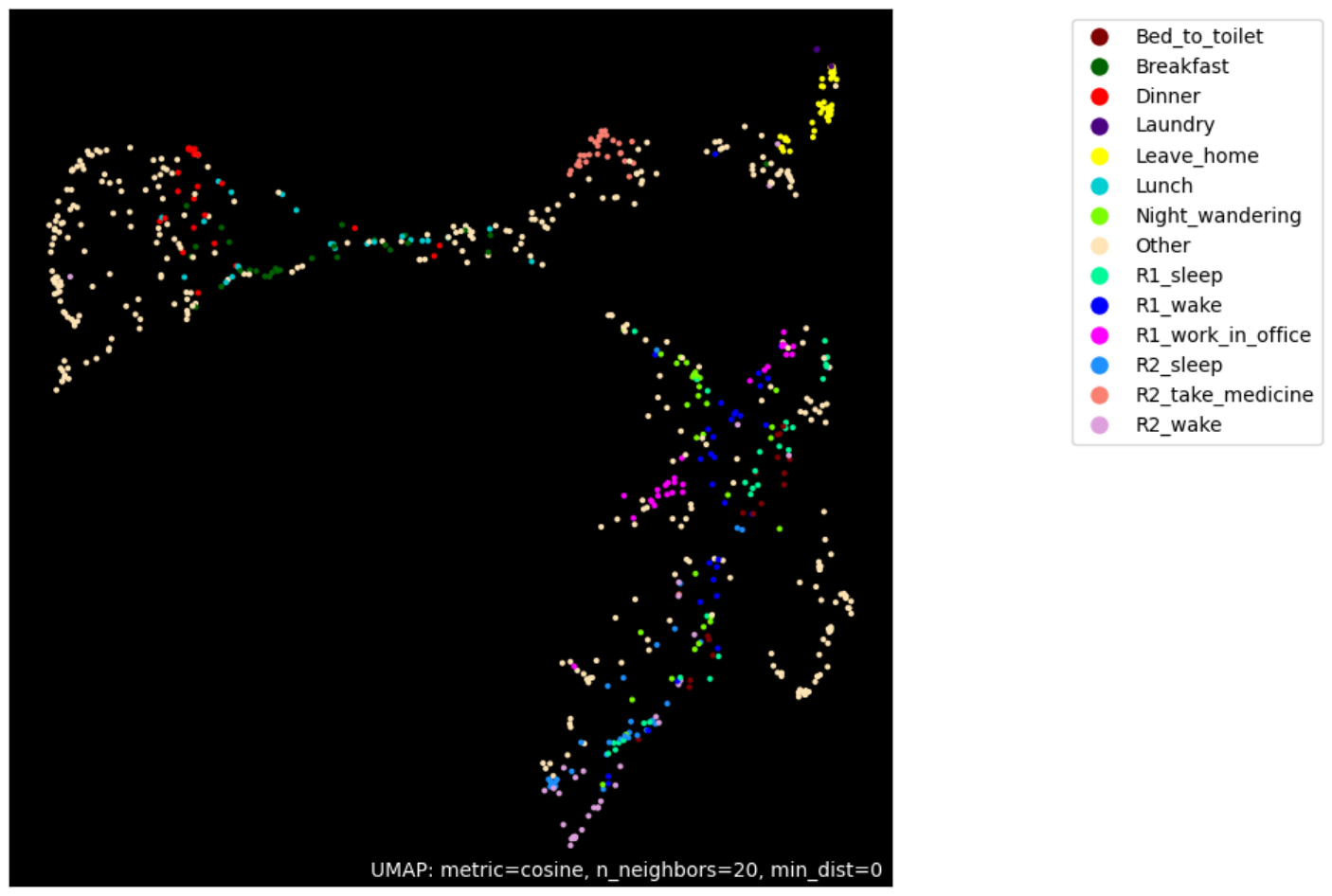}
         \vspace{-50pt}
         \caption{GPT}
         \label{fig:Cairo_GPT_Emb}
         \vspace{10pt}
     \end{subfigure}
    \caption{Visualization of Activity Embeddings for the Cairo Dataset (Test Set) Generated by ELMo (Top) and GPT (Bottom)}
    \label{fig:emb_cairo}
\end{figure}

To generate the embedding of activity sequences, each pre-segmented activity from the datasets is processed through the respective embedding model. The resulting output vectors are then averaged to produce a unique vector representation for each activity. Subsequently, these vectors are dimensionally reduced using the Uniform Manifold Approximation and Projection (UMAP) algorithm, enabling their representation in a 2-dimensional space. Each point in this space is color-coded according to its corresponding activity label. Figures \ref{fig:emb_aruba}, \ref{fig:emb_milan}, and \ref{fig:emb_cairo} demonstrate the ELMo-based and GPT-based activity sequence embeddings for the Aruba, Milan, and Cairo datasets, respectively.

We observe that the embeddings are generally similar. However, it is crucial to note that the GPT embedding is trained from random chunks that may not contain only one activity but potentially a subpart or parts of multiple activities in contrary of ELMo which use pre-segmented activities. We can observe, in the Milan dataset, the GPT embedding appears to more distinctly separate some clusters, for instance, it isolate the 'master\_bathroom' activities. For the Cairo dataset, the GPT embedding seems to segregate all activities into two main clusters: one grouping day activities and the other night activities.


\end{document}